%% file: acl.tex
\pdfoutput=1

\documentclass[11pt]{article}

\usepackage{acl}

\usepackage{times}
\usepackage{latexsym}

\usepackage[T1]{fontenc}

\usepackage[utf8]{inputenc}

\usepackage{microtype}

\usepackage{xspace}
\usepackage{xcolor}
\usepackage{soul}
\usepackage{booktabs}
\usepackage{graphicx}
\usepackage{amsmath}
\usepackage{comment}
\usepackage{svg}
\usepackage{multirow}
\usepackage{tabularx}
\usepackage{amssymb}

\usepackage{ulem}

\definecolor{colorGrammarUsage}{HTML}{F5F5F5}
\definecolor{colorRedundant}{HTML}{FFC2BA}
\definecolor{colorOffPrompt}{HTML}{FFFFB3}
\definecolor{colorSelfContradiction}{HTML}{B9DEFF}
\definecolor{colorIncoherent}{HTML}{FCCDE5}
\definecolor{colorJargon}{HTML}{FEC88B}
\definecolor{colorBadMath}{HTML}{D9D9D9}
\definecolor{colorCommonsense}{HTML}{D7F3E7}
\definecolor{colorEncyclopedic}{HTML}{C8E792}
\definecolor{colorNeedsGoogle}{HTML}{E4B7FF}

\definecolor{colorFuture}{HTML}{EDFFFF}

\definecolor{someSpanText}{HTML}{000000} 

\setul{0.1ex}{0.4ex}

\newcommand{\errGrammarUsage[1]}{\sethlcolor{colorGrammarUsage}{\textbf{\hl{~#1~}}}}
\newcommand{\errRedundant[1]}{{\sethlcolor{colorRedundant}\textcolor{someSpanText}{\textbf{\hl{~#1~}}}}}
\newcommand{\errRedundantAntecedent[1]}{\setulcolor{colorRedundant}{\textcolor{someSpanText}{\textbf{\ul{~#1~}}}}}
\newcommand{\errOffPrompt[1]}{\sethlcolor{colorOffPrompt}{\textbf{\hl{~#1~}}}}
\newcommand{\errSelfContradiction[1]}{\sethlcolor{colorSelfContradiction}{\textcolor{someSpanText}{\textbf{\hl{~#1~}}}}}
\newcommand{\errSelfContradictionAntecedent[1]}{\setulcolor{colorSelfContradiction}{{\textbf{\ul{~#1~}}}}}
\newcommand{\errIncoherent[1]}{\sethlcolor{colorIncoherent}{\textcolor{someSpanText}{\textbf{\hl{~#1~}}}}}
\newcommand{\errJargon[1]}{\sethlcolor{colorJargon}{\textcolor{someSpanText}{\textbf{\hl{~#1~}}}}}
\newcommand{\errBadMath[1]}{\sethlcolor{colorBadMath}{\textcolor{someSpanText}{\textbf{\hl{~#1~}}}}}
\newcommand{\errCommonsense[1]}{\sethlcolor{colorCommonsense}{\textbf{\hl{~#1~}}}}
\newcommand{\errEncyclopedic[1]}{\sethlcolor{colorEncyclopedic}{\textcolor{someSpanText}{\textbf{\hl{~#1~}}}}}
\newcommand{\errNeedsGoogle[1]}{\sethlcolor{colorNeedsGoogle}{\textcolor{someSpanText}{\textbf{\hl{~#1~}}}}}

\newcommand{\nameGrammarUsage}{\errGrammarUsage[Grammar and Usage]}
\newcommand{\nameRedundant}{\errRedundant[Redundant]}
\newcommand{\nameRedundantAntecedent}{\errRedundantAntecedent[antecedent]}
\newcommand{\nameOffPrompt}{\errOffPrompt[Off-Prompt]}
\newcommand{\nameSelfContradiction}{\errSelfContradiction[Self-Contradiction]}
\newcommand{\nameSelfContradictionAntecedent}{\errSelfContradictionAntecedent[antecedent]}
\newcommand{\nameIncoherent}{\errIncoherent[Incoherent]}
\newcommand{\nameJargon}{\errJargon[Technical Jargon]}
\newcommand{\nameBadMath}{\errBadMath[Bad Math]}
\newcommand{\nameCommonsense}{\errCommonsense[Commonsense]}
\newcommand{\nameEncyclopedic}{\errEncyclopedic[Encyclopedic]}
\newcommand{\nameNeedsGoogle}{\errNeedsGoogle[Needs Google]}

\newcommand{\futureGeneric}{\sethlcolor{colorFuture}{\textbf{\hl{~Generic~}}}}
\newcommand{\futureAdequacy}{\sethlcolor{colorFuture}{\textbf{\hl{~Adequacy~}}}}

\newcommand{\iconDecreasing}{\includegraphics[width=10pt]{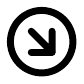}}
\newcommand{\iconHumansHighest}{\includegraphics[width=10pt]{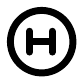}}
\newcommand{\iconModelPlateau}{\includegraphics[width=10pt]{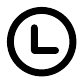}}
\newcommand{\iconUpDown}{\includegraphics[width=10pt]{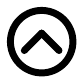}}

\newcommand{\method}{\textsc{Scarecrow}\xspace}

\title{
Is GPT-3 Text Indistinguishable from Human Text? \\
\textsc{Scarecrow:} A Framework for Scrutinizing Machine Text
}

\author{
Yao Dou\thanks{\hspace{4pt}Equal contribution}\hspace{4pt}$^{\dagger}$~~~Maxwell Forbes\footnotemark[1]\hspace{4pt}$^{\dagger}$~~~Rik Koncel-Kedziorski$^{\dagger}$~~~Noah A. Smith$^{\dagger\ddagger}$~~~Yejin Choi$^{\dagger\ddagger}$ \\

$^\dagger$Paul G. Allen School of Computer Science \& Engineering, University of Washington\\
$^\ddagger$Allen Institute for AI\\
{
\tt\small
\{douy,mbforbes,nasmith,yejin\}@cs.washington.edu~~~
kedzior@uw.edu
}
}

\begin{document}
\maketitle

\input{src/abstract}
\input{src/intro}
\input{src/takeaways}
\input{src/bg}
\input{src/method}
\input{src/data}
\input{src/prediction}
\input{src/related_work}
\input{src/conclusion}

\section*{Acknowledgments}
The authors thank members of xlab for their feedback on this work.
This research is supported in part by NSF (IIS-1714566), DARPA MCS program through NIWC Pacific (N66001-19-2-4031), DARPA SemaFor program, and Allen Institute for AI.

\bibliography{anthology,custom}
\bibliographystyle{acl_natbib}

\newpage

\appendix

\input{src/schema}
\input{src/annotation-more}
\input{src/verification}
\input{src/data-details}
\input{src/analysis}
\input{src/more-analysis}
\input{src/future-work}

\end{document}

%% file: src/abstract.tex
\begin{abstract}
Modern neural language models can produce remarkably fluent and grammatical text. So much, in fact, that recent work by \newcite{clark-etal-2021-thats} has reported that conventional crowdsourcing can no longer reliably distinguish between machine-authored (GPT-3) and human-authored writing. 
%
%
%
As errors in machine generations become ever subtler and harder to spot, it poses a new challenge to the research community for robust machine text evaluation. 

We propose a new framework called \method for scrutinizing machine text via crowd annotation. To support the broad range of real machine errors that can be identified by laypeople, the ten error categories of \method---such as \errRedundant[redundancy], \errCommonsense[commonsense errors], and \errIncoherent[incoherence]---are identified through several rounds of crowd annotation experiments without a predefined ontology.

We then use \method to collect over 41k error spans in human-written and machine-generated paragraphs of English language news text.
We isolate factors for detailed analysis, including parameter count, training data, and various decoding-time configurations. 
Our approach successfully quantifies measurable gaps between human authored text and generations from models of several sizes, including fourteen configurations of GPT-3. 
In addition, our analysis unveils new insights, with detailed rationales provided by laypeople, e.g., that the commonsense capabilities have been improving with larger models while math capabilities have not, and that the choices of simple decoding hyperparameters can make remarkable differences on the perceived quality of machine text. 
We release our training material, annotation toolkit and dataset at \textbf{\url{https://yao-dou.github.io/scarecrow/}}.
\end{abstract}

%% file: src/intro.tex
\section{Introduction}

\input{fig/example-filled-fig}

\input{fig/ontology-table}

\input{fig/per-label-wide}



\newcite{clark-etal-2021-thats} demonstrated the challenges of human evaluation in the era of GPT-3 \cite{brown2020language}, as crowd workers are no longer able to reliably distinguish GPT-3's generations  from human-written text.

Or are they? In this paper, we propose a new framework for systematically scrutinizing machine text so that even crowd workers, despite the known challenges reported by recent literature, can successfully critique seemingly fluent generations. We not only quantify a measurable gap between machine text and human text, but reveal the distributions of specific categories of issues, and pinpoint their occurrences in text written by several sizes of language models as well as humans.


To achieve this, we develop \method, a methodology for eliciting categorical judgements of errors in machine-generated text from crowd workers. 
One goal in
natural language generation (NLG)
is to produce fluent outputs which can be read by laypeople.
As such, we propose that important errors to address are those which are recognized by readers without NLP expertise. 
Our framework allows crowd workers to annotate problems in model outputs at the span level.
A single such annotation is shown in Figure~\ref{fig:example-filled}.

To make this possible, we establish a categorization of shortcomings commonly found in machine generated text (Table \ref{tab:onto-ex}).
This error schema covers a broad scope of problems as identified by experts, but has been honed
according to what is salient to non-expert readers
through several pilot rounds of crowd annotation without a fixed label set.
The result is a framework that is usable by everyday people with minimal training, but covers the error phenomena found in real machine-generated text.
Labeling spans of text using specific error types creates a picture of contemporary model generations with an unprecedented level of detail.
In contrast to judging text holistically \cite{celikyilmaz2021evaluation}, insights from this method are specific and practical, as it measures exactly how and where problems arise.

We conduct a large-scale analysis of human-written and machine-generated text using \method, collecting 13k annotations of 1.3k paragraphs, amassing 41k spans labeled with error type, severity, and an explanation.
Through this, we characterize in which ways GPT-3's generations are better than those of previous models, and which aspects do not improve with increased data and parameters.
We also provide a rigorous error analysis of text generated by several other contemporary language models, examining the impact of model size, training data, and decoding strategy. 

We provide our detailed
annotator training system and task interface so that future researchers may
employ and refine them for
error analyses of machine-generated text.
We hope this will contribute to the standardization of NLG human evaluation \cite{howcroft-etal-2020-twenty}.

%% file: fig/example-filled-fig.tex
\begin{figure}[t!]

\begin{center}
\includegraphics[width=0.99\linewidth]{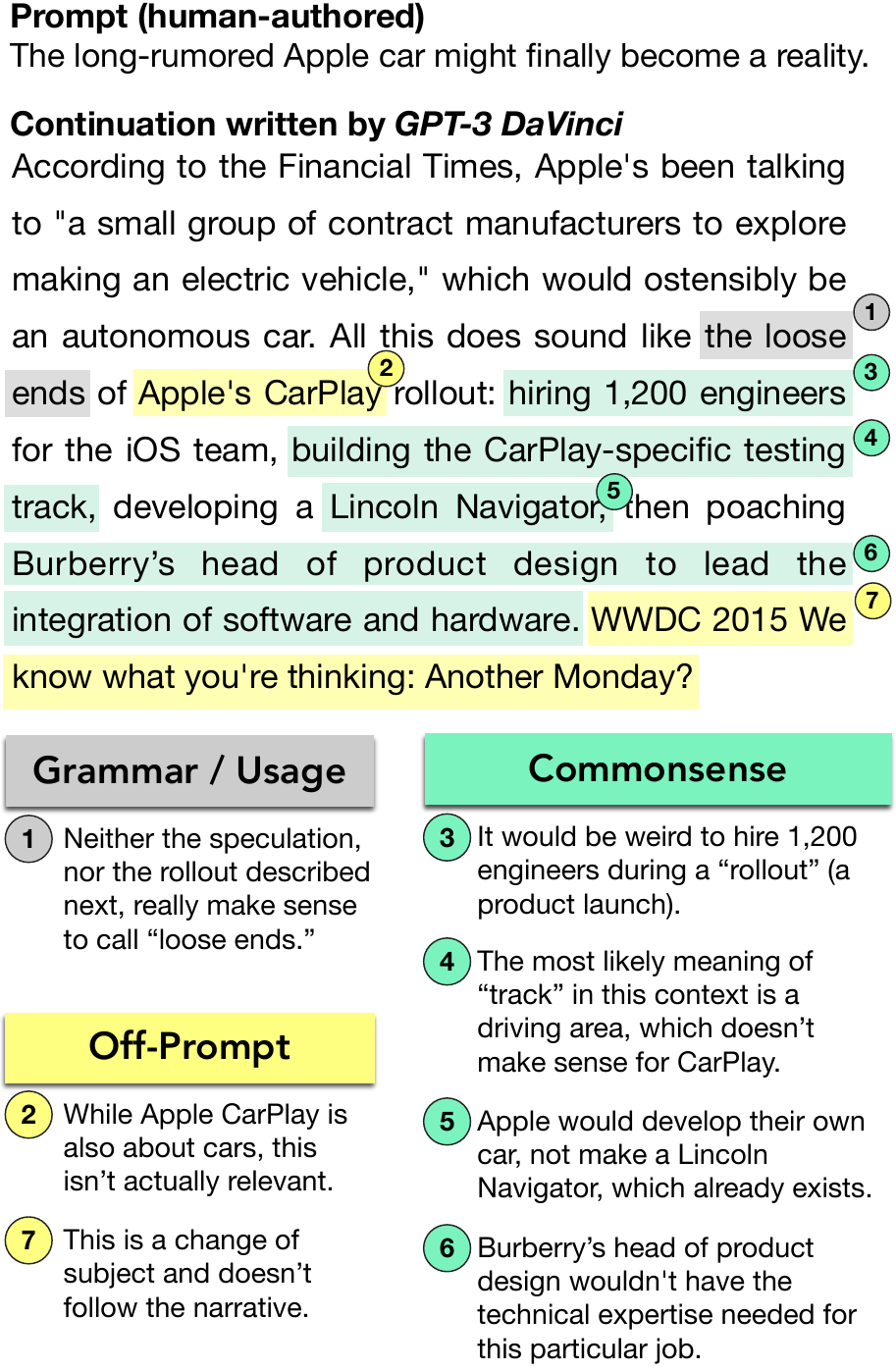}
\end{center}

\caption{
After a model (here, GPT-3 DaVinci) has read the prompt (top sentence) and generated a continuation (next paragraph), the \method annotation framework provides a systematic way for humans to mark issues throughout the text and explain what is wrong. Our own annotations are pictured here.
}
\label{fig:example-filled}
\end{figure}

%% file: fig/ontology-table.tex
\renewcommand{\arraystretch}{1.2} 
\begin{table*}[h]
\centering
\small
\begin{tabularx}{\textwidth}{lXX}
\toprule
\textsc{error type} & \textsc{definition} &  \textsc{example} \\ \midrule

{\bf Language Errors} &&\\ 
\hspace{1em} \nameGrammarUsage & Missing, extra, incorrect, or out of order words & \ldots explaining how cats feel \errGrammarUsage[emoticons] \ldots\\
\hspace{1em} \nameOffPrompt & Generation is unrelated to or contradicts prompt &\textbf{\textsc{prompt:}} Dogs are the new kids. \textbf{\textsc{generation:}} Visiting \errOffPrompt[the dentist can be scary] \\
\hspace{1em} \nameRedundant & Lexical, semantic, or execessive topical repetition &  Merchants worry about \errRedundantAntecedent[poor service] \errRedundant[or service that is bad] \ldots \\
\hspace{1em} \nameSelfContradiction & Generation contradicts itself & Amtrak plans to
    \errSelfContradictionAntecedent[lay off many employees,] though \errSelfContradiction[it has no plans cut employee hours.] \\
\hspace{1em} \nameIncoherent & Confusing, but not any error type above & Mary gave her kids cheese toast but
    \errIncoherent[drew a map of it on her toast.] \\ \midrule
    
{\bf Factual Errors} && \\
\hspace{1em} \nameBadMath & Math or conversion mistakes & {\it \ldots it costs over \pounds1,000 \errBadMath[(\$18,868)] \ldots} \\
\hspace{1em} \nameEncyclopedic & Facts that annotator knows are wrong & {\it \errEncyclopedic[Japanese Prime Minister Justin Trudeau] said Monday \ldots } \\
\hspace{1em} \nameCommonsense& Violates basic understanding of the world & The dress was made at the \errCommonsense[spa.] \\ \midrule
    
{\bf Reader Issues} && \\
\hspace{1em} \nameNeedsGoogle & Search needed to verify claim & {\it \errNeedsGoogle[Jose Celana, an artist based in Pensacola, FL], \ldots } \\
\hspace{1em} \nameJargon & Text requires expertise to understand & {\it \ldots an 800-megawatt \errJargon[photovoltaic] plant was built \ldots }\\ 

\bottomrule
\end{tabularx}%
\caption{Error types in the \method framework, grouped into three categories. The categories are explained further in \S\ref{sec:error-types}, and detailed definitions and examples for each error type is provided in Appendix \ref{sec:schema}.
}
\label{tab:onto-ex}
\end{table*}
\renewcommand{\arraystretch}{1} 

%% file: fig/per-label-wide.tex
\begin{figure*}[t]

\begin{center}

\includegraphics[width=0.195\linewidth]{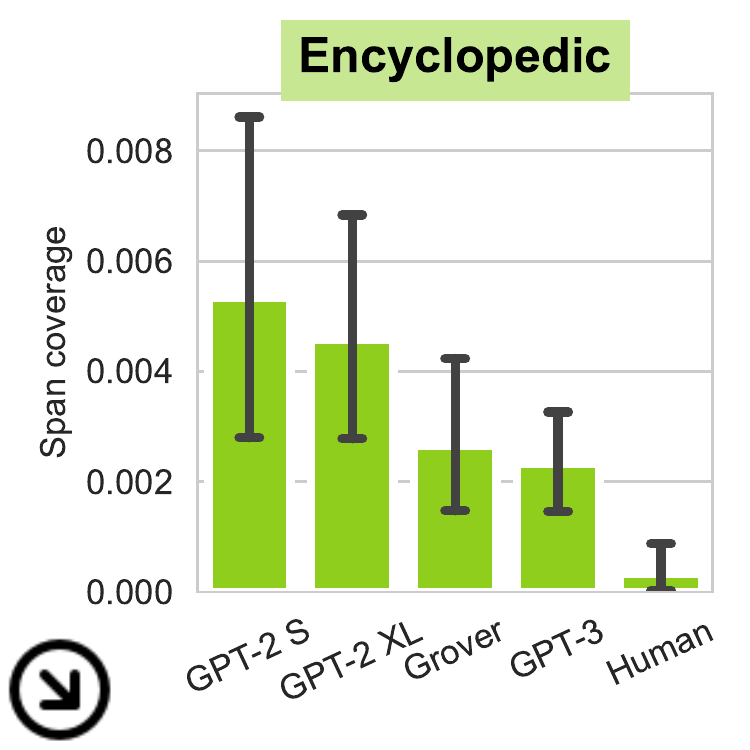}
\includegraphics[width=0.195\linewidth]{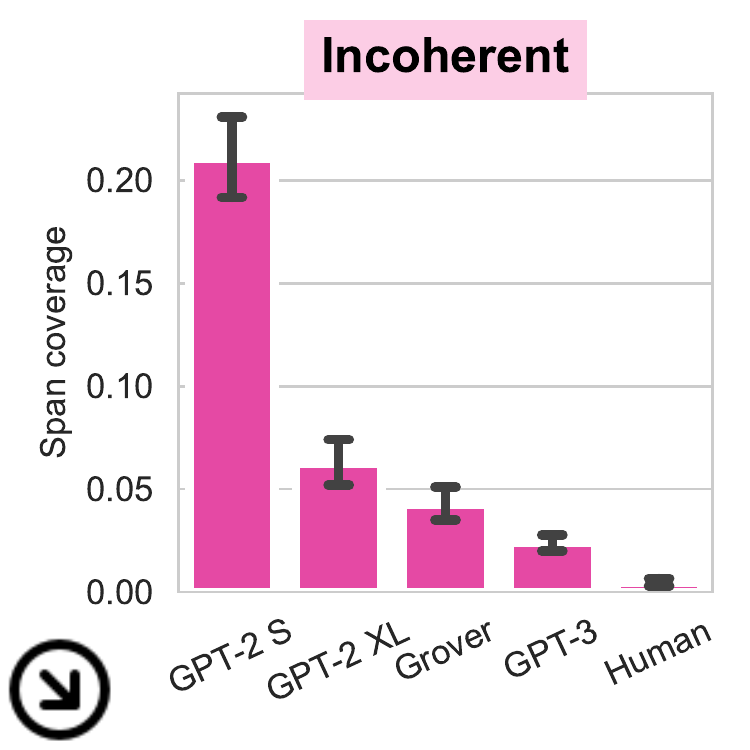}
\includegraphics[width=0.195\linewidth]{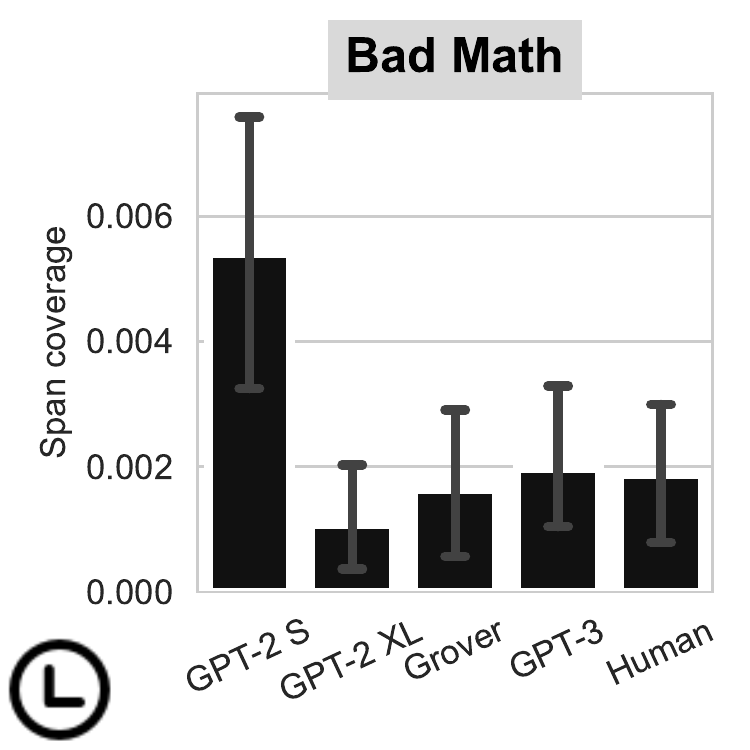}
\includegraphics[width=0.195\linewidth]{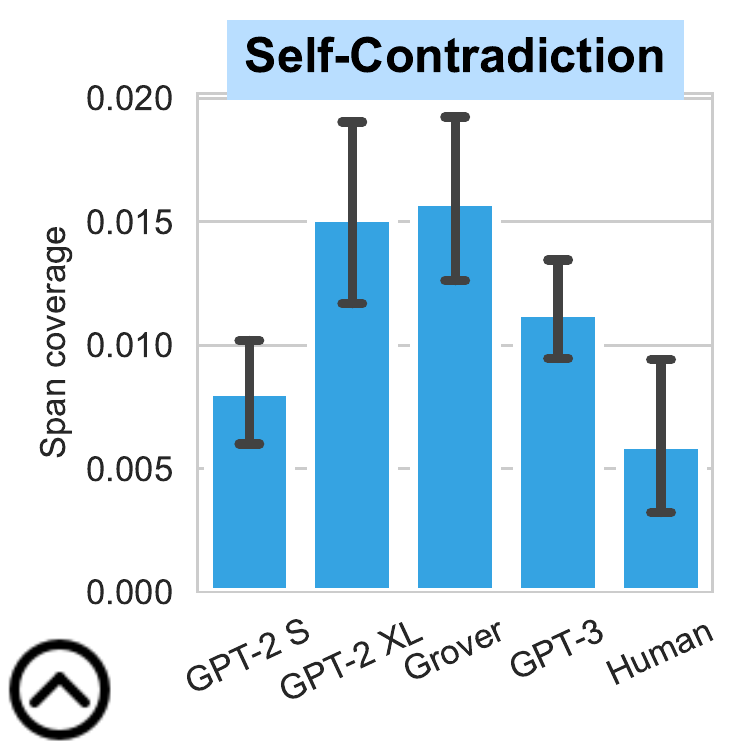}
\includegraphics[width=0.195\linewidth]{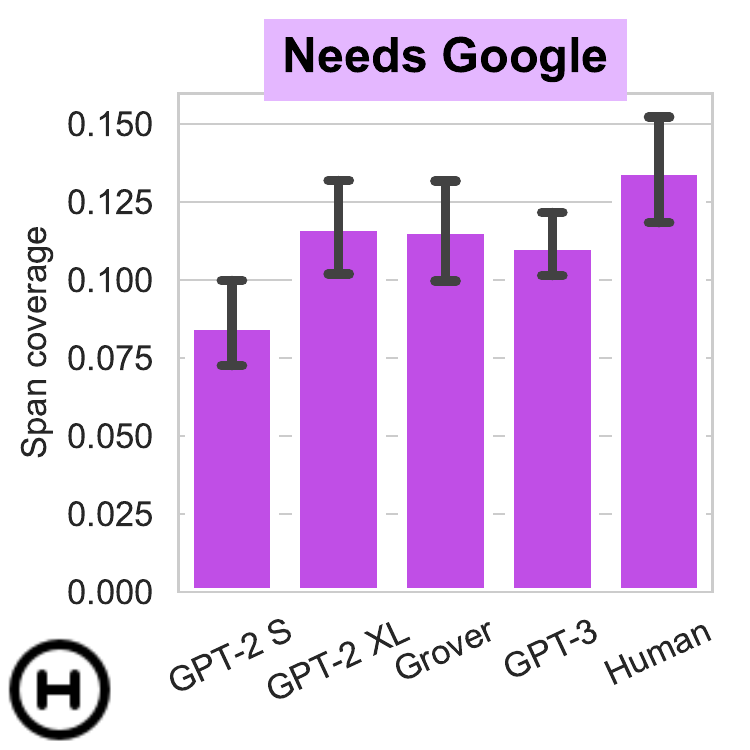}

\includegraphics[width=0.195\linewidth]{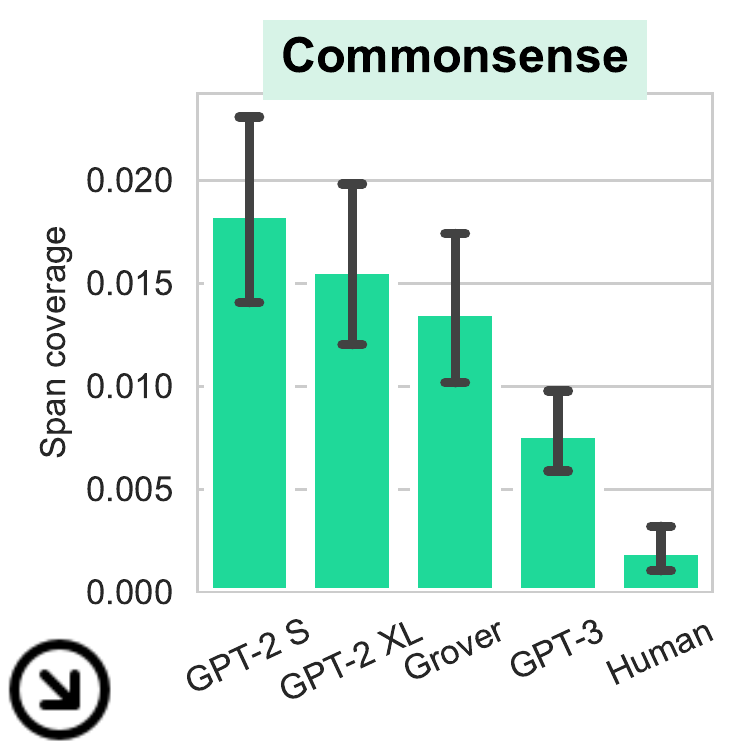}
\includegraphics[width=0.195\linewidth]{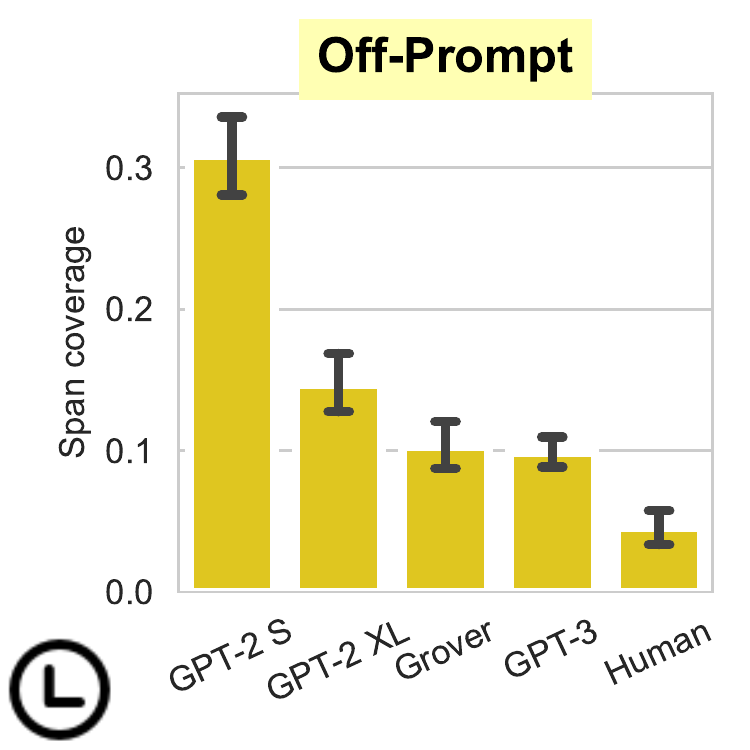}
\includegraphics[width=0.195\linewidth]{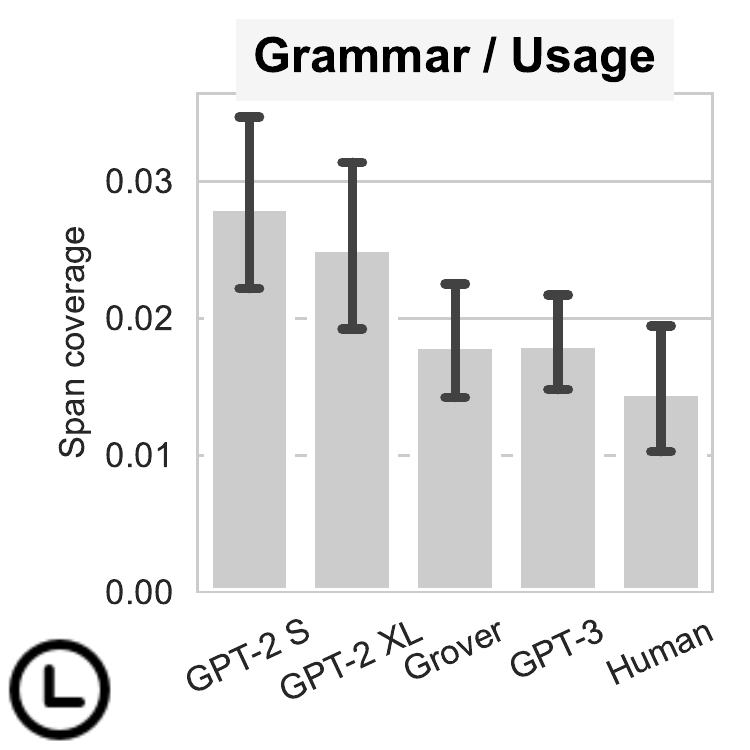}
\includegraphics[width=0.195\linewidth]{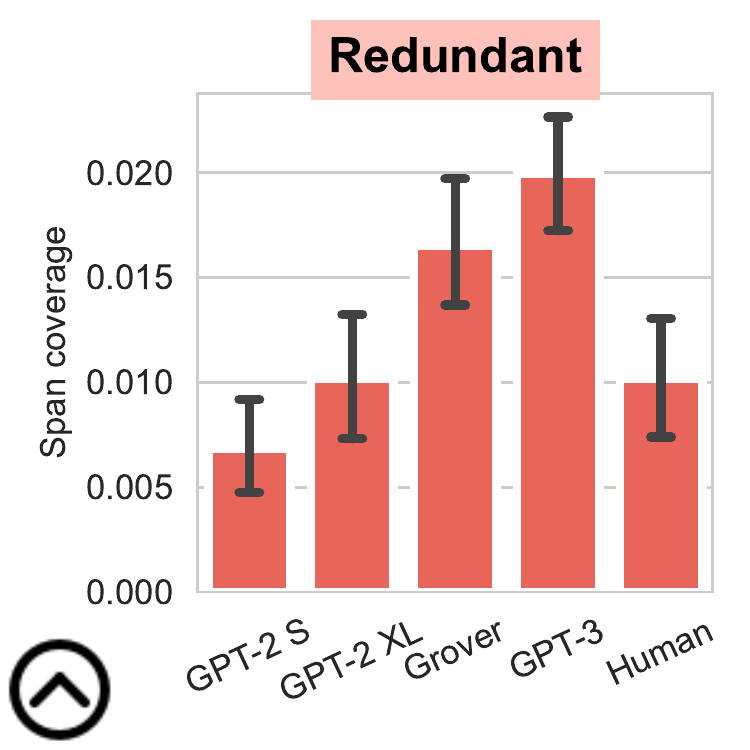}
\includegraphics[width=0.195\linewidth]{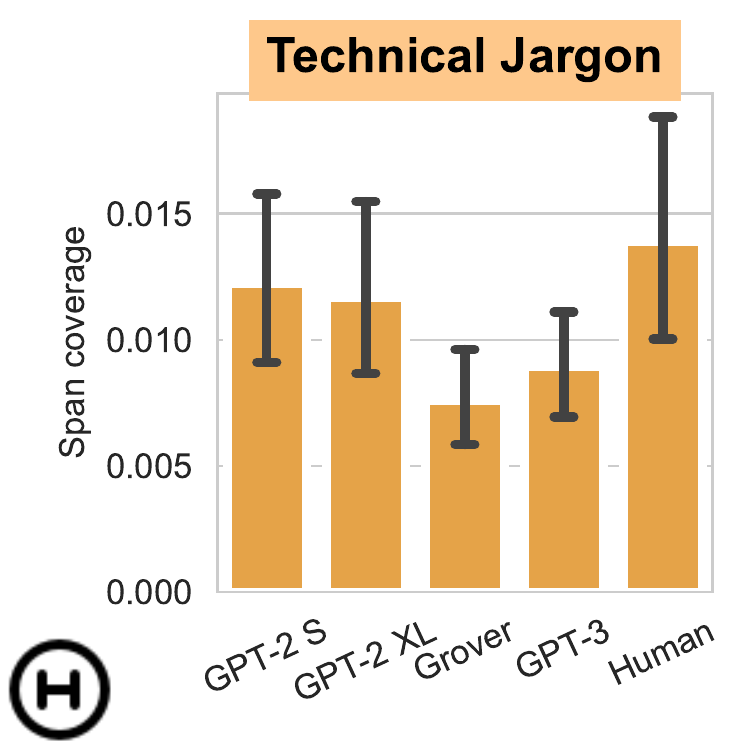}

\end{center}

\caption{
Average portion of tokens annotated with each error type ($y$-axis) across models ($x$-axis), with 95\% confidence intervals.
We group the trends into several broad categories.
\iconDecreasing \textbf{Decreasing:} fine-tuning and increasing model size improves performance.
\iconModelPlateau \textbf{Model plateau:} increasing model size to GPT-3 does not correlate with further improvements.
\iconUpDown \textbf{Rising and falling:} errors become more prevalent with some models, then improve.
\iconHumansHighest \textbf{Humans highest:} these spans are labeled most on human-authored text; both are \textit{reader issues} (distinct from \textit{errors}; see Table \ref{tab:onto-ex}).
Details: all models, including GPT-3, use the same ``apples-to-apples'' decoding hyperparameters: top-$p$=0.96, temperature=1, and no frequency penalty.
}

\label{fig:per-label-wide}
\end{figure*}

%% file: src/takeaways.tex
\section{Key Findings}
\label{sec:takeaways}

We perform a large-scale annotation of errors in English news text generated by five sources (four models and ground truth articles).
We present Figures \ref{fig:per-label-wide}, \ref{fig:stacked-bar-overall}, and \ref{fig:model-variants} as summaries of our main results. As a reminder to readers, Grover \cite{zellers2019neuralfakenews} is the same model size and architecture as GPT-2 XL \cite{radford2019language}, but trained in-domain (on news text). As such, our results cover three increasing model sizes (GPT-2 Small, XL, and GPT-3 \cite{brown2020language}), one change in domain (Grover), and ground-truth text (Human). For GPT-3, we also study a variety of decoding configurations (Figure \ref{fig:model-variants}).

The main quantity we measure (on $y$-axes) is \textit{span coverage}, which is the average portion of tokens that ends up covered by annotations of a particular error type. 
Since it is possible that multiple spans nest or overlap, there is no upper bound for this quantity.
(See Figure \ref{fig:error-measuring-comparison} for a comparison of span coverage with other measurement alternatives.)
Figure \ref{fig:per-label-wide} measures span coverage for each type of span separately, Figure \ref{fig:stacked-bar-overall} stacks them, and Figure \ref{fig:model-variants} removes non-error spans (reader issues) before adding them (as in Figure \ref{fig:stacked-bar-overall}, but without showing the individual types).

The following are our key findings.\\



\noindent
\textbf{1. Scaling pays off to improve \nameEncyclopedic, \nameCommonsense, and \nameIncoherent~errors (Fig. \ref{fig:per-label-wide}).}
These error categories \iconDecreasing \textbf{decrease} with in-domain training (Grover) and larger model size (GPT-3). Human text still shows the fewest of these kinds of errors.\\

\noindent
\textbf{2. Scaling benefits plateau for \nameOffPrompt, \nameBadMath, and \nameGrammarUsage~errors (Fig. \ref{fig:per-label-wide}).}
These three error categories see a \iconModelPlateau \textbf{model plateau} in error reduction when scaling to GPT-3. Of these error types, humans still commit fewer \nameOffPrompt~(more: \S\ref{sec:analysis:offPrompt}) and \nameGrammarUsage~errors, but \nameBadMath~appears saturated for our domain.\\

\noindent
\textbf{3. \nameSelfContradiction~and \nameRedundant~errors exhibit more complex scaling behavior (Fig. \ref{fig:per-label-wide}).}
We roughly categorize these trends as \iconUpDown \textbf{rising and falling}:
increasing for medium or large-scale models, but dropping for human-authored text. 
Text generated by GPT-2 Small is so often incoherent that there is little possibility for \nameSelfContradiction~(more: \S\ref{sec:analysis:selfContradiction}), and the increase in \nameRedundant~errors varies based on how errors are counted (more: \S\ref{sec:analysis:redundant}).\\

\noindent
\textbf{4. Human-authored text produces the most reader issues (Figs. \ref{fig:per-label-wide} and \ref{fig:stacked-bar-overall}).}
The \nameNeedsGoogle~and \nameJargon~span categories both have a \iconHumansHighest \textbf{humans highest} trend, and both fall under \textit{reader issues}: problems that are not necessarily \textit{errors}, but that still prevent full comprehension or factual verification of the text (more: \S\ref{sec:analysis:readerIssues}).
    
Furthermore, human-authored text is not free from error annotations (Figure \ref{fig:stacked-bar-overall}). This can serve either as a control for baseline error rates (more: \S\ref{sec:analysis:whatsNext}), or as a mechanism for critiquing human writing.\\

\input{fig/stacked-bar-overall-fig}

\input{fig/model-variants-fig}

\noindent
\textbf{5. Decoding hyperparameters have a huge impact (Figure \ref{fig:model-variants}).}
For the previous findings, we fix the sampling configuration for all models to an apples-to-apples setup for fair comparison: top-$p$ = 0.96, (softmax) temperature = 1, and no frequency penalty (i.e., word repetition penalty; defined precisely in \S\ref{sec:decoding-strategies}, Equation \ref{eq:frequency-penalty}). To study the effects of these decoding settings, we annotate text generated by GPT-3 using a variety of values for top-$p$ and temperature, both with and without a frequency penalty. 
    
To our surprise, the decoding hyperparameters considerably affected error rates (more: \S\ref{sec:analysis:decodingConfigs}). As seen in Figure \ref{fig:model-variants}, the worst sampling procedure for GPT-3 (argmax sampling with no frequency penalty) performed even worse than GPT-2 XL. But the best sampling procedure (surprisingly, also argmax sampling, but with a frequency penalty) produced text with as few apparent \method error spans as those authored by humans (more: \S\ref{sec:analysis:whatsNext}).


All of these findings are discussed in more detail in Appendix~\ref{sec:analysis}. 

%% file: fig/stacked-bar-overall-fig.tex
\begin{figure}[t!]

\begin{center}
\includegraphics[width=0.99\linewidth]{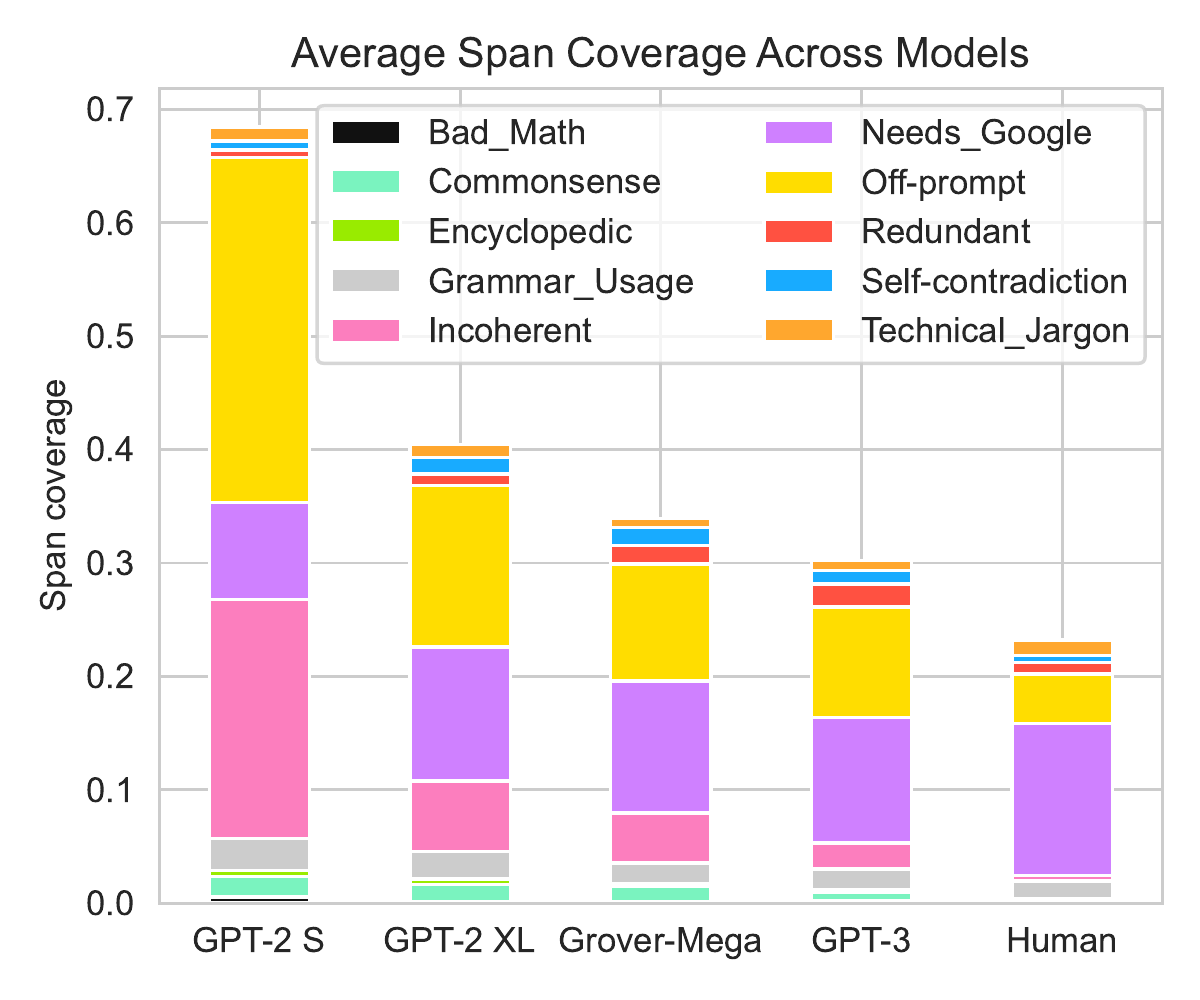}
\end{center}

\caption{
Average portion of tokens covered by span annotations, broken down by error type. All models, including GPT-3, use the same apples-to-apples decoding  hyperparameters: top-$p$=0.96, temperature=1, and no frequency penalty. We scale each span by its token length, normalize by generation token lengths, and remove severity-1 \nameGrammarUsage~errors (see \S\ref{sec:data-quality}).
}
\label{fig:stacked-bar-overall}
\end{figure}

%% file: fig/model-variants-fig.tex
\begin{figure}[t!]

\begin{center}
\includegraphics[width=0.99\linewidth]{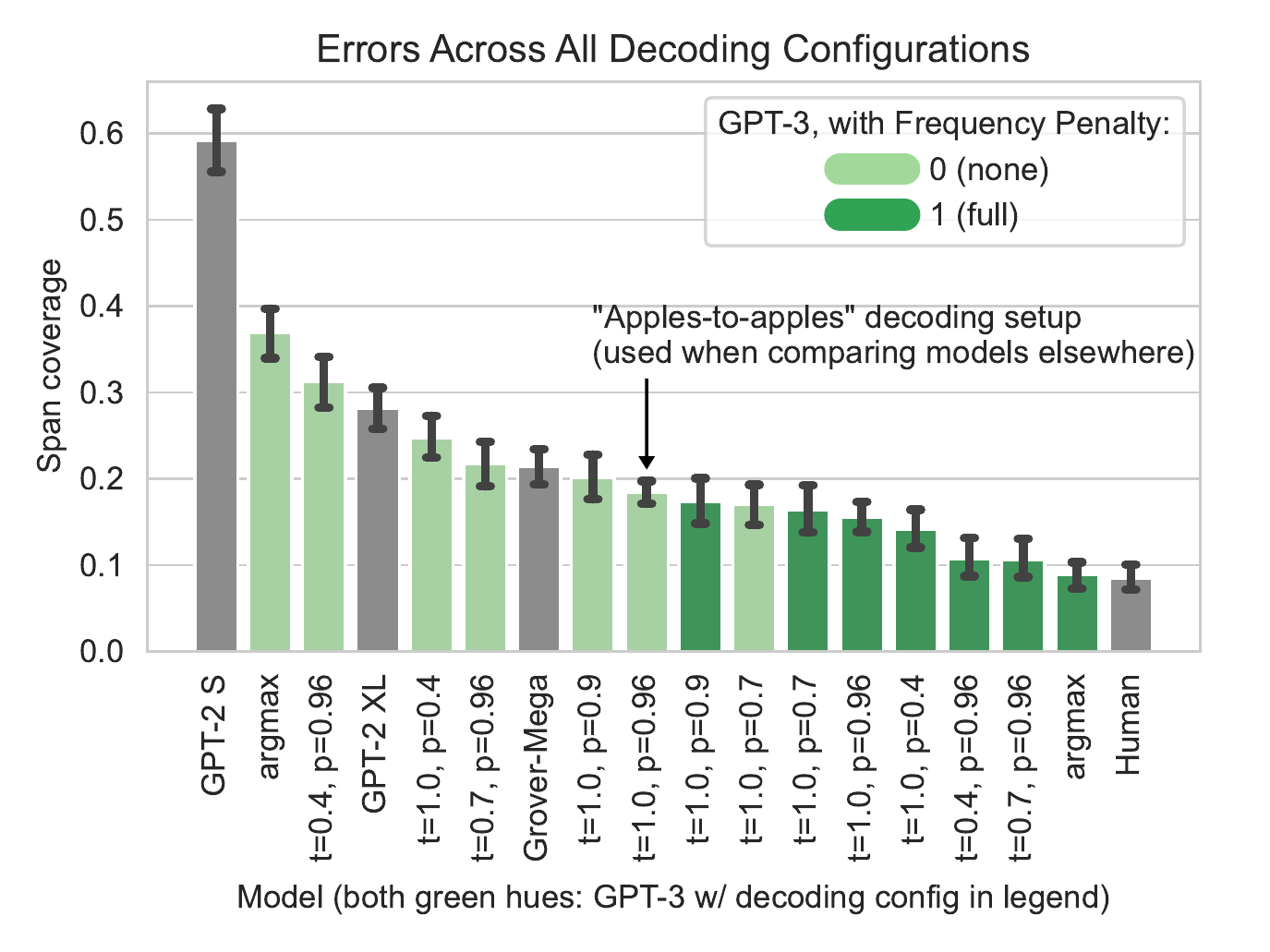}
\end{center}

\caption{
Taking the average span coverage (Figure \ref{fig:stacked-bar-overall}) and removing reader issues (\nameJargon~and \nameNeedsGoogle), we plot values and 95\% confidence intervals for all models, including all decoding hyperparameters we tested for GPT-3. We find a surprisingly large change in annotated errors depending on the decoding setting used.
}
\label{fig:model-variants}
\end{figure}

%% file: src/bg.tex
\section{Evaluation of Natural Language Generation}
\label{sec:oe-nlg}

We make our study in the area of open-ended natural language generation, a loose term for generating longer texts with an increased level of creative freedom. 
The common factor in all open-ended generation tasks such as story, blog, and dialog generation is the wide and diverse nature of target outputs. 
Lexically and even semantically dissimilar responses to the same prompt could be equally valid.
For example, a model prompted with the blog title ``Recipes for success this Holiday season'' could describe how to roast a turkey or strategies for dealing with the stresses of holiday travel.

This allowable variation poses a particular difficulty for the evaluation of generation systems. 
Traditionally, text generation quality for tasks like machine translation or graph-to-text generation has been measured by word overlap with human-authored references \citep{papineni2002bleu,lin2004rouge}. 
Though measures like BLEU allow for multiple references, they break down when the space of allowable outputs is large, as in open-ended generation.
Recently introduced metrics seek to remedy this problem \citep{hashimoto-etal-2019-unifying,pillutla2021mauve}, but the gold standard for evaluating generated text is still human judgment.

However, current approaches to eliciting human judgement of generated text often do not provide detailed insight into where models are making progress, where they are failing, and the scope of these failures. 
A/B-style testing allows for directly comparing one system against others \cite{clark-smith-2021-choose}, but can only express relative improvements. 
Simple Likert scale judgements can assess text quality, but do not explain why a generated text receives a given rating, or which segment of the text is problematic.
Insights into model failures often come instead from a small scale expert analysis of outputs. 
However, these ``error analyses,'' once a staple of NLP research, have become less common in recent years, perhaps due to their small size and high variance. 

A hypothesis of the current work is that a well designed error analysis annotation framework could be used by crowdworkers to annotate large amounts of text, thereby providing detailed information about model progress and failures as well as actionable directions for future research.
Such a framework would be easy to learn, reusable, and independent of particular models or experimental conditions.
In what follows, we outline the details of such a method.

%% file: src/method.tex
\section{\method Annotation Methodology}
\label{sec:method}
This section describes the high-level annotation methodology for \method. 

\input{fig/error-interface-fig}


\subsection{Prompt and Generation}

Our annotations consider two segments of text: a one-sentence prompt, and a one-paragraph generation. The \emph{prompt} is human-written. It provides both starting tokens for model generation, as well as context for humans to evaluate whether a model is able to stay on-prompt---both topically and factually. Annotators know that the prompt is written by a human.

The \emph{generation} is either text sampled from a language model, or the human-authored continuation to the prompt. Annotators, who do not know whether the generation came from a model or humans, assess this text. A paragraph length (80--145 tokens) is chosen to balance expressiveness with scope. For expressiveness, models must be given a sufficient number of tokens to express their capabilities lexically, syntactically, and semantically. One paragraph allows for significantly more variation than a single sentence. On the other hand, assessing multiple paragraphs is challenging, both as a crowdsourcing task itself, and because it broadens the kinds of errors to include larger narrative scope. We leave extensions of \method to longer narrative lengths for future work.

\subsection{Span Labeling}

Annotators select spans that contain problems in the generation. The spans are automatically snapped to word boundaries. We choose spans to balance specificity (i.e., vs. simply commenting on the text as a whole) with ease of use (vs. imposing a more structured annotation schema).

\subsection{Span Selection}

We instruct workers to select the smallest span---minimally a single word---that contains an issue. Sometimes this involves an entire phrase, sentence, or multiple sentences. We aim for specificity because during aggregation, it is possible to ``back off'' annotations to larger spans, but not the inverse.

Once they select a span, workers (1) label the error type, (2) choose a severity level, and (3) explain their reasoning behind the error. Workers use the annotation interface shown in Figure \ref{fig:err-interface} to mark a span with these three steps. We describe each step in greater detail in the next three sections.

\subsection{Error Types}
\label{sec:error-types}

Each selected span is labeled with exactly one error type. Multiple errors may be marked with partially or fully overlapping spans in the case that one text segment contains multiple problems.

We chose ten error types to balance three criteria: linguistic analysis, observed errors in generated text, and capabilities of everyday people with one to two hours of training.\footnote{The complete training material is available for download.} We developed the schema by starting with the first two criteria (linguistic analysis and observed errors), and refining it over several pilot annotation studies, with 30 crowd workers performing 750 total annotations of 60 paragraphs before beginning data collection. 

We broadly group the errors into three categories: \emph{language} errors, \emph{factual} errors, and \emph{reader issues}.  Language errors are issues with internal and external structure of text: which ideas are expressed, and whether they are expressed coherently and consistently. Factual errors denote that the information presented is known to be incorrect. Reader issues, on the other hand, are cases where the text is too technical or obscure to assess its factuality. Hence, reader issues are not errors, per se, but regions where a reader would need assistance outside of the text itself for comprehension. 

We present the ten error types  in Table~\ref{tab:onto-ex} (several pages back).
Appendix~\ref{sec:schema} provides more details, examples, and explanations for all error types.

\subsection{Severity}
\label{sec:severity}

Errors naturally vary in how jarring they are to a reader.
We define three error severity levels, and ask annotators to pick one for each error.

The severity levels are as follows. \textbf{(1)} Almost no impact on quality; just a small problem. \textbf{(2)} Understandable, but difficult; what's written is still comprehensible, but there's clearly an issue. \textbf{(3)} Very difficult to understand; the error almost completely ruins the text.

We provide examples of each severity in Appendix \ref{sec:severity-examples}.
In this paper, we omit an analysis of the severity labels (except for an illustration in Figure \ref{fig:error-measuring-comparison}), but include it in our data release for future work to explore.

\subsection{Explanation}
Finally, we ask annotators to explain their reasoning behind each error in natural language. We provide example explanations during training, but do not impose strict guidelines. This paper primarily focuses on quantitative error analysis, but we anticipate the error explanations may warrant future investigation.

\subsection{Annotation Process}
\label{sec:crowdsourcing-details}

We use Amazon Mechanical Turk (AMT) for all data collection.

\paragraph{Training} We first pay each worker \$40 to take an extensive qualification task, which both trains them in the span categorization scheme and quizzes their understanding. 
We pass workers if they score $\geq$ 90 points out of 100 points (details in Appendix~\ref{sec:grading-details}).

\paragraph{Annotation} Workers annotate each paragraph using a custom annotation interface (shown partially in Figure \ref{fig:err-interface}), for which we pay \$3.50. We calculated \$3.50 per annotation by aiming to pay workers at least \$15/hour. After several annotation rounds, we observed considerable variation in time per annotation,\footnote{Median: 212s, mean: 265s, std. dev.: 199s.} so this cost should not be necessarily seen as a requirement for \method annotations.


%% file: fig/error-interface-fig.tex
\begin{figure}[t]

\begin{center}
\includegraphics[width=0.99\linewidth]{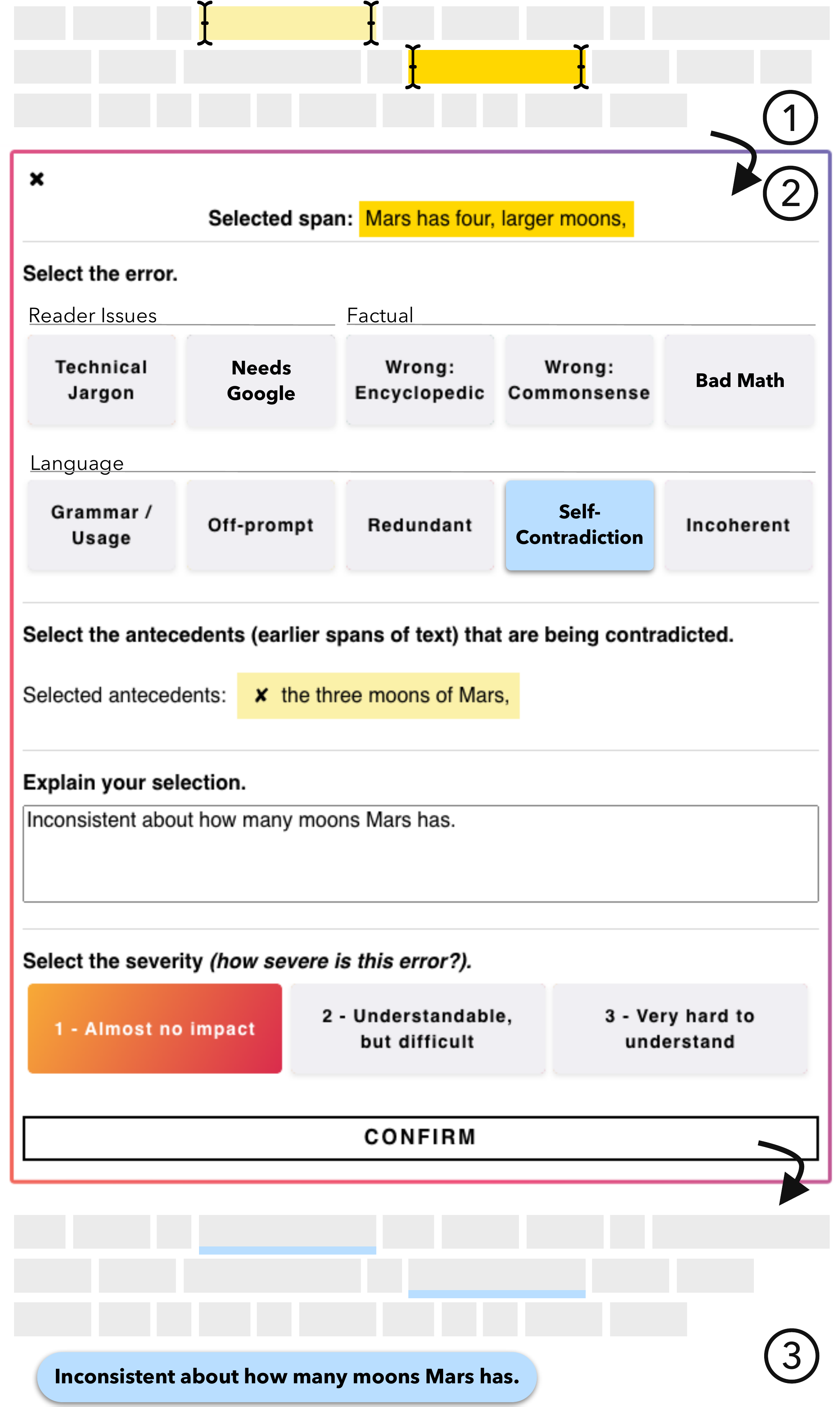}
\end{center}

\caption{
\method interface for annotating a single span: (1) highlighting a span  (and later, an antecedent); (2) completing the annotation, with the error type, explanation, and severity; (3) the error annotation is saved---interactive controls allow detailed viewing and editing of spans (not shown).
}
\label{fig:err-interface}
\end{figure}

%% file: src/data.tex
\section{Data Collection}
\label{sec:data-collection}

We collect 13k human annotations of 1.3k paragraphs using \method, resulting in over 41k spans.

\subsection{Models}

We consider four model configurations to test recent state-of-the-art transformer-based \cite{vaswani2017attention} models.

\paragraph{GPT-2 Small} \cite{radford2019language} The 117M parameter variant of GPT-2, which is pretrained on WebText, without additional fine-tuning.

\paragraph{GPT-2 XL} \cite{radford2019language} The 1.5B parameter variant of GPT-2, (WebText, no fine-tuning).

\paragraph{Grover-Mega} \cite{zellers2019neuralfakenews} The 1.5B parameter variant of Grover, a model with the same architecture and parameter count of GPT-2, trained on news articles and their metadata.

\paragraph{GPT-3 DaVinci} \cite{brown2020language} The 175B parameter variant of GPT-3, which is trained on a version of the Common Crawl web scrape with additional filtering and deduplicating.


In addition, we also use the actual human-written text from the data sources we draw from, which we denote as \textbf{Human}.

\subsection{Decoding strategies}
\label{sec:decoding-strategies}

We consider three main hyperparameters when sampling from models: $p$ for \textit{top-p} or \textit{nucleus sampling} \cite{holtzman2019curious}, an alternative to \textit{top-k};\footnote{We omit separate studies of \textit{top-k}, due to results presented by \newcite{holtzman2019curious}, and OpenAI's removal of \textit{top-k} from the GPT-3 API.} \textit{t} for the \textit{softmax temperature}; and \textit{f.p.} for \textit{frequency penalty}. The frequency penalty scales a token's likelihood based on how many times it was already generated by applying the following modification to the model's output:

\begin{equation}
    \ell_i(t) \leftarrow \ell_i(t) - c_{<i}(t) \cdot \alpha_f
    \label{eq:frequency-penalty}
\end{equation}

\noindent
where $\ell_i(t)$ is the model's output for token $t$ at the $i$-th position,\footnote{While $\ell_i(t)$ is defined to be \textit{``logits (un-normalized log-probabilities),''} because it is un-normalized, we anticipate that it is simply the model's output before the $\log(\text{softmax}(\cdot))$ is applied. See OpenAI's description of frequency and presence penalties: \url{https://beta.openai.com/docs/api-reference/parameter-details}} $c_{<i}(t)$ is the count of token $t$'s sampled occurrences prior to the $i$-th position, and $\alpha_f$ is the frequency penalty.
We omit studying \textit{presence penalty}, another hyperparameter offered for GPT-3, simply due to annotation budget constraints.

To compare models as consistently as possible, we set identical decoding strategies for our primary data collection. We refer to this as the ``apples-to-apples'' decoding setup throughout the paper:

\begin{align*}
    p = 0.96 \qquad t = 1.0 \qquad \textit{f.p.} = 0
\end{align*}

However, we also wish to study the effects of these decoding strategies. We annotate generations from the strongest available model (currently, GPT-3) varying the following parameters:

\begin{align*}
    p &\in \{0.4, 0.7, 0.9, 0.96\}\\
    t &\in \{0.0 \textit{ (argmax)}, 0.4, 0.7, 1.0\}\\
    \text{f.p.} &\in \{0 \text{ (none)}, 1 \text{ (full)}\}
\end{align*}


For budget reasons, we only vary $p$ and $t$ independently---i.e., we set $p=0.96$ when varying $t$, and $t=1.0$ when varying $p$.

\subsection{Prompt Selection}
\label{sec:data:prompt}
We use news articles as the sources of prompts for models to condition on for generation.
Specifically, we use news articles found in the Common Crawl.
We select the first sentence as the prompt.

Our use of news text is constrained by two factors. First GPT-3 is trained on the Common Crawl, from 2016 through 2019. We wish to avoid testing GPT-3 by generating from articles it saw during training, due to the possibility of copying \cite{Carlini2021ExtractingTD}.
Second, news articles began heavily covering the COVID-19 pandemic beginning around February 2020. Though testing models' capabilities to generate text about unseen events is a valuable line of study, the distribution shift caused by COVID-19 in news writing about all aspects of life is difficult to overstate.

As such, to make the comparison more amenable to models' training data, we consider news articles from January 2020. We select articles where there is a known topic---such as \textit{Food} or \textit{Sports}---from the Common Crawl metadata, to allow for studying any effect of coarse-grained subject.
\subsection{Generation}

We generate between 80 and 145 tokens\footnote{Counted by Stanza tokenization \cite{qi-etal-2020-stanza}, not byte-pair encoding (BPE) or whitespace-separated tokens.} from each model as a continuation to the first sentence of the news article. We stop generating when we heuristically detect the first sentence boundary after 80 tokens. If the model does not end a sentence between 80 and 145 tokens, we sample again. For the \textit{Human} setting, we use the remainder of the article, similarly stopping after the first sentence boundary after 80 tokens.

\subsection{Annotation}
\label{sec:data:annotation}

\paragraph{Crowdsourcing} Workers first complete training and qualification tasks. We provide more details in \ref{sec:crowdsourcing-details}. From pilot studies, we discovered that each error, depending on its severity and clarity, has only a low to moderate chance of being identified by each worker. However, most worker-identified errors were truly problems. In other words, annotators labeled issues with high precision and low recall. To account for this, we have \textbf{10 workers} annotate each paragraph. We examine the agreement and variability of annotations in Appendix~\ref{sec:data-quality}.

\paragraph{Dataset statistics} We provide detailed dataset statistics in Appendix~\ref{sec:data-statistics}.

%% file: src/prediction.tex
\section{Error Prediction}
\label{sec:prediction}

A natural question is: using this data, can machines learn to detect and classify errors in machine generated text?

\paragraph{Task} We frame this problem as a span classification task. 
Given a span from a generated text, the goal is to classify its error type or output “No Error” if there is none.
Positive examples for each error class are taken from our data.
We sample random spans that were not labeled with any error type as negative examples. 
To ensure a breadth of  span lengths, we sample 3 negative spans for every length of error span in the generated text. 
We split the generated texts into train, development, and test sets using 1063 texts (28029 error spans), 100 texts (2538 spans) and 100 texts (2677 spans) respectively.

\paragraph{Model} We use a standard span classification model inspired by \citet{wadden-etal-2019-entity}.
This model encodes every generated text using a pretrained language model (RoBERTa-large). 
Spans are represented with the final layer of this encoding. 
Following previous work, we concatenate the start and end tokens with a task-specific learned length embedding. 
The resulting vector is passed through a feedforward network which reduces its dimensionally to the number of error categories plus a ``No Error’’ option. 
The resulting model has 357M trainable parameters. 
The model is trained to minimize the cross entropy of the correct span category. 
We train for 15 epochs using AdamW with a learning rate of $10^{-6}$. 
We validate after each epoch and use the checkpoint with the lowest validation loss (epoch 8). 

\paragraph{Evaluation} To evaluate the error prediction model, we use per-token precision, recall, and F$_1$ score per error category. 
We classify every span up to length 30 in a generated text. 
We take as gold labels the aggregated human error spans collected in our data. In other words, models predict the combined spans of all 10 annotators. 
For comparison, we also report as \textit{Human} the average metrics of one annotator versus the others (i.e., 1-vs-9).\footnote{The difference in available references (10 for models, 9 for humans) mean this setup makes it easier for models to score higher in precision, and for humans to score higher in recall. Despite this, humans still achieve higher precision, and models still achieve higher recall.}

\paragraph{Results}
Table~\ref{tab:preds} shows the error prediction capability of this model in terms of precision and recall. 
As we noted earlier, a single human annotator can be thought of as a high precision, low recall judge. 
These results bear out this claim.
For all but one category, humans have higher precision annotations.
However, the models trained on the aggregation of human labels can achieve considerably higher recall. 
For half of the error categories, this leads to higher model F$_1$ scores than the human annotators.

We see that the model is successful at identifying information that human's would have to manually verify (\nameNeedsGoogle), achieving nearly perfect recall with precision close to 0.6. 
The model can also identify \nameGrammarUsage, \nameIncoherent, and \nameRedundant~errors with higher recall than an individual human annotator, though at the cost of precision (sometimes in the .20s).


\begin{table}[t]
    \centering
    \resizebox{\linewidth}{!}{%
    \begin{tabular}{lccc|ccc}
    \toprule
    \multirow{2}{*}{\bf Error} & \multicolumn{3}{c}{ Model }& \multicolumn{3}{c}{ Human } \\ 
    & P & R & F$_1$ &P &R &F$_1$\\
    \midrule
    \nameBadMath & -- & 0 & -- & 0.72 & 0.14 & {\bf 0.24} \\
    \nameCommonsense & 0.77 & 0.06 & {\bf 0.10} & 0.17 & 0.02 & 0.04 \\
    \nameEncyclopedic & -- & 0 & -- & 0.22 & 0.03 & {\bf 0.05} \\
    \nameGrammarUsage & 0.29 & 0.23 & {\bf 0.26} & 0.30 & 0.04 &  0.08 \\
    \nameIncoherent & 0.59 & 0.34 & {\bf 0.43} &0.69 & 0.15 &  0.24 \\
    \nameOffPrompt & 0.67 & 0.29 & 0.41 & 0.88 & 0.31 & {\bf 0.46} \\
    \nameRedundant & 0.23 & 0.82 & 0.36 & 0.88 & 0.35 & {\bf 0.50} \\
    \nameSelfContradiction & 0.08 & 0.23 & 0.12 & 0.51 & 0.09 & {\bf 0.16} \\ \midrule
    \nameJargon & 0.18 & 0.74 & {\bf 0.29} & 0.61 & 0.12 & 0.20 \\
    \nameNeedsGoogle & 0.59 & 0.96 & {\bf 0.73} & 0.78 & 0.20 & 0.32 \\
    \bottomrule
    \end{tabular}
    }
    \caption{
    Model prediction results against combined spans of 10 annotators, compared with humans scored as one-vs-rest (i.e., 1-vs-9). Bold F$_1$ scores denote the higher average; values marked ``--'' cannot be computed due to division by zero. \textbf{Takeaway:} Humans have higher precision in every error type except \nameCommonsense, but relatively sparse annotations lead to lower computed recall. This allows the model to achieve higher F$_1$ scores for half of the span categories.
    }
    \label{tab:preds}
\end{table}

%% file: src/related_work.tex
\section{Related Work}

Automated evaluation metrics such as BLEU \cite{papineni2002bleu}, ROUGE \cite{lin2004rouge}, METEOR \cite{banerjee2005meteor}, and BERTScore \cite{zhang2019bertscore} compute a generation's score based on a (set of) reference(s).
Their use is well-established in tasks like machine translation and summarization, but they are less helpful in open-ended text generation, where there is a vast diversity of possible high-quality continuations.

Recent studies propose automated metrics for open-ended text generation evaluation such as: Perception Score \cite{Gu2021PerceptionSA}, which diffuses evaluation onto a multidimensional space and assigns a single score; UNION \cite{Guan2020UNIONAU}, which learns to distinguish human-written stories from negative samples by generating perturbations of human-written stories; and MAUVE \cite{pillutla2021mauve}, which compares the distribution of machine-generated text to that of human language.

An alternate recent approach to assessing open-ended text generation was presented in TuringAdvice \cite{zellers-etal-2021-turingadvice}, where crowd workers assess machine-generated advice in response to Reddit posts. In their error analysis, \citeauthor{zellers-etal-2021-turingadvice} connect problems in generated text to core NLP tasks, such as \nameSelfContradiction~errors as instances of failed natural language inference \cite{monz2001light}, or \nameOffPrompt~errors as cases of failed reading comprehension \cite{richardson2013mctest}. While past work has attempted to guide text generation using discriminative models trained for such tasks \cite{holtzman-etal-2018-learning}, it remains an open challenge.

Comparative human evaluations of natural language generations ask annotators to rank system outputs relative to each other. Text is typically evaluated using a few global criteria, such as fluency and relevance, using discrete (e.g., 5-point) \cite{sai2020survey} or continuous scales \cite{Novikova2018RankMERH}.
Recent work even automates this approach, running a human evaluation alongside automatic metrics on leaderboard submissions \cite{khashabi2021genie}.
In the RoFT system \citep{dugan2020roft}, annotators attempt to detect the boundary between human- and machine-written text as a proxy for assessing quality. Table~\ref{tab:related-work-comparison} summarizes the differences between these schemes and \method. See \newcite{celikyilmaz2021evaluation} for a recent survey of text generation evaluation techniques across both human and automatic metrics.

\begin{table}[t]
\centering
\resizebox{\linewidth}{!}{
\begin{tabular}{@{}lccccccc@{}}
\toprule
\textbf{Method} & \multicolumn{1}{|c}{\textbf{GC}} & \multicolumn{1}{c|}{\textbf{SET}}  & \textbf{DE} & \multicolumn{1}{c|}{\textbf{RR}} & \textbf{EE} & \multicolumn{1}{c|}{\textbf{RS}} & \textbf{SA} \\ \midrule
Likert-Scale & \checkmark  &    &  \checkmark  &   & & \checkmark & \\
RankME & \checkmark &    &    &  \checkmark &   & \checkmark & \\
RoFT &  \checkmark  &    &  \checkmark & & \checkmark  &  & \\
\method &  &    \checkmark   &  \checkmark   &   &  \checkmark &  & \checkmark\\ \bottomrule
\end{tabular}
}
\caption{
Comparison of different natural language generation human evaluations. Here, \textbf{GC} : General Criteria, \textbf{SET} : Specific Error Type, \textbf{DE} : Direct Evaluation, \textbf{RR} : Relative Ranking, \textbf{EE} : Error Explanation, \textbf{RS} : Rating Scale, \textbf{SA} : Span Annotation.
}
\label{tab:related-work-comparison}
\end{table}

While these approaches may be helpful---sometimes \cite{card.2020}---at ranking systems, they do not give us insight into exactly \textit{which} parts of a generation fall short, and \textit{why}.
One approach related to or annotation method is pursued by \newcite{wood2018rethinking}, who develop a collaborative mobile app where users draw ``graffiti'' commentary on news articles.
\method aims to assess model generations the way we would critique human-written text: by locating, coarsely categorizing, and explaining problems.

%% file: src/conclusion.tex
\section{Conclusion}
\label{sec:conclusion}

We present \method, a method for identifying and explaining issues in generated text.
Along with the annotation framework, we present an analysis of the \method method applied to several large neural language models in an open-ended news generation task.
We release our data and methodology to the community.

%% file: src/schema.tex
\section{\method Annotation Schema}
\label{sec:schema}

Here, we present in greater detail the \method annotation error types.\footnote{All example annotations here are our own. Many are provided to annotators during training.}
A visual summary is shown in Figure~\ref{fig:ontology-condensed}.

While we annotate using this schema, the essence of our study is to embrace language users' abilities to detect when something may be wrong with text.
In other words, we do not wish for our span definitions to get in the way of humans describing problems with text.
To this end, we encourage researchers to embrace label back off (to coarser categories), merging labels (based on empirical observations), and refining the annotation ontology over time.
The central goal is to collect \emph{what people find wrong with text.}

\subsection{Language Errors}

We define five categories of \emph{language errors}, which concern the selection of ideas in a text and how they are expressed. These range from grammar and syntax problems to issues of semantics and pragmatics.

\subsubsection{\nameGrammarUsage}

This category of errors includes missing words, extra words, and incorrect or out of order words. 

\begin{small}
\begin{quote}
    \textsc{\textbf{example}}\\
    A PhD student from the University of Kent in the UK claims to have discovered a clever way to explain the positive \errGrammarUsage[emoticons] in cats.\vspace{0.5em}\\
    \textit{\textbf{Explanation:} The word should probably be ``emotions.''}
\end{quote}
\end{small}

We also label \nameGrammarUsage for inserted words or small phrases that could be deleted to resolve the issue:

\begin{small}
\begin{quote}
    A couple is facing criticism for their extravagant birthday party. The bewitching pair had first stripped down to fishnets \errGrammarUsage[and backward.]
    \vspace{0.5em}\\
    \textit{\textbf{Explanation:} This phrase can simply be deleted.}
\end{quote}    
\end{small}

We avoid partitioning \nameGrammarUsage~errors into more detailed categories based on the observation that large language models produce fewer issues of syntax and diction (aside from \nameRedundant~errors, described next). As such, we focus instead on semantic and pragmatic errors, captured by the upcoming error types.

\input{fig/ontology-condensed-fig}

\subsubsection{\nameRedundant}
While ``redundant'' can also include extra unnecessary information, we specifically use the \nameRedundant~label to mark repetition.
In identifying redundant text, our schema annotates both the \nameRedundantAntecedent~(first mention) and the \errRedundant[redundant text] (when the repetition occurs).
Sometimes the exact word or phrase will be repeated.

\begin{small}
\begin{quote}
    \textsc{\textbf{example}}\\
    Many merchants worry about the possibility of \errRedundantAntecedent[poor service] \errRedundant[or service] for certain categories of customers.
\end{quote}
\end{small}

Other times, generated text expresses the same idea repeatedly using different words.

\begin{small}
\begin{quote}
    \textsc{\textbf{example}}\\
    They then made decisions based on Kondo’s instructions, to the extent that they \errRedundantAntecedent[created de-cluttered spaces] \errRedundant[and got rid of clutter and clutter-filled spaces].
\end{quote}
\end{small}

\subsubsection{\nameOffPrompt}

The prompt is a human-written sentence used as context from which the model generates a continuation. Models sometimes generate text that is unrelated to the prompt.

\begin{small}
\begin{quote}
    \textsc{\textbf{example}}\\
    \textbf{Prompt:} Dogs are the new kids.\\
    \textbf{Generation:} Statistics suggest that most Americans would be happier with dogs than children. \errOffPrompt[In fact, four out of five don't even visit the dentist annually, much less every six months.] Dog owners report much higher rates of happiness than non-dog owners.
\end{quote}
\end{small}

Other times, the text may be related, but it contradicts what is stated in the prompt.

\begin{small}
\begin{quote}
    \textsc{\textbf{example}}\\
    \textbf{Prompt:} China sets new record for Economic Growth\\
    \textbf{Generation:} The Chinese economy \errOffPrompt[fell 10\% this month, the third such loss this year.]
\end{quote}
\end{small}

\subsubsection{\nameSelfContradiction}

When a model generates text that contradicts the prompt, that is labeled as \nameOffPrompt. But when a model generates text that contradicts \emph{itself}, that is labeled as \nameSelfContradiction. We also mark the \nameSelfContradictionAntecedent (original statement).

\begin{small}
\begin{quote}
    \textsc{\textbf{example}}\\
    McDonald's is considering a design which will replace the \errSelfContradictionAntecedent[cardboard packaging.] Mr Gore-Cotter said: ``We recognise the concern around waste. We are now looking at a new design that minimises the \errSelfContradiction[plastic bag.'']\vspace{0.5em}\\
    \textit{\textbf{Explanation:} The idea of minimizing the plastic bag contradicts the stated goal of replacing cardboard packaging.}
\end{quote}
\end{small}

\begin{small}
\begin{quote}
    \textsc{\textbf{example}}\\
    Mall of America plans to
    \errSelfContradictionAntecedent[lay off and furlough hundreds of its employees.]
    \errSelfContradiction[It has no plans to restrict the number of hours workers can work.]\vspace{0.5em}\\
    \textit{\textbf{Explanation:} Furloughed workers are explicitly restricted from working.}
\end{quote}
\end{small}

\subsubsection{\nameIncoherent}

Generated text is sometimes grammatical, not redundant, on prompt, and not contradictory, but still confusing. We provide the \nameIncoherent label for such sentences.

\begin{small}
\begin{quote}
    \textsc{\textbf{example}}\\
    Melody Mitsugi, 28, had never given her kids cheese toast before her husband
    \errIncoherent[drew a map of it on her toast.]
    \vspace{0.5em}\\
    \textit{\textbf{Explanation:} One can’t exactly draw a map of Cheese Toast, and one probably wouldn't draw it on toast itself.}
\end{quote}
\end{small}

\begin{small}
\begin{quote}
    \textsc{\textbf{example}}\\
    Cats naturally show anxiety and fear by at times
    \errIncoherent[breaking apart different parts of the brain in an attempt to keep the others from escaping.]
    \vspace{0.5em}\\
    \textit{\textbf{Explanation:} It's difficult to even imagine what is happening in this passage.}
\end{quote}
\end{small}

\subsection{Factual Errors}

We define three categories of factual errors, which encompass known incorrect statements.

\subsubsection{\nameBadMath}

Generated text will sometimes have issues with basic mathematical operations of known quantities (e.g., \textit{``half of ten apples is four''}), problems converting fixed units (e.g., \textit{m to cm}).

\begin{small}
\begin{quote}
    \textsc{\textbf{example}}\\
    One account, @Iain\_Rowling1, had over 500,000 followers at one point, but in just four days they fell by \errBadMath[around half - some 4,000.]
\end{quote}
\end{small}

We also include problems converting currencies that are wildly implausible under modern assumptions (e.g., \textit{\pounds1 = \$18 US }).

\begin{small}
\begin{quote}
    \textsc{\textbf{example}}\\
    ... compared with just over \pounds1,000 \errBadMath[(\$18,868)] for previous versions of Samsung’s flagship phone.
\end{quote}
\end{small}

\subsubsection{\nameCommonsense}

These errors mark spans that violate our everyday basic understanding of the world. Though it is challenging to precisely define \textit{commonsense knowledge} \cite{liu2004conceptnet}, we include non-encyclopedic knowledge and basic reasoning.

The following example concerns broadly sensible numerical ranges.

\begin{small}
\begin{quote}
    \textsc{\textbf{example}}\\
    The picture is from high above the South Pole, where close to \errCommonsense[100,000] Astronauts live and work.
    \vspace{0.5em}\\
    \textit{\textbf{Explanation:} Even if we don't know the exact number of astronauts in space, it is common knowledge that 100k is far too many.} 
\end{quote}
\end{small}

The next example involves world knowledge, akin to scripts \cite{schank1977scripts}.

\begin{small}
\begin{quote}
    \textsc{\textbf{example}}\\
    You can get the dress custom-made and stitched at your favorite
    \errCommonsense[spa.]
    \vspace{0.5em}\\
    \textit{\textbf{Explanation:} Spas don't offer stitching.} 
\end{quote}
\end{small}

The following example involves lexical entailment.

\begin{small}
\begin{quote}
    \textsc{\textbf{example}}\\
    The thinness of our bodies isn't an answer to all common human health problems like
    \errCommonsense[obesity] or diabetes
    \vspace{0.5em}\\
    \textit{\textbf{Explanation:} While most of the statement is acceptable, it’s impossible to be ``thin'' and ``obese'' at the same time. } 
\end{quote}
\end{small}

The final example involves time.

\begin{small}
\begin{quote}
    \textsc{\textbf{example}}\\
    Now in 2021, NASA is measuring California wildfire temperatures using an instrument on the International Space Station. This year's record-shattering heat has had global repercussions in \errCommonsense[2017], forcing sea level rise on California and increasing the risk of deadly wildfires.
    \vspace{0.5em}\\
    \textit{\textbf{Explanation:} Events in 2021 can't affect events in 2017. } 
\end{quote}
\end{small}

\subsubsection{\nameEncyclopedic}

These errors are ones that we \textit{know} are factually wrong, and that we could look up in, say, Wikipedia. 

\begin{small}
\begin{quote}
    \textsc{\textbf{example}}\\
    \errEncyclopedic[Japanese Prime Minister Justin Trudeau] said he will be halting all imports and exports until the current situation can be contained.
    \vspace{0.5em}\\
    \textit{\textbf{Explanation:} Justin Trudeau is the Prime Minister of Canada, not Japan.}
\end{quote}    
\end{small}
    
The distinction between \nameEncyclopedic~errors, and the upcoming \nameJargon~and \nameNeedsGoogle~issues, depends on the reader's knowledge.

\begin{small}
\begin{quote}
    \textsc{\textbf{example}}\\
    The gas contains something known as \errEncyclopedic[phyto-romatic acid, a common chemical element in the periodic table.]
    \vspace{0.5em}\\
    \textit{\textbf{Explanation:} Acids aren't elements.}
\end{quote}
\end{small}

\subsection{Reader Issues}

We define two categories of reader issues. These are words or statements a reader cannot verify without using an external resource.

\subsubsection{\nameJargon}

Sometimes generated text includes specific words from a field that requires expertise to understand.

\begin{small}
\begin{quote}
    \textsc{\textbf{example}}\\
    In Chile, an 800-megawatt \errJargon[photovoltaic] plant was built for a record low cost of \$129 per megawatt-hour last year.
\end{quote}
\end{small}

Which words are jargon depends on the reader's particular expertise. This means \nameJargon~spans are more accurately thought of as \emph{potential issues} rather than known errors.

\begin{small}
\begin{quote}
    \textsc{\textbf{example}}\\
    He uses a spirit \errJargon[mash] made from white corn and malted barley and a \errJargon[neutral grain], which he describes as a "whiskey grain.”
\end{quote}
\end{small}

\input{fig/4_examples_fig}

\subsubsection{\nameNeedsGoogle}

Many facts---especially those involving specific people, events, dates, or numbers---could be categorized as encyclopedic knowledge. However, whether the fact is accurate may require additional verification by the everyday reader. To make this distinction between \textit{known} encyclopedic knowledge and \textit{trivia}, we introduce this label to denote that a reader would need to search online to verify whether it is true. 

We instruct annotators to \emph{not} look up facts marked with the \nameNeedsGoogle~span. We do this to keep the focus of the task on classification, rather than factuality detection. As a result, \nameNeedsGoogle~spans mark \textit{statements that would need to be verified}, rather than known errors.

\begin{small}
\begin{quote}
    \textsc{\textbf{example}}\\
    It was promoted by \errNeedsGoogle[Dr. Michael Fanning, the Executive Director of the Foundation for Mental Health Awareness, Inc.]
\vspace{0.5em}\\
    \textit{\textbf{Explanation:} A reader would likely need to look up whether there is a Dr. Fanning who holds this position.}
\end{quote}
\end{small}

\begin{small}
\begin{quote}
    \textsc{\textbf{example}}\\
    ... an \errNeedsGoogle[800-megawatt photovoltaic plant] was built for a \errNeedsGoogle[record low cost of \$129 per megawatt-hour] last year. 
\vspace{0.5em}\\
    \textit{\textbf{Explanation:} In addition to potential \nameJargon~spans, there are at least two \nameNeedsGoogle~spans: 1. whether such a plant can be roughly 800-megawatt, 2. whether \$129/megawatt-hour is a sensible cost measure, and the value is reasonable.}
\end{quote}
\end{small}

To illustrate the annotation methodology and schema in practice, we present four complete example annotations in Figure 
\ref{fig:4_examples}. This figure also illustrates how much variation we see across models.

%% file: fig/ontology-condensed-fig.tex
\begin{figure}[t]

\includegraphics[width=0.99\linewidth]{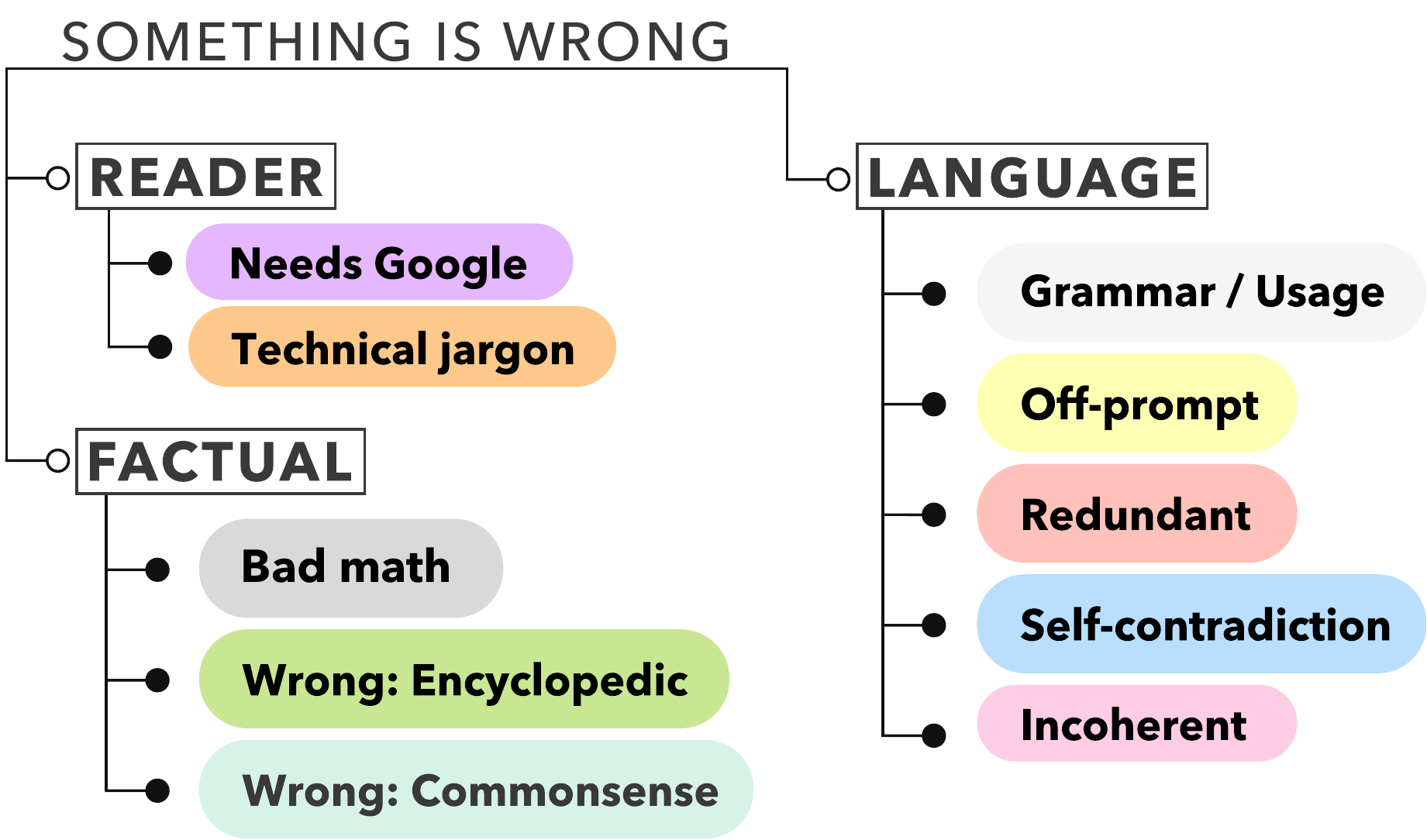}

\caption{
A visualization of \method spans: three categories (reader, language, and factual) composed of ten types. Annotators choose directly from the ten error types.
}
\label{fig:ontology-condensed}
\end{figure}

%% file: fig/4_examples_fig.tex
\begin{figure*}[th!]

\begin{center}
\includegraphics[width=0.99\linewidth]{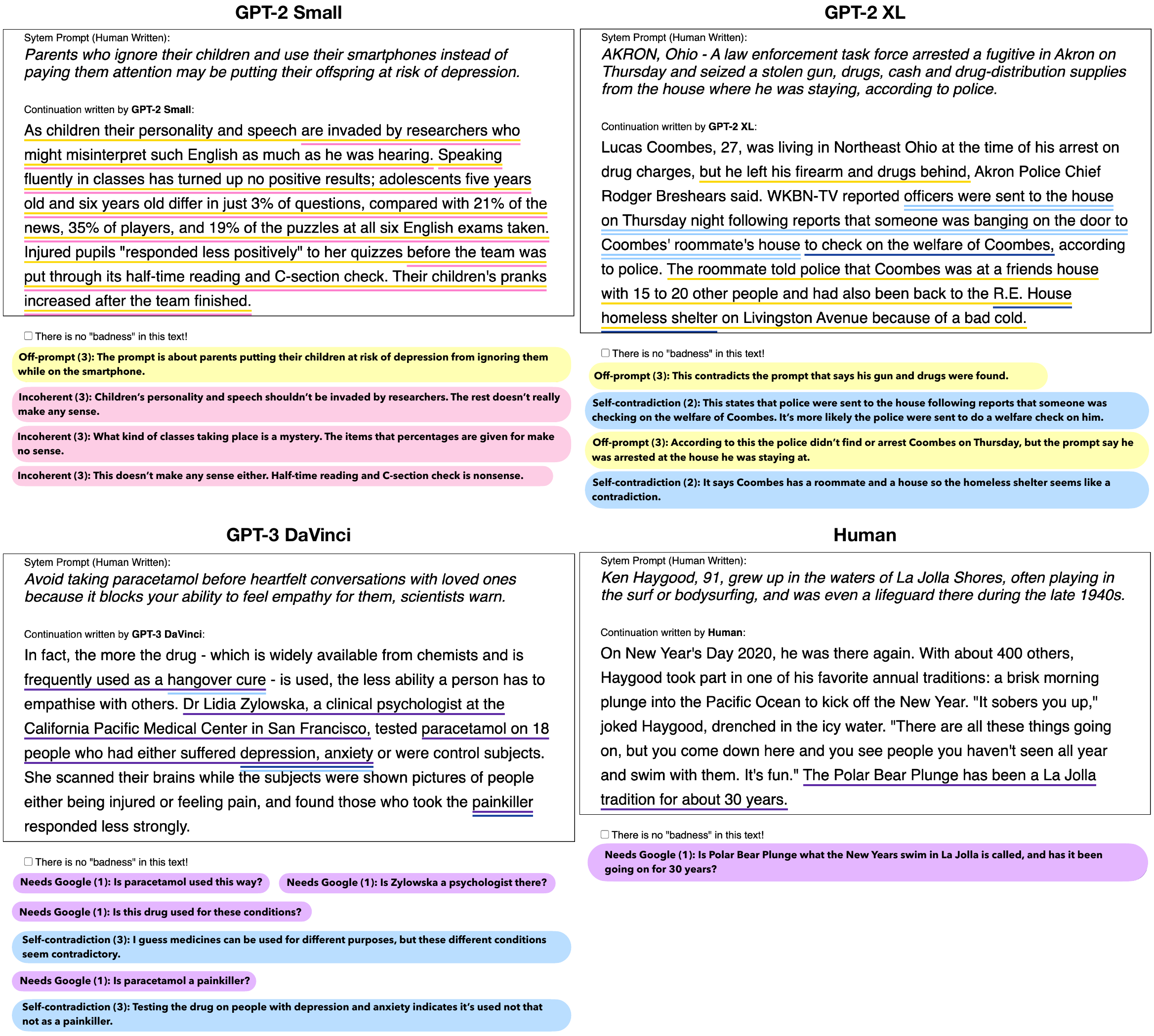}
\end{center}

\caption{
Example \method annotations (for a single annotator) of three model generations and one ground truth continuation, demonstrating the shift in number, type, and severity of errors. The entirety of the \textbf{GPT-2 Small} generation is \nameOffPrompt~and/or \nameIncoherent, with high severity (3/3). \textbf{GPT-2 XL} is instead only about two-thirds covered by errors---still sometimes \nameOffPrompt, but also \nameSelfContradiction, and with high severity (2--3/3). In contrast, \textbf{GPT-3 DaVinci} receives several \nameNeedsGoogle~marks---less severe than errors, as they only indicate that fact-checking is needed---though it also commits two high-severity \nameSelfContradiction~errors by generating inconsistent claims. The \textbf{Human} (ground-truth) continuation only receives one \nameNeedsGoogle~span.
}
\label{fig:4_examples}
\end{figure*}

%% file: src/annotation-more.tex
\section{Annotation Details}
\label{sec:annotation-more}

\subsection{Error Severity}
\label{sec:severity-examples}

We provide here examples for each of the three error severity levels, which we also give to annotators during training.

\begin{small}

\begin{quote}
    \textsc{\textbf{example}}\\
    Paul Campbell-Hughes, from the University of Aberdeen, explains how \errGrammarUsage[she] managed to locate colonies of honey bees in Kent.
    \vspace{0.5em}\\
    \textit{\textbf{Severity: 1.} Since Paul is usually a male name, the model should have used ``he.'' But this error is pretty minor.}
\end{quote}

\begin{quote}
    \textsc{\textbf{example}}\\
    Paul Campbell-Smith, a PhD student from the University of Kent in the UK, claims to have discovered a clever way to explain the positive \errGrammarUsage[emoticons] in cats.
    \vspace{0.5em}\\
    \textit{\textbf{Severity: 2.} The word should probably be ``emotions.'' We can guess what was being said, but it's definitely wrong.
}
\end{quote}

\begin{quote}
    \textsc{\textbf{example}}\\
    \textbf{Prompt:} Whether you're on Facebook, Instagram, Snapchat or TikTok, many people make huge efforts to curate the best version of themselves online.\\
    \textbf{Generation:} \errOffPrompt[This year we've got something for you: a Love Match Custom Size Poster featuring Mather, Phoenix, Kashun and all her friends, divided among six different covers, creating a beautiful custom size poster for your own personal high school reunion.]
    \vspace{0.5em}\\
    \textit{\textbf{Severity: 3.} Even ignoring the end of the generation (a poster for a personal high school reunion?), this whole generation is way off the prompt and does not make sense.
}
\end{quote}

\end{small}

\subsection{Grading Details}
\label{sec:grading-details}
In the training material, there are 10 annotation exercises, 10 multiple choice questions, and 1 real task question to test workers' understanding.

\paragraph{Annotation Exercise}
After going through each error type, there is an annotation exercise. Workers are asked to mark the span with that particular error in a short text. Each exercise is worth 5 points.

\paragraph{Multiple Choice Question}
After going through all \emph{language} errors, and going through all \emph{factual} errors and \emph{reader issues}, there is a \emph{language} error label quiz and a \emph{reader} and \emph{factual} error label quiz respectively. Each label quiz consists of 5 multiple choice questions, where workers are asked to choose the error type of a marked span in a short text. Each multiple choice question is worth 3 points.

\paragraph{Real Task Question}
At the end of the whole training material, workers are asked to apply what they learn in an actual task where they annotate a given paragraph with full tool like ones shown in Figure~\ref{fig:4_examples}. This question is worth 20 points. We mark 7 error spans as the solution. As long as they can mark 5 of 7 error spans, they get a full 20 points. Otherwise, 4 points will be deducted for each missing error span.

In total, there are 100 points. We pass workers if they score $\geq$ 90 points, and then they are provided with the solution to review.

%% file: src/verification.tex
\section{Data Quality}
\label{sec:data-quality}

\input{fig/spans-viz-fig}

Identifying and classifying errors in potentially noisy machine-generated text is a challenging task.
How consistent are the annotations collected from crowd workers? 
In this section, we examine the agreement and variability of the collected annotations.

At a high level, we observe either acceptable or high inter-annotator agreement across error categories. 
For rare error types such as \nameBadMath, high agreement stems from the prevalence of spans with no error. 
For such categories, {\it we recommend treating each annotator as a high precision, low recall judge}, and considering the information from their aggregate annotations.
Figure \ref{fig:spans-viz} gives an example of the perspective gained by viewing all 10 annotations of a single generation.


\begin{table}[h]
    \centering
    \resizebox{\linewidth}{!}{%
    \begin{tabular}{lrr}
    \toprule
    {\bf Error} & Krippendorff's $\alpha$ & {\it Two Agree} (\%) \\ \hline
    \nameBadMath & 0.99 & 30\\
    \nameCommonsense & 0.88 & 20\\
    \nameEncyclopedic & 0.98 & 12\\
    \nameGrammarUsage~$^{>1}$ & 0.72 & 30\\
    \nameIncoherent & 0.73 & 49\\
    \nameOffPrompt & 0.71 & 61\\
    \nameRedundant & 0.88 & 38 \\
    \nameSelfContradiction & 0.87 & 26\\
    \bottomrule
    \end{tabular}
    }
    \caption{Per-token inter-annotator agreement metrics by error category. The $^{>1}$ indicates that we omit severity-1 \nameGrammarUsage~ errors in all analyses in this paper due to higher variance; including them would drop the Krippendorf's $\alpha$ to 0.56.}
    \label{tab:agree}
\end{table}

\paragraph{Agreement} 
Table~\ref{tab:agree} shows token-level inter-annotator agreement statistics aggregated over all collected data. 
Since a single annotator can label a single span with multiple errors, we break the agreement statistics down by error category.
We report Krippendorff's $\alpha$ coefficient, a chance-corrected measure of agreement for multiple annotators \citep{krippendorff2018content}.
Due to computational constraints, we calculate this coefficient per generation and report the average across the dataset.
The agreement shown here is  high for most categories ($>$0.8) and acceptable ($>$0.6) for all error types.

The Krippendorff measure may be deceptively high for some error types such as \nameBadMath, where 99\% of tokens are not annotated with this error. 
The {\it Two Agree} measure in Table~\ref{tab:agree} gives a different characterization of this data. 
{\it Two Agree} for a given error label is the percentage of tokens labeled by at least one annotator that were also labeled by one or more additional annotators. 
This metric allows us to see where annotators agree that particular errors exist while ignoring the majority of tokens (for most error categories) which annotators agree are not errors. 
{\it Two Agree} shows significantly lower rates for sparse errors with high Krippendorff scores, such as \nameEncyclopedic. 
However, it reveals stronger agreement among \nameIncoherent~and \nameOffPrompt~errors than might be expected given the Krippendorff coefficient. 

A limitation for both metrics is the use of token-based overlap.

\paragraph{Bootstrap}
One issue we face is high variance of annotations. To determine the impact of this variance for lower-data settings, we perform a bootstrap analysis using largest subset of our data (GPT-3, top-$p=0.96$, $t=1$, f.p.$=0$, for which we have annotations of 200+ generations). 
We choose 50 generations (roughly 500 annotations) and calculate the error statistics therein. 
We repeat this process 1000 times and report the mean, standard deviation, and coefficient of variation in Table~\ref{tab:bootstrap}.
We also calculate the coefficient of variation for different numbers of samples, shown in Figure~\ref{fig:cov}.
We see that as the number of samples increases, the coefficient of variation decreases as expected, though less precipitously after 30 examples. 
These results show that with as few as 50 documents, the \method error analysis should yield relatively robust results. 
However, this varies by error type: rare errors like \nameBadMath~and \nameEncyclopedic~show greater variance.
Here, again we repeat our recommendation to treat annotations for these categories in aggregate. 
These results motivate our collection of at least 500 annotations per condition studied. 

\input{fig/figcov}

\begin{table}[h]
    \centering
    \resizebox{\linewidth}{!}{%
    \begin{tabular}{lrrr}
    \toprule
    {\bf Error}& mean & std. & c.v. (\%) \\ \midrule
    \nameBadMath &8.51 &3.78& 44.5 \\
    \nameCommonsense &39.40&8.67& 22.0\\
    \nameEncyclopedic &13.56&3.94& 29.1\\
    \nameGrammarUsage &126.19&16.81& 13.3 \\
    \nameIncoherent & 96.89 & 16.58& 17.1\\
    \nameOffPrompt & 167.29 & 23.39 & 14.0 \\
    \nameRedundant & 114.77 & 22.53 & 19.6 \\
    \nameSelfContradiction & 60.54 & 11.94 & 19.7 \\ \midrule
    \nameJargon & 100.95 & 24.09 & 23.9 \\
    \nameNeedsGoogle & 482.84 & 42.22 & 8.7 \\ \midrule
    Total errors & 1268.48 & 55.59 & 19.72\\    
    \bottomrule
    \end{tabular}
    }
    \caption{
    Bootstrap analysis (sampling 50 generations) of error \textit{counts}, by category (c.v. is the coefficient of variation).
    }
    \label{tab:bootstrap}
\end{table}

%% file: fig/spans-viz-fig.tex
\begin{figure*}[t]

\begin{center}
\includegraphics[width=0.99\linewidth]{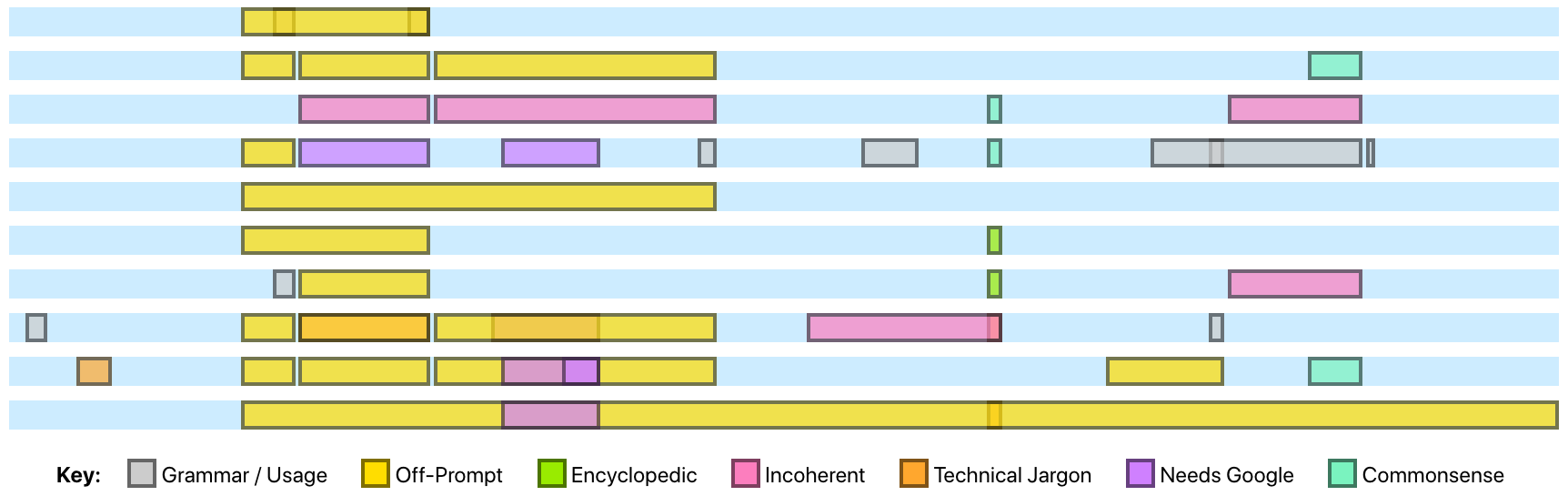}
\end{center}

\caption{
A visual representation of the 10 annotations we collected for one paragraph. Each blue bar represents one annotator, where the width of the bar represents the text of the paragraph. Colored bars drawn on top of the blue bar represent spans marked as errors. We draw bars semi-transparently to show overlapping errors. We can see that some problematic spans (e.g., the \nameOffPrompt~section) are marked by almost all workers and given the same label. Other spans are marked by only a subset of the workers (e.g., \nameCommonsense~and \nameIncoherent~spans on the right side), or have some label disagreement.
}
\label{fig:spans-viz}
\end{figure*}

%% file: fig/figcov.tex
\begin{figure}[t!]

\begin{center}
\includegraphics[width=0.99\linewidth]{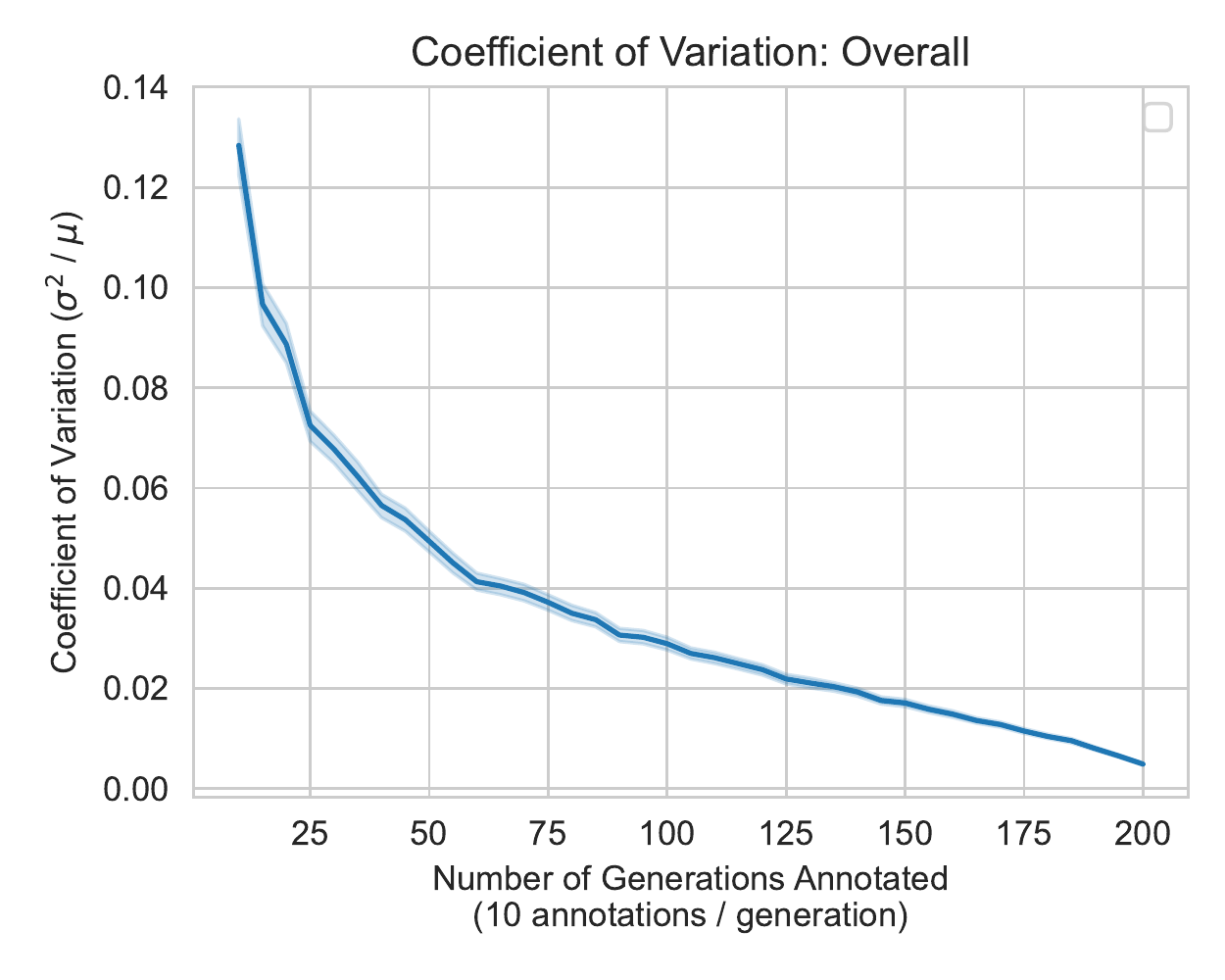}
\includegraphics[width=0.99\linewidth]{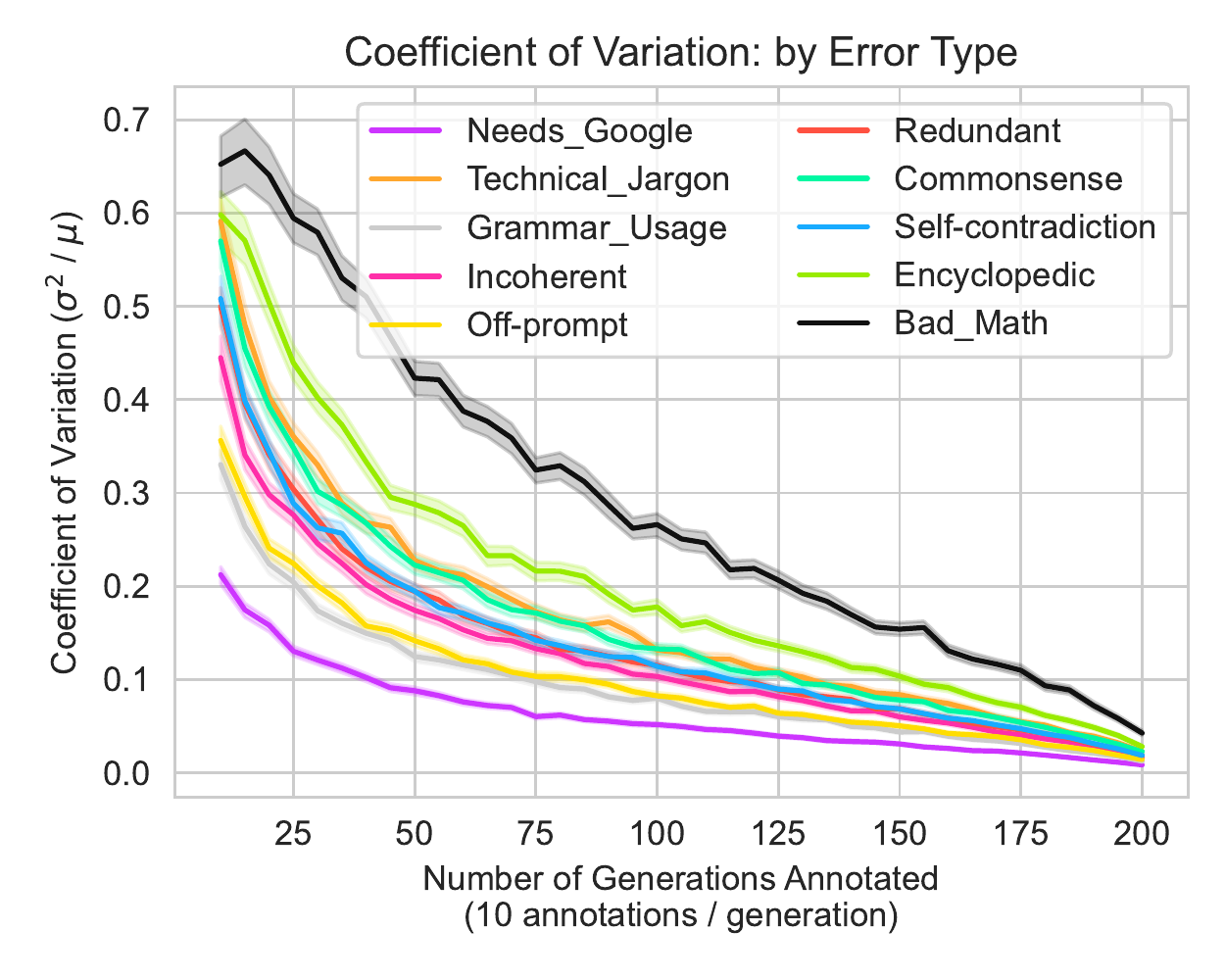}
\end{center}

\caption{
Change in coefficient of variation as number of bootstrap samples increases overall (top), and by error type (bottom), with 95\% confidence intervals. Data shown for GPT-3 with apples-to-apples decoding configuration (top-$p=0.96$, $t=1$, no \textit{f.p.}).
}
\label{fig:cov}
\end{figure}

%% file: src/data-details.tex
\section{Dataset Statistics}
\label{sec:data-statistics}

\input{fig/dataset-statistics}
\input{fig/dataset-stats-pies}
\input{fig/span-lengths-fig}



We list the data collection quantities in Table~\ref{tab:dataset-stats}, and plot visualizations of three aspects: prompt topic and annotated span proportions are shown in Figure~\ref{fig:dataset-stats-pies}, and average span lengths are shown in Figure~\ref{fig:span-lengths}.

%% file: fig/dataset-statistics.tex
\begin{table}[t]
\centering
\resizebox{\linewidth}{!}{%
\begin{tabular}{@{}lrrrrrr@{}}
\toprule
\textsc{model} & top-$p$ & $t$ & \textsc{f.p.} & \textsc{gens} & \textsc{anns} & \textsc{spans} \\ \midrule
\textsc{gpt-2 s} & 0.96 & 1.00 & 0 & 81 & 809 & 3694 \\
\textsc{gpt-2 xl} & 0.96 & 1.00 & 0 & 81 & 806 & 3087 \\
\textsc{grover-mega} & 0.96 & 1.00 & 0 & 80 & 796 & 3006 \\
\midrule[\cmidrulewidth]
\textsc{gpt-3} & 0.40 & 1.00 & 0 & 66 & 660 & 2064 \\
 & 0.70 & 1.00 & 0 & 65 & 648 & 1841 \\
 & 0.90 & 1.00 & 0 & 63 & 629 & 1794 \\
 & \textit{n/a} & argmax & 0 & 66 & 659 & 2153 \\
 & 0.96 & 0.40 & 0 & 65 & 650 & 2249 \\
 & 0.96 & 0.70 & 0 & 61 & 610 & 1865 \\
 & 0.96 & 1.00 & 0 & 206 & 2055 & 6234 \\
 & 0.40 & 1.00 & 1 & 50 & 500 & 1280 \\
 & 0.70 & 1.00 & 1 & 53 & 530 & 1481 \\
 & 0.90 & 1.00 & 1 & 54 & 540 & 1717 \\
 & \textit{n/a} & argmax & 1 & 51 & 509 & 1384 \\
 & 0.96 & 0.40 & 1 & 53 & 530 & 1401 \\
 & 0.96 & 0.70 & 1 & 50 & 498 & 1369 \\
 & 0.96 & 1.00 & 1 & 84 & 838 & 2947 \\
\midrule[\cmidrulewidth]
\textsc{human} &  &  &  & 79 & 789 & 2296 \\
\midrule \midrule
\textsc{total} &  &  &  & 1308 & 13056 & 41862 \\ \bottomrule
\end{tabular}%
}
\caption{Statistics of data annotated with \method. $t$ is the (softmax) temperature, and \textsc{f.p.} is a frequency penalty for already-generated words (explained in \S\ref{sec:decoding-strategies}). \textsc{gens, anns,} and \textsc{spans} are then number of generations, annotations over those generations, and error spans marked during the annotations, respectively. We perform the most annotations on the strongest available generative model (GPT-3). 
}
\label{tab:dataset-stats}
\end{table}

%% file: fig/dataset-stats-pies.tex
\begin{figure}[t]

\begin{center}
\includegraphics[width=0.47\linewidth]{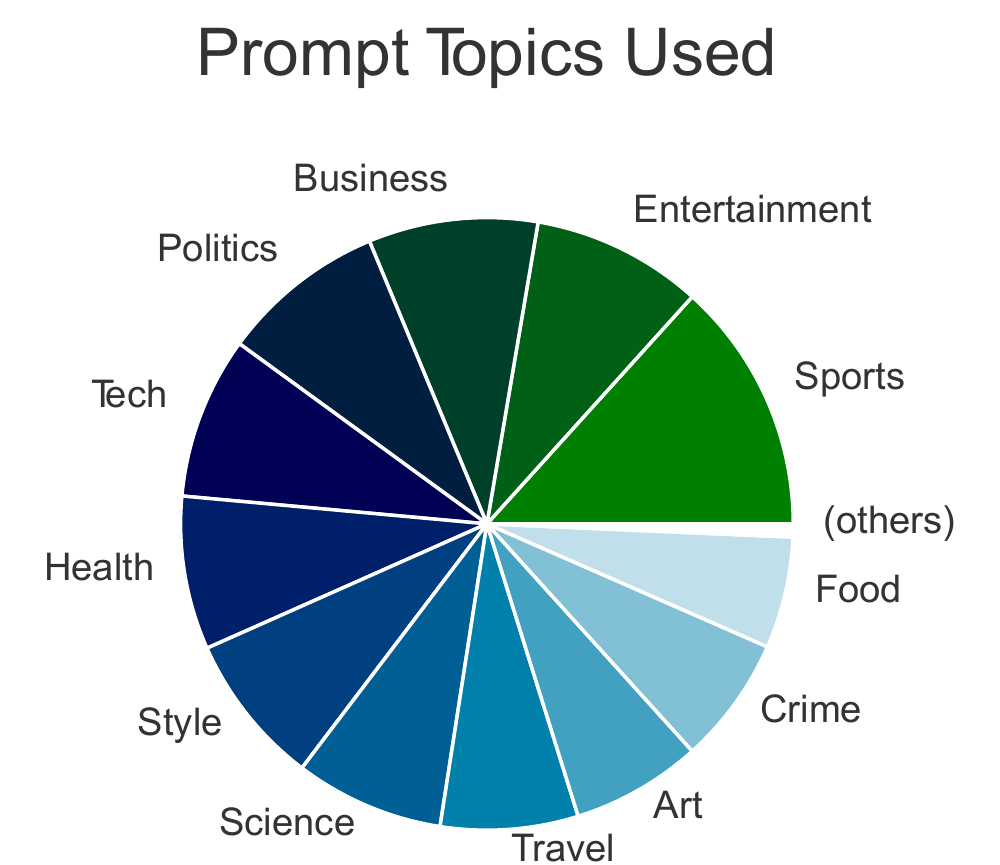}
\includegraphics[width=0.51\linewidth]{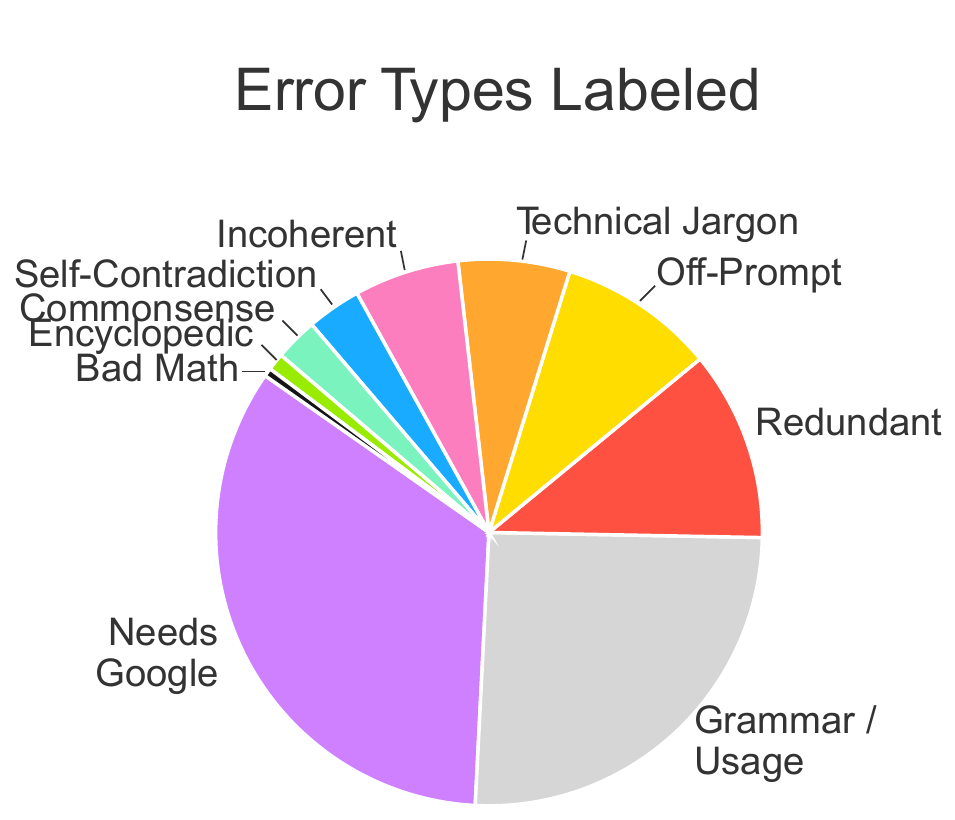}
\end{center}

\caption{
Visual overviews of the distribution of prompt topics used for generating the 1.3k paragraphs used in the annotation (left), and the types of the 41k spans labeled during the annotation (right).
}
\label{fig:dataset-stats-pies}
\end{figure}

%% file: fig/span-lengths-fig.tex
\begin{figure}[ht]

\begin{center}
\includegraphics[width=0.99\linewidth]{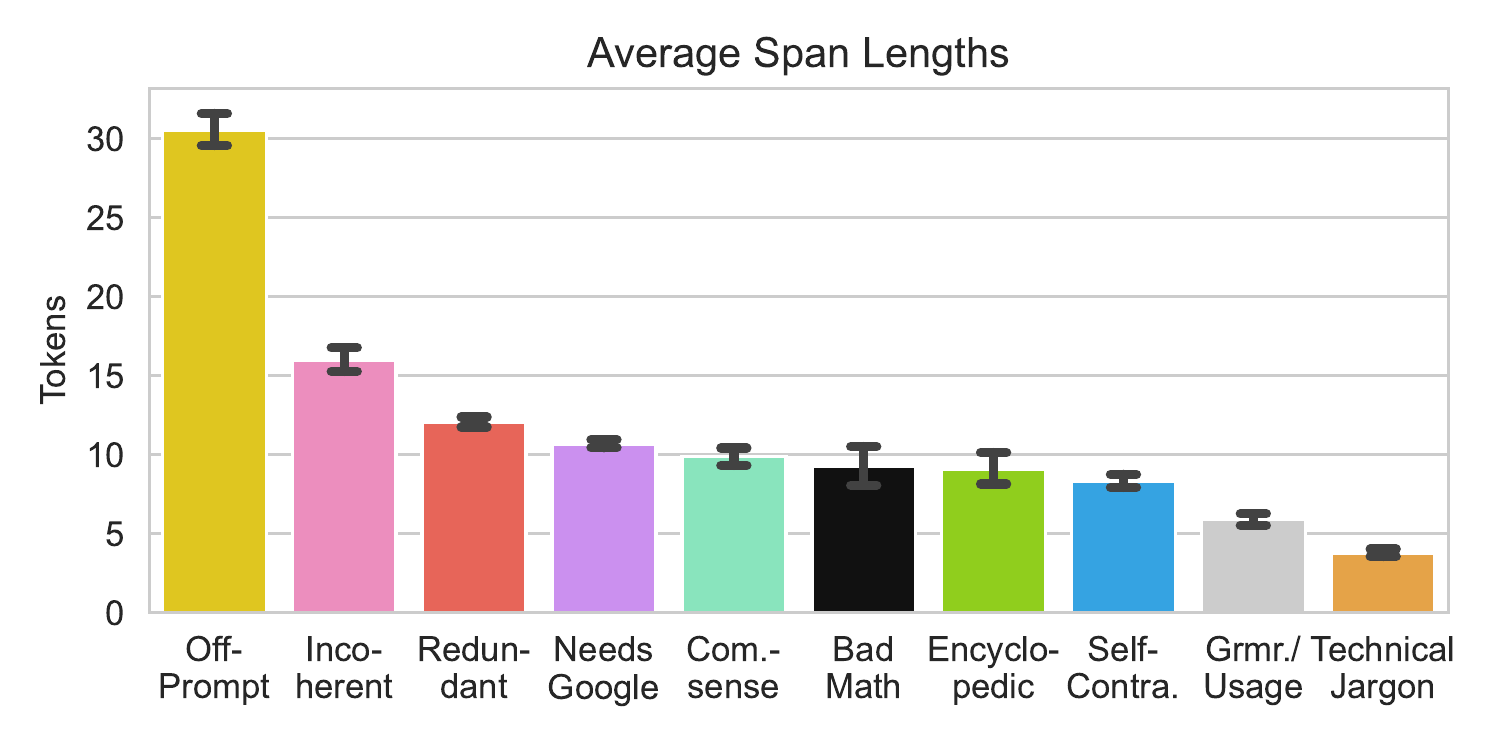}
\end{center}

\caption{
Average number of tokens covered by each annotated span.
We observe span length correlates with how abstract the error category is, from word-level issues (\nameJargon), through phrase-level semantics (e.g., \nameCommonsense), and into problems of pragmatics (\nameOffPrompt).
}
\label{fig:span-lengths}
\end{figure}

%% file: src/analysis.tex
\section{Detailed Analysis}
\label{sec:analysis}

\input{fig/error-measuring-comparison}

In this section we perform a detailed analysis of the trends of individual error types and decoding configurations.

To begin, we consider apples-to-apples model decoding configurations. To expand on these results, 
originally presented in Figure \ref{fig:per-label-wide}, 
we also present two additional ways of counting error spans, which we show in Figure \ref{fig:error-measuring-comparison}. While our method for counting errors throughout the paper takes into account the number of tokens covered in each span (\textit{span coverage}), we also show plots for scaling each span by its severity level (\textit{span coverage $\times$ severity}), and by ignoring both severity and token length (simply \textit{span counts}). These changes in measurement further illuminate model error characters, which we discuss in the upcoming sections (refer to Figure \ref{fig:error-measuring-comparison}).

\subsection{\nameOffPrompt}
\label{sec:analysis:offPrompt}
Under initial analysis of \textit{span coverage}, \nameOffPrompt~errors show a \iconModelPlateau \textit{model plateau} at GPT-3. Measuring \textit{span counts} offers barely perceptible improvement, indicating that scaling language models over more in-domain training does not guarantee topicality.

This observation is consistent with growing work on \textit{prompt programming} as a new technique for attempting to steer large pretrained models to complete the desired task \cite{branwen_2020,gao2020making,reynolds2021prompt}. In practice, we observe that while GPT-3 will sometimes continue a prompt by writing an article, other times, it may elaborate on the prompt itself:

\begin{small}
\begin{quote}
    \textsc{\textbf{prompt}}\\
    Do you prefer the idea of being outdoors in the fresh air to being stuck inside with phones ringing and messages pinging?
\vspace{0.5em}\\
    \textbf{GPT-3}\\
    Can you leave work at work? Are you flexible enough to \errOffPrompt[cover holidays or take on additional responsibilities? Can you prioritize tasks? If your boss comes to you on Tuesday to confirm the new social media strategy, are you able to pick up the ball and get the messaging hammered out by Thursday?] ...
\end{quote}
\end{small}

Of course, this generation is not \textit{literally} \nameOffPrompt, but it is out of place when other generations are continuations of the prompt, rather than further elaborations of it.

While avoiding \nameOffPrompt~errors for language models is worth exploring with prompt programming and other avenues, an investigation of these techniques is outside the scope of this work.

Finally, we note that \nameOffPrompt~spans are the most prevalent \textit{error} (not reader issue) marked for human-authored text. We suggest that a higher rate of false positives for this error type, coupled with its prevalence in model-generated text, makes further refinement of this error a compelling avenue for further study.

\subsection{\nameSelfContradiction}
\label{sec:analysis:selfContradiction}

While changing from \textit{span coverage} to \textit{span counts} alters the relative order of GPT-2 XL and Grover (though still within confidence bounds), the puzzling question is why GPT-2 Small performs better than most (or all) other models. Why would the smallest model produce the fewest \nameSelfContradiction~errors?

We posit the reason is that GPT-2 generations are so \nameIncoherent~and \nameOffPrompt~that there is little opportunity for relevant, comprehensible points to be made and then reversed. For example, see the GPT-2 Small annotated generation in the top left of Figure \ref{fig:4_examples}. The entire text is covered by \nameOffPrompt~and \nameIncoherent~errors.\footnote{The high double-error coverage reveals another consideration: to what \textit{depth} (i.e., number of overlapping spans) will annotators mark? By the design of our framework, \nameIncoherent~errors serve as a fall-back, but without it, we might imagine poor generations splatter-painted by other error types.} If we look at GPT-2 Small's error distribution in Figure \ref{fig:stacked-bar-overall}, we see most of its added density comes from significantly more \nameOffPrompt~and \nameIncoherent~tokens.

\subsection{\nameRedundant}
\label{sec:analysis:redundant}

The different counting methods shown in Figure \ref{fig:error-measuring-comparison} reveal a change in the results for \nameRedundant~errors. Rather than repetition simply increasing as models grow larger, we observe that GPT-3 repeats in a similar number of cases (lower span \textit{counts}), but for more tokens (higher span \text{coverage}). This matches the qualitative observation that GPT-3 produces larger \textit{topically} repetitive blocks, rather than simple word or phrase repetitions generated by GPT-2-sized models:

\begin{small}
\begin{quote}
    \textbf{GPT-2 Small}\\
    ... owners have started growing their own breeds and dogs are \errRedundant[starting to start] so there's really ...
    \vspace{0.5em}\\
    \textbf{GPT-3}\\
    The focus of your thoughts should be on the task at hand, \errRedundant[not on your productivity. You shouldn't be thinking about how you can be more productive. You should be thinking about how you can be productive right now.] ...
\end{quote}
\end{small}

Such repetitions can be more difficult to clearly isolate, because even slight wording changes produce variations in tone and connotation. Rather than being identical \textit{semantically}, we observe GPT-3 will seem stuck on a particular \textit{topic}, elaborating on and rephrasing similar ideas more times than a human writer (hopefully) would.

\subsection{Reader Issues}
\label{sec:analysis:readerIssues}

\input{fig/topic-label-heatmaps}

We observe the highest number of \nameNeedsGoogle~and \nameJargon~issues in human-authored text.

\nameNeedsGoogle~issues broadly represent any specific claim that could be fact-checked. In our domain (news articles), these are primarily whether an event happened on a particular day, whether a person holds a role, or whether a mechanism works as described (e.g., chemical or technical). As seen in Figure \ref{fig:heatmaps-topic} (which shows GPT-3's span distribution), \nameNeedsGoogle~issues happen roughly equally for all topics. We believe this trend is due to the news article domain, which is prone to a high density of specific information. As such, for other domains, this trend may be less prevalent, more difficult to label (e.g., subtle claims assumed to be true in long running text), or both.

We observe that \nameJargon~issues are influenced by topic (Figure \ref{fig:heatmaps-topic}, bottom), occurring significantly more frequently in \textit{Business, Health, Science,} and \textit{Technology} topics than in others. This trend displays a clear topic-dependence even within a single broader domain (news). These results indicate that both reader issues are characteristics of natural text. Of course, one might wish to measure or minimize potential reader issues for a particular application---for example, claim verification, or controlling for reading level.

\subsection{Decoding Hyperparameters}
\label{sec:analysis:decodingConfigs}

\input{fig/heatmaps-per-label-fig}

\input{fig/heatmaps-frequency-penalty-fig}

We discuss the effects of the decoding hyperparameters we consider---top-$p$, temperature, and frequency penalty---on generation quality. For the sake of annotation cost, we only vary these parameters for the strongest model available, GPT-3.

\input{fig/whats-next-figs}

First, we show the effect of varying top-$p$ and temperature alone (i.e., with no frequency penalty) on different error types. Figure \ref{fig:heatmaps-per-label} shows the effect on two salient spans: \nameOffPrompt~and\nameRedundant. (We omit others for space.) We observe that annotators naturally label errors the way we would intuitively expect the model to produce them, given the hyperparameter changes. The bottom-right corner of each subplot, where $t=1$ and $p=0.96$, is the configuration with the highest amount of randomness from sampling. As we move away from that corner---either left by lowering temperature, or up by lowering top-$p$---we lower the amount of randomness. We observe a positive correlation with randomness and \nameOffPrompt~errors, and an inverse correlation with \nameRedundant~errors. In other words, sampling from a larger set of words makes the model more prone to changing topics, but less likely to repeat itself, and vice versa.

After confirming these intuitive measures, we turn our attention to Figure \ref{fig:heatmaps-frequency-penalty}, which investigates the overall error spans for GPT-3 both without (left) and with (right) the frequency penalty. (Note that unlike Figure \ref{fig:heatmaps-per-label}, both heatmaps in Figure \ref{fig:heatmaps-frequency-penalty} have the same color scale.) We observe that introducing the frequency penalty lowers error rates for every value of temperature and top-$p$ that we try. Furthermore, it appears to reverse the trend seen without a frequency penalty: that sampling from a larger set of words produces fewer errors.

The overall results for all decoding configurations were shown previously in Figure \ref{fig:model-variants}. In the next section, we focus on the GPT-3 decoding configuration that produced the fewest number of errors, and compare it to human authored text.

\subsection{Best GPT-3 vs. Humans}
\label{sec:analysis:whatsNext}

The best GPT-3 configuration shown in Figure \ref{fig:model-variants}---argmax sampling with frequency penalty = 1---appears to match error rates seen in human text. Is the text generated by this model truly as error-free as news articles?

We first look at the error composition of both sets of annotations. To get a clear picture of the potential problems, we plot only error spans (ignoring reader issues), and we omit length scaling, instead plotting span counts. This breakdown is shown in the left plot of Figure \ref{fig:whats-next-figs}. The error compositions are similar, the largest differences being more \nameRedundant~errors for GPT-3, and more \nameGrammarUsage~errors for human-authored text.

Next, we perform a manual analysis of 160 errors, sampling 10 at random from each of the 8 error types for each model (GPT-3 and human-authored text). We show the results in the center plot of Figure \ref{fig:whats-next-figs}. We notice that a greater portion of errors in human-authored text were due to artifacts present in the text-only format of the Common Crawl. For example, links to other articles or advertisements sometimes appear in the middle of an article's text. While annotators were quick to mark these spans, they reflect errors in formatting, not in writing. We partition these errors separately and exclude them from the subsequent calculations.\footnote{GPT-3's generations also sometimes exhibited what appeared to be formatting errors due to training on web-scraped text, though more rarely. For example, some generations contained \textit{Which?} after vague noun phrases, which appear to be learned from Wikipedia, where under-specified information is tagged by an editor with this word. For fairness, we removed these errors from GPT-3's tally as well, though they were few enough we do not plot them separately.}

Finally, we scale each error type's prevalence for each model (i.e., the left plot of Figure \ref{fig:whats-next-figs}) by the portion of errors that we estimate to be legitimate based on our manual annotation (i.e., Figure \ref{fig:whats-next-figs}, center) to produce the right plot of Figure \ref{fig:whats-next-figs}. After taking into account each error type's frequency, we estimate that 48\% of GPT-3's worker-annotated errors overall are legitimate, compared to 9\% for human-written articles.

This analysis suggests two findings. First, human-authored news paragraphs contain many times fewer issues than text authored by GPT-3 using the best decoding configuration we tested. Second, the noise of error annotations may be as high as 90\% when assessing high-quality text. Though it would require further manual annotation to verify, we conjecture that the trend of GPT-3's error spans being more reliable (only 50\% noise) would continue, and that text generated by GPT-2 would contain even fewer false positives. We note that such rates are not fixed---after all, the manual annotations were done by one of the authors simply by reading carefully---but that more realistic text may require correspondingly more effort by human annotators.

%% file: fig/error-measuring-comparison.tex
\begin{figure*}[ht!]

\begin{center}

\includegraphics[width=0.19\linewidth]{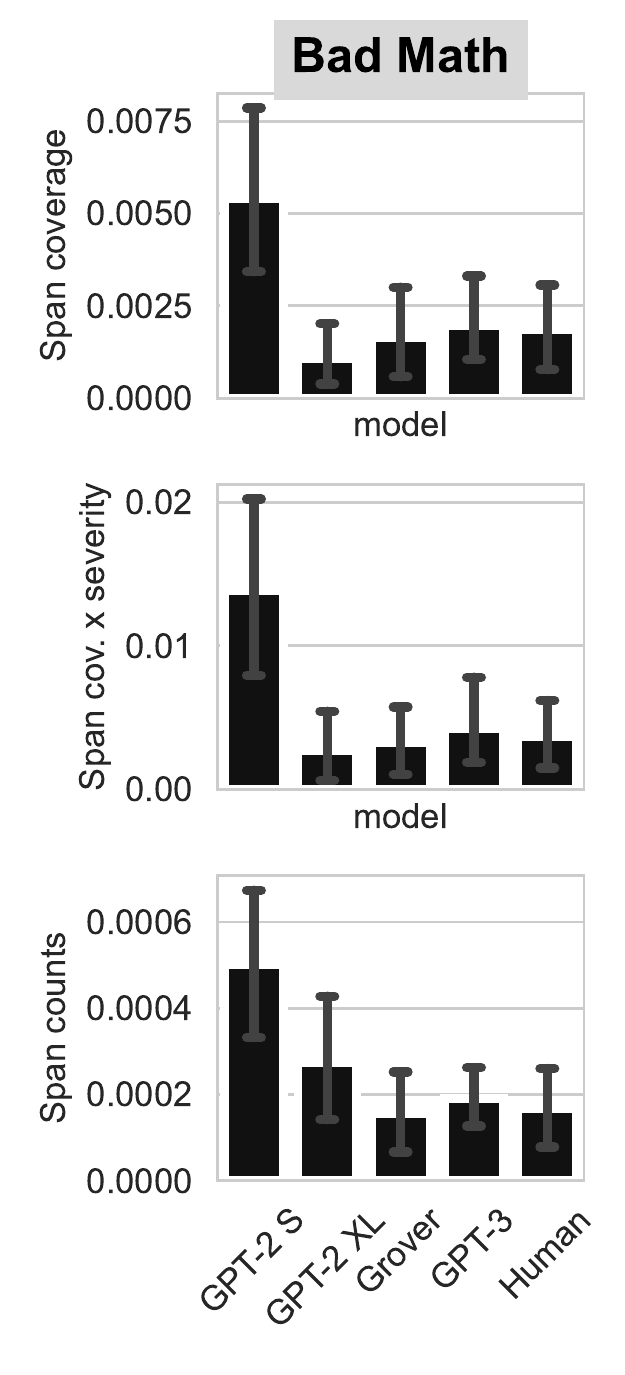}
\includegraphics[width=0.19\linewidth]{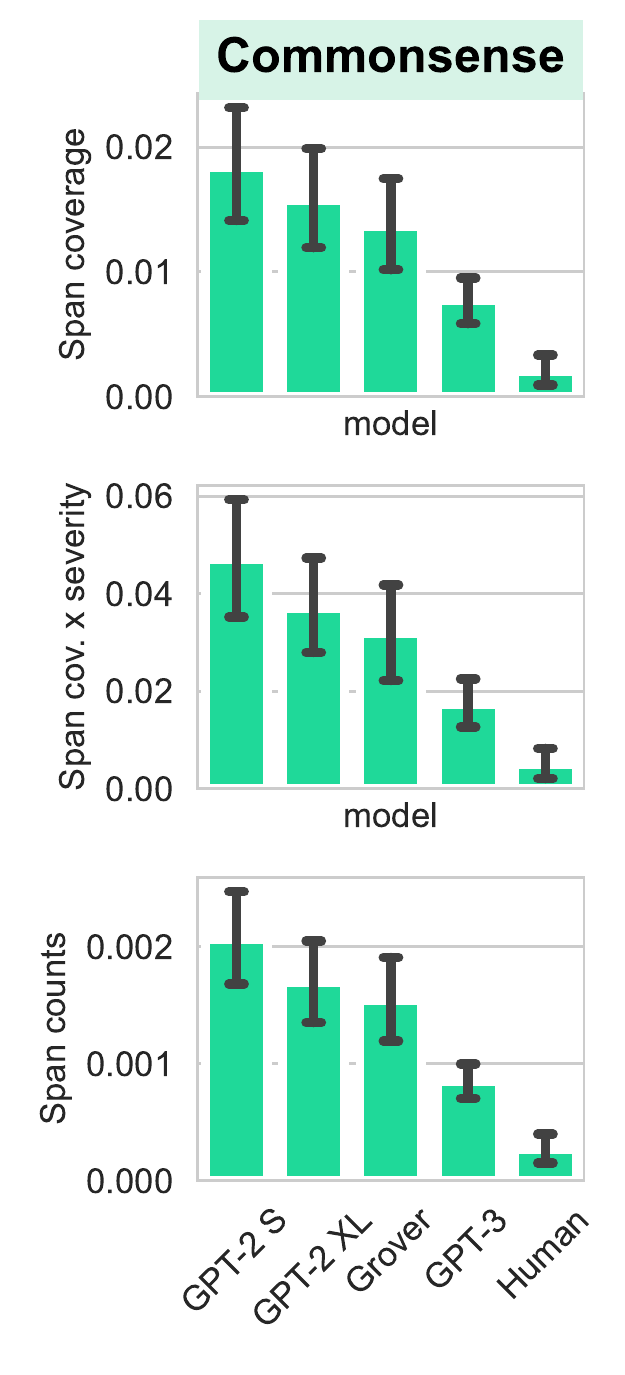}
\includegraphics[width=0.19\linewidth]{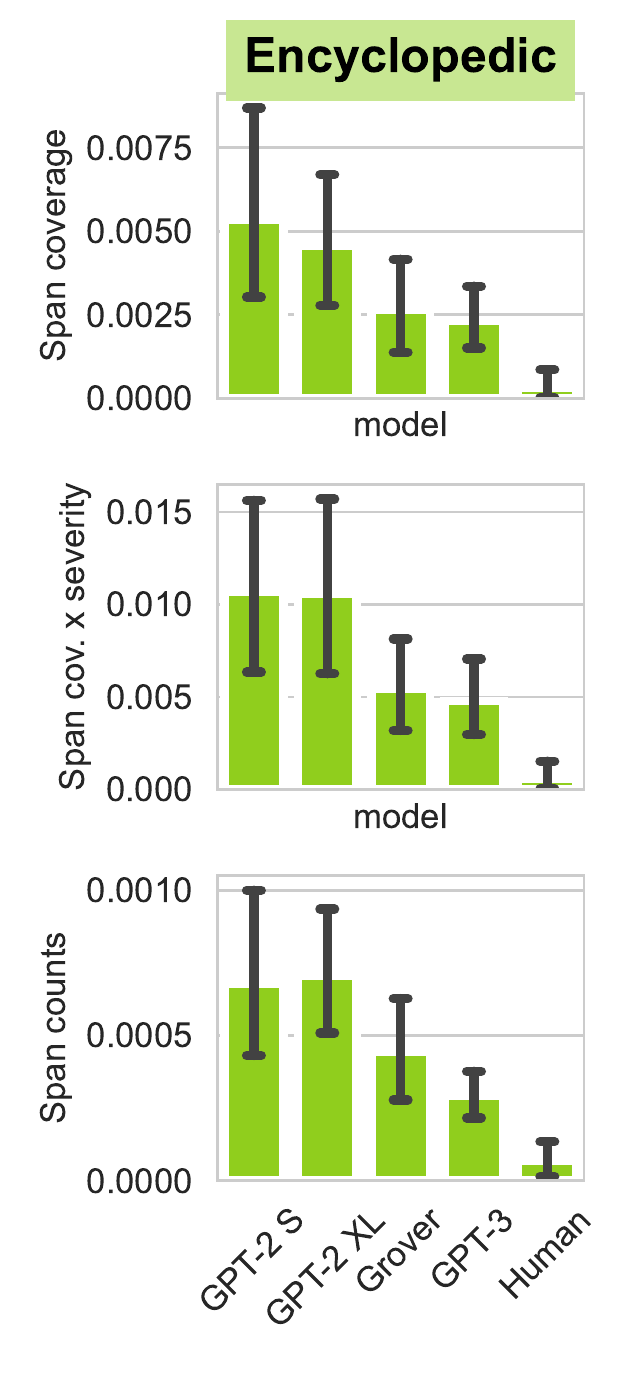}
\includegraphics[width=0.19\linewidth]{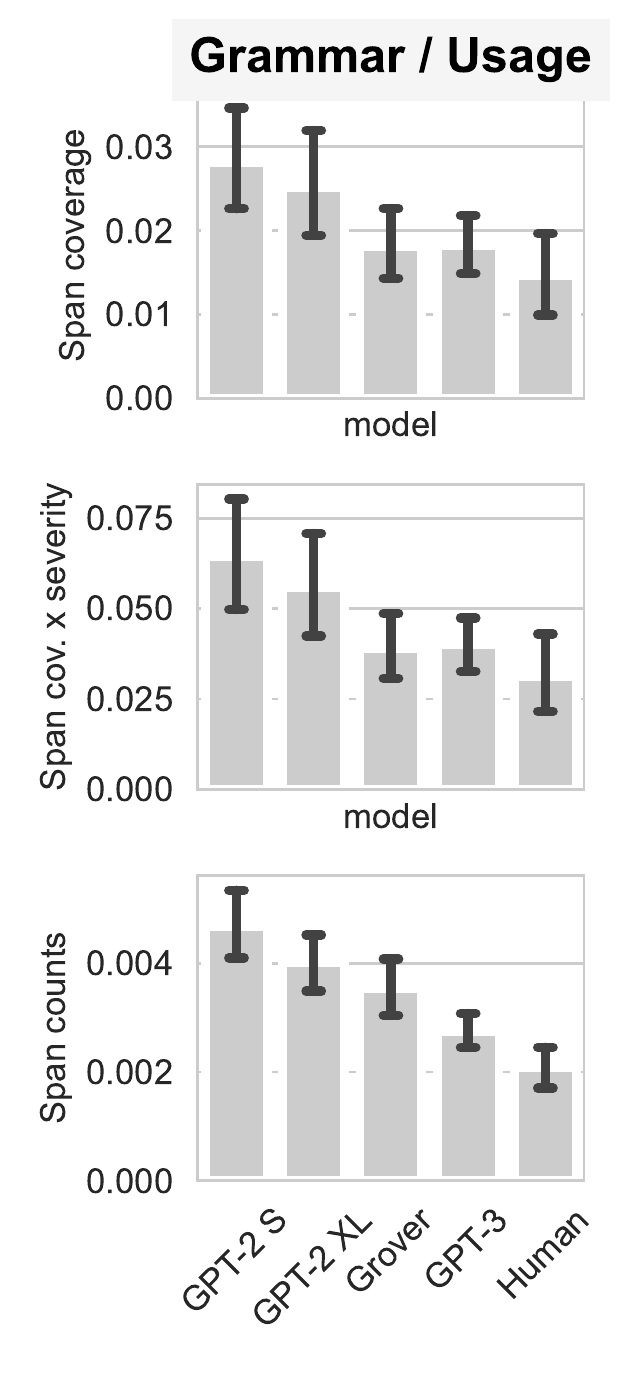}
\includegraphics[width=0.19\linewidth]{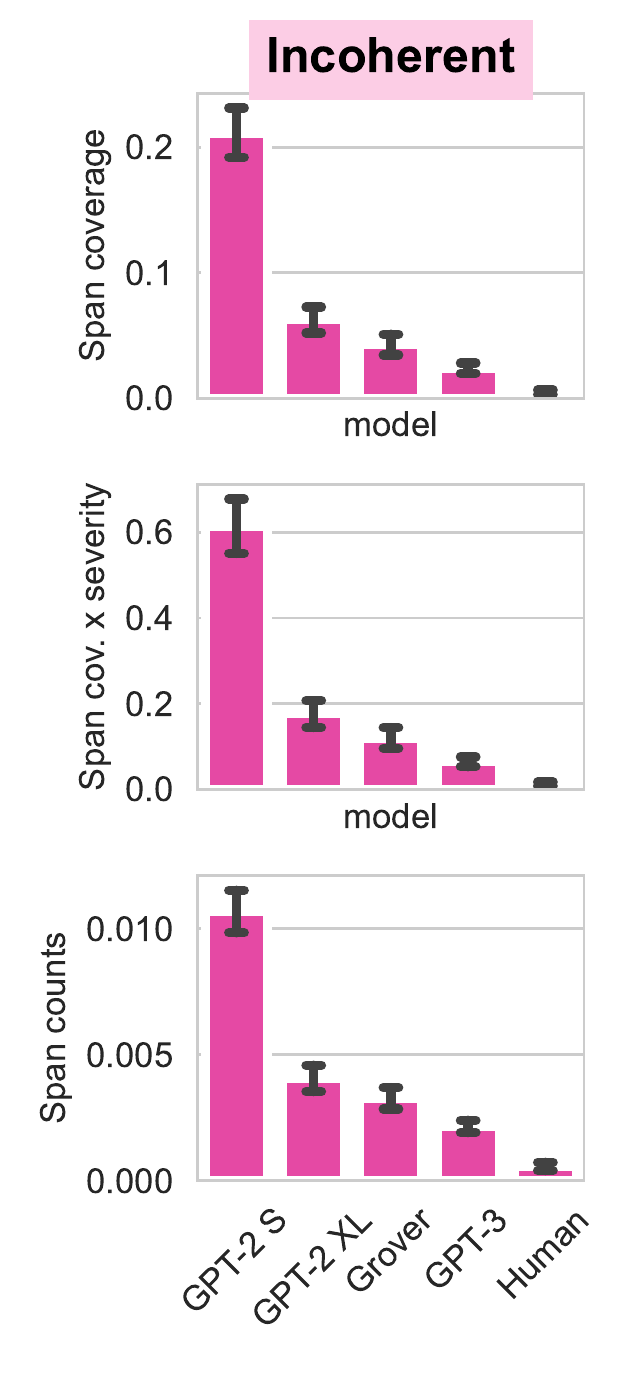}

\includegraphics[width=0.19\linewidth]{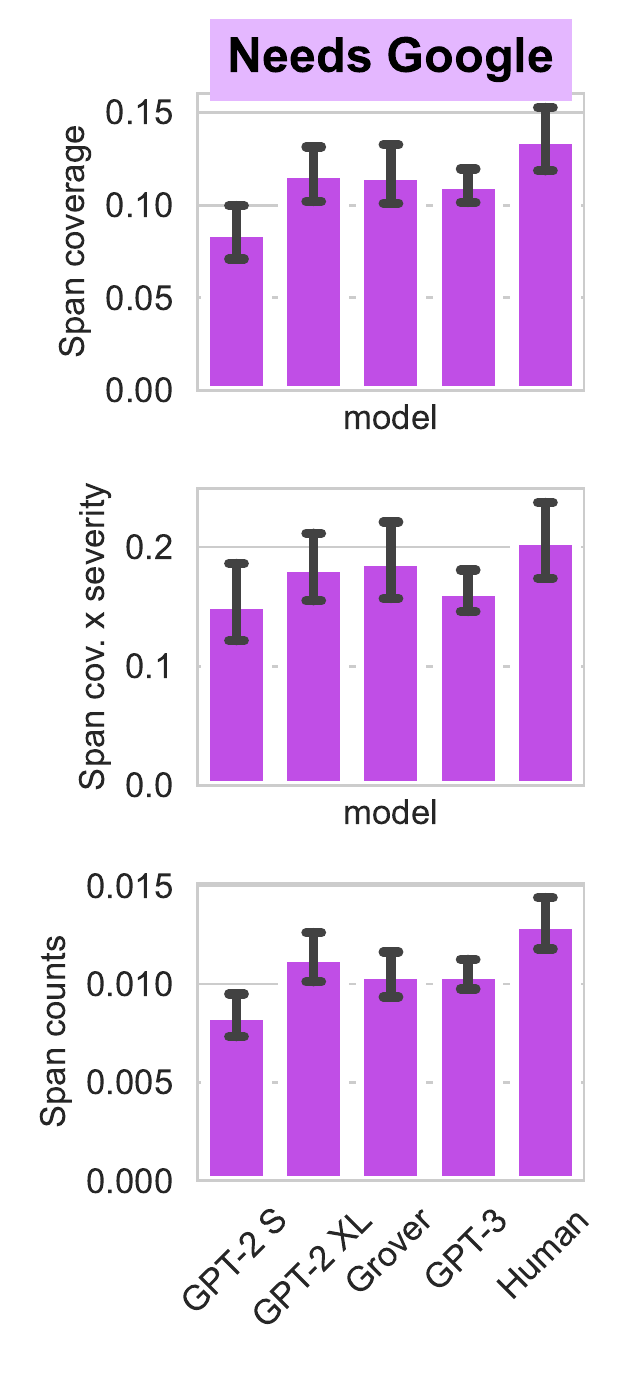}
\includegraphics[width=0.19\linewidth]{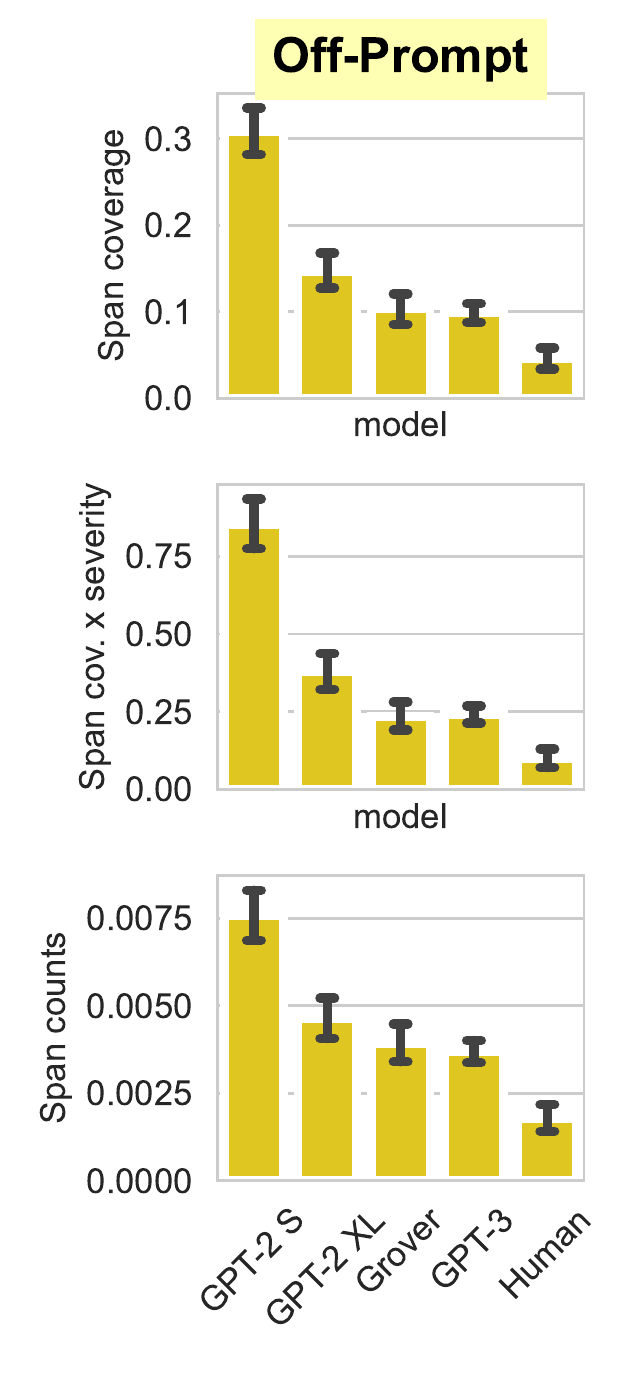}
\includegraphics[width=0.19\linewidth]{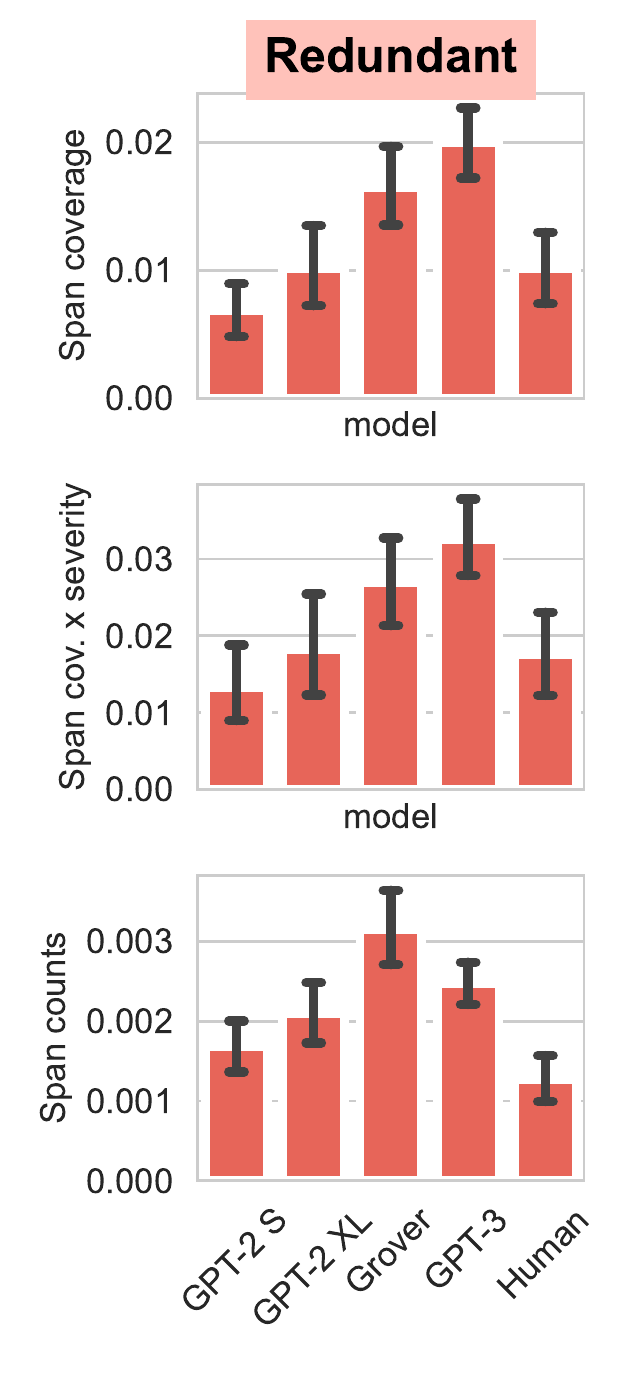}
\includegraphics[width=0.19\linewidth]{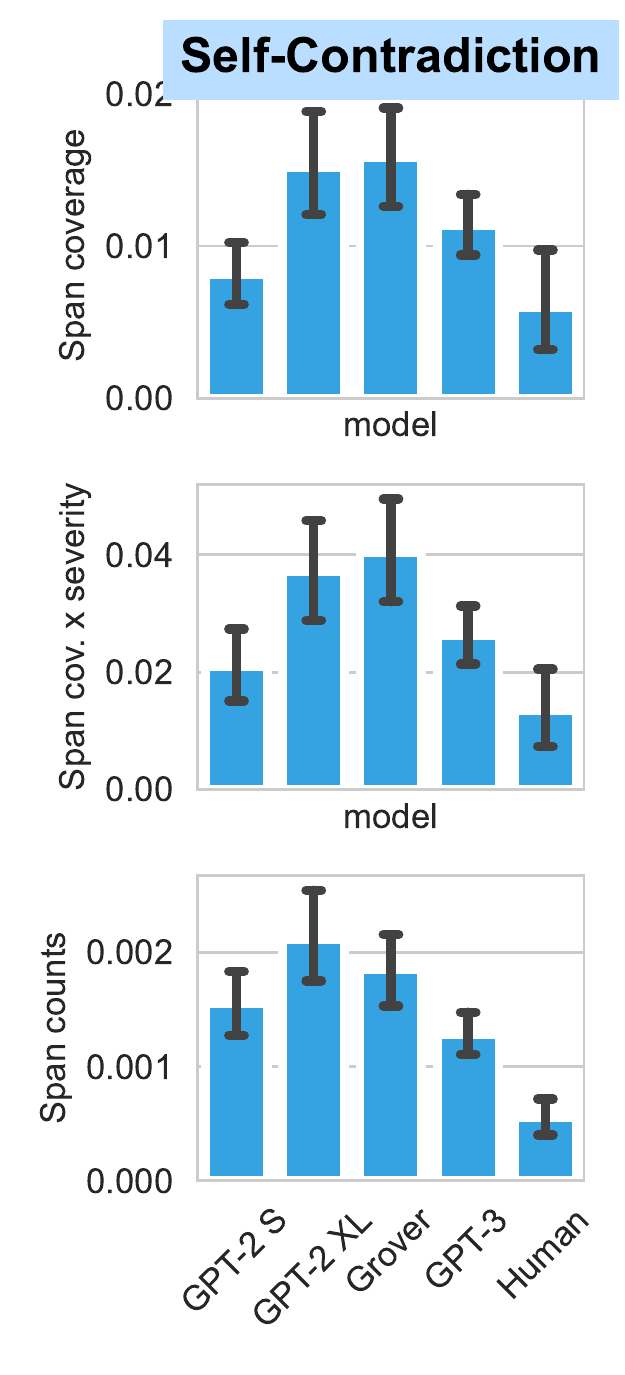}
\includegraphics[width=0.19\linewidth]{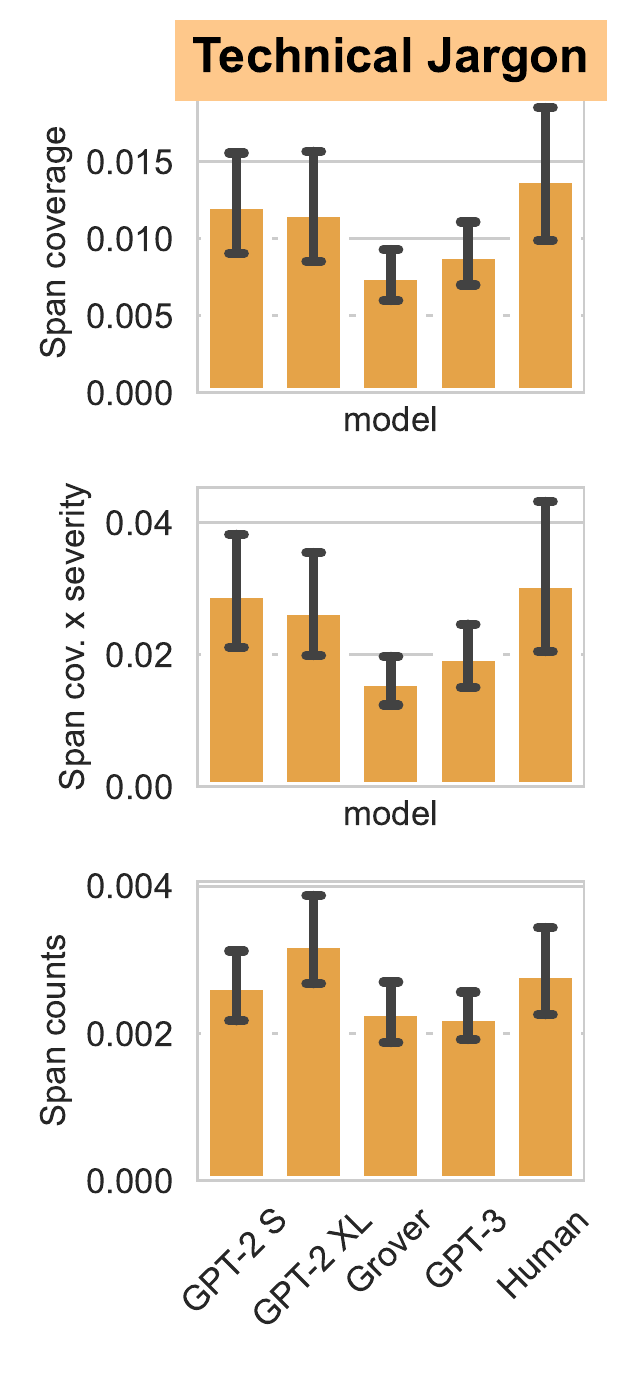}

\end{center}

\caption{
Comparison of three different ways of measuring quantities of error span annotations, shown per label. 
(The top plot for each error type is identicial to the one shown in Figure \ref{fig:per-label-wide}.)
The top method (\textit{span coverage}) is used in the rest of the paper; we provide the comparisons here to illustrate how this decision affects analysis.
\textbf{Top subplots:} \textit{span coverage}, where the number of tokens annotated as the error span are divided by the length of each annotation. (Annotations with no spans count as 0.) Intuitively, this measures the expected portion of tokens that will be covered by an error span.
\textbf{Middle subplots:} \textit{span coverage $\times$ severity}, like the top measure, but each span's token count is multiplied by its severity, more harshly penalizing errors intuitively marked as worse.
\textbf{Bottom subplots:} \textit{span counts}, where each error span simply counts as 1, regardless of the span length.
In all cases, model configurations are set as closely as possible (top-$p=0.96$, $t=1.0$, no frequency penalty), severity-1 grammar errors are removed (see \S\ref{sec:data-quality}), and 95\% confidence intervals are shown as bands.
\textbf{Takeaways:} Compared to the approach used in the rest of the paper (\textit{span coverage;} top), scaling by severity (middle) does not affect the relative model ordering, primarily widening confidence intervals. However, ignoring span lengths (bottom) does affect the results in several cases. \nameGrammarUsage~and \nameEncyclopedic~develop clearer \iconDecreasing \textit{decreasing} shapes, previously suffering from various levels of \iconModelPlateau \textit{model plateau} at GPT-3. Furthermore, the relative model ordering is changed for \nameRedundant, \nameSelfContradiction, and \nameJargon~spans.
}
\label{fig:error-measuring-comparison}
\end{figure*}

%% file: fig/topic-label-heatmaps.tex
\begin{figure}[th]

\begin{center}
\includegraphics[width=0.99\linewidth]{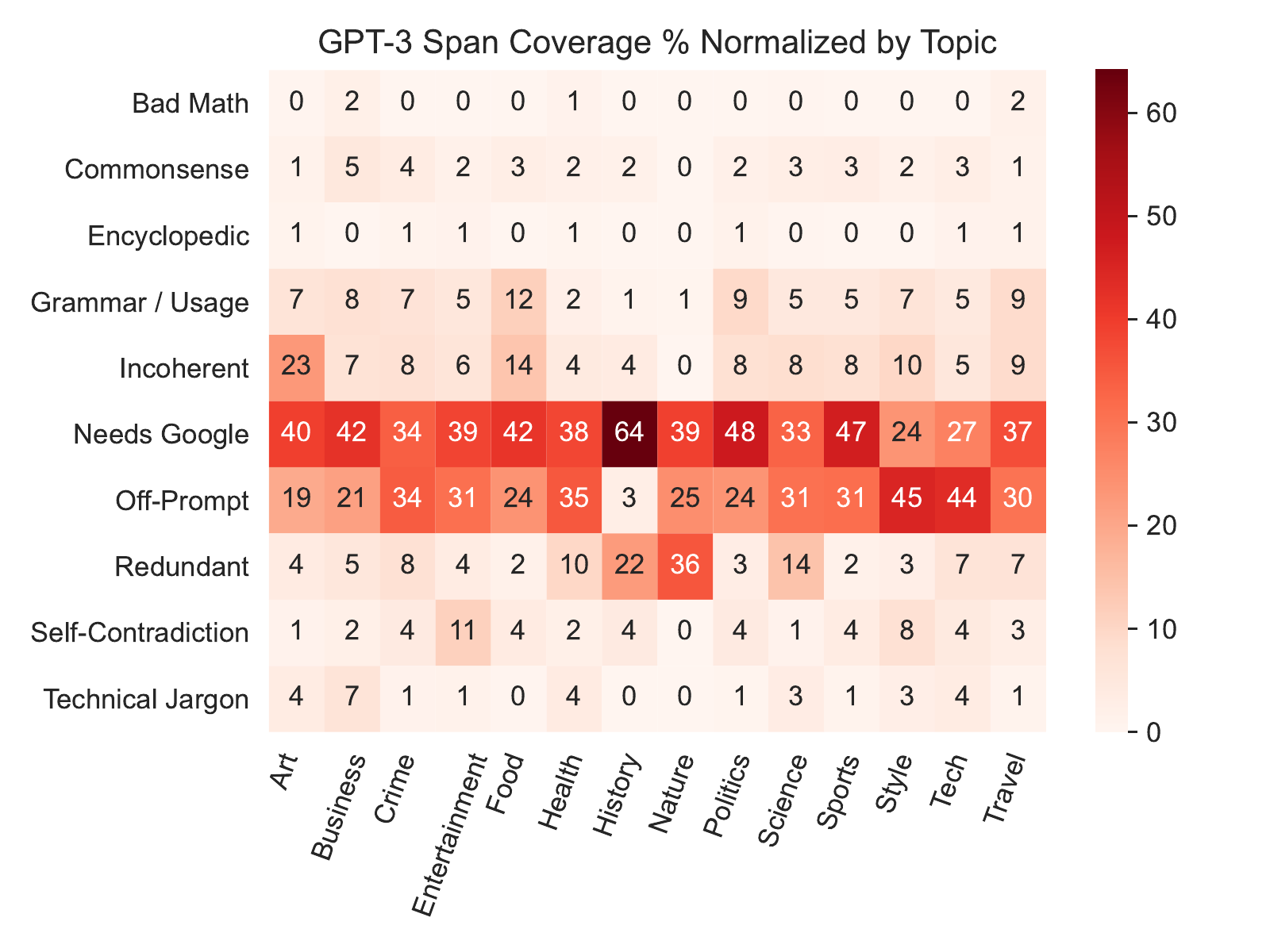}\\
\includegraphics[width=0.99\linewidth]{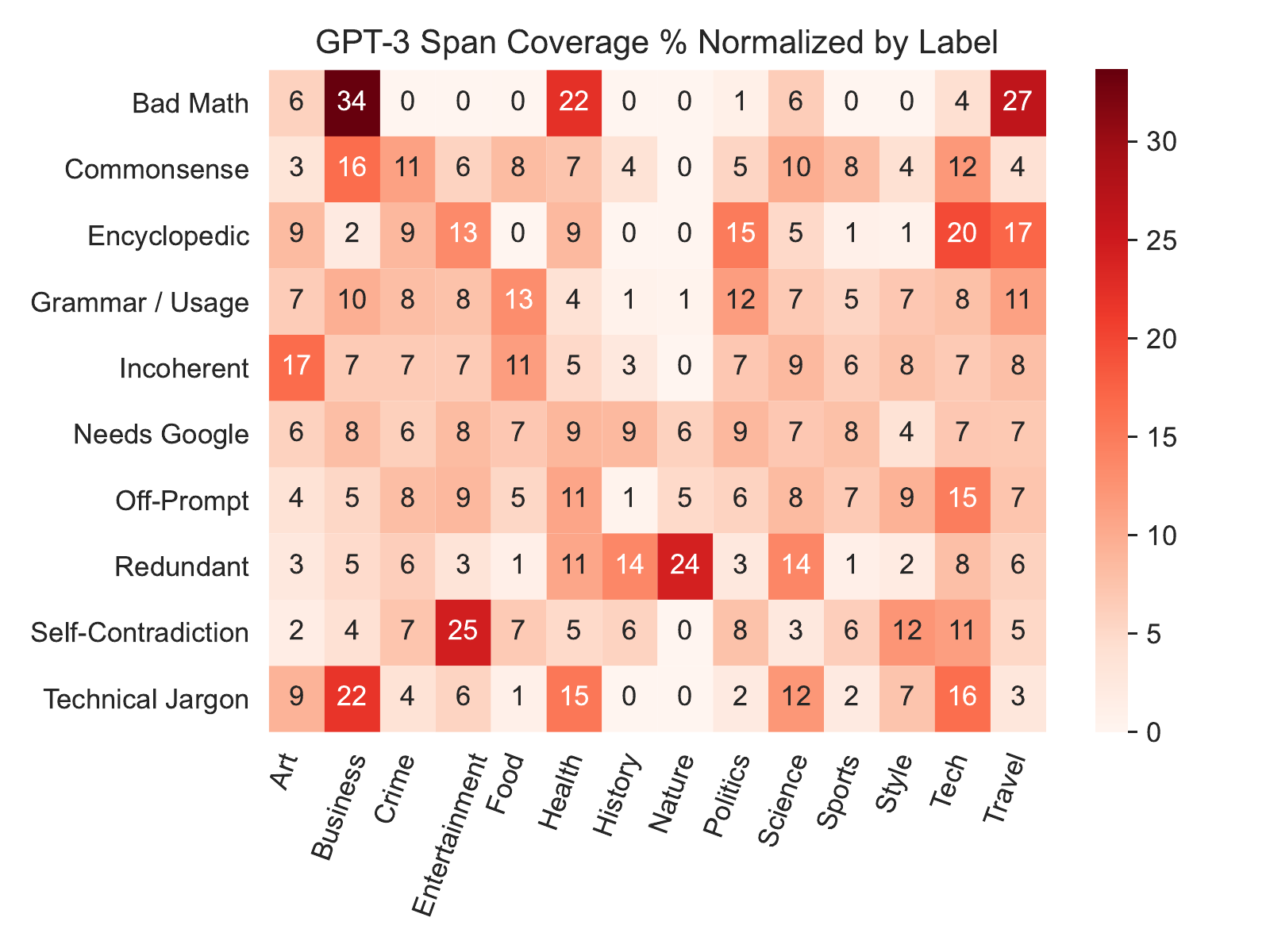}
\end{center}

\caption{
Span coverage across both topic (x-axis) and span label (y-axis) for GPT-3 generated spans (\textit{apples-to-apples} decoding: $p=0.96, t=1,$ and no frequency penalty).
\textbf{Top:} normalized by topic (column); \textbf{bottom:} normalized by error type (row).
}
\label{fig:heatmaps-topic}
\end{figure}

%% file: fig/heatmaps-per-label-fig.tex
\begin{figure}[ht]

\begin{center}
\includegraphics[width=0.49\linewidth]{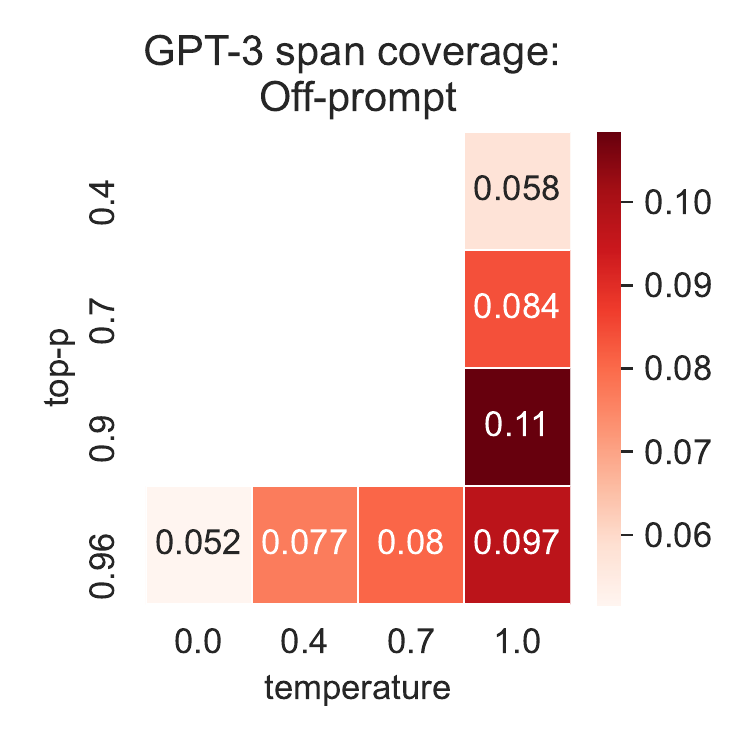}
\includegraphics[width=0.49\linewidth]{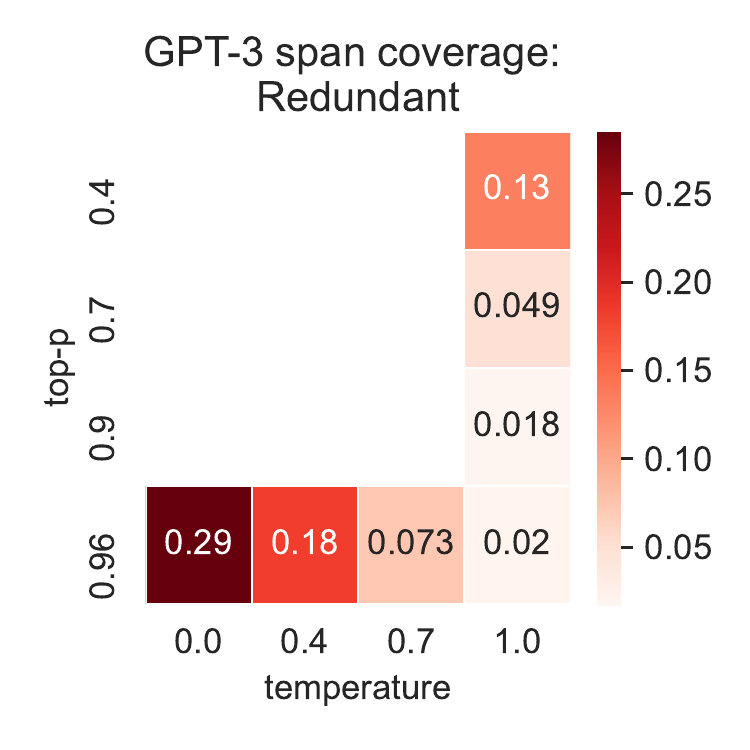}
\end{center}

\caption{
GPT-3 span coverage for \nameOffPrompt~(left) and \nameRedundant~(right) for values of top-$p$ and temperature ($t=0$ is argmax; both plots with no frequency penalty; argmax sampling is agnostic to the top-$p$ value, so we simply plot it in the $p=0.96$ cell).
\textbf{Takeaway:} Our annotation confirms intuitive expectations of the effect of sampling on two error categories. When sampling from a larger pool of words (higher $p$ and $t$), a model is more likely to veer \nameOffPrompt, but less likely to produce \nameRedundant~text.
}
\label{fig:heatmaps-per-label}
\end{figure}

%% file: fig/heatmaps-frequency-penalty-fig.tex
\begin{figure}[ht]

\begin{center}
\includegraphics[width=0.49\linewidth]{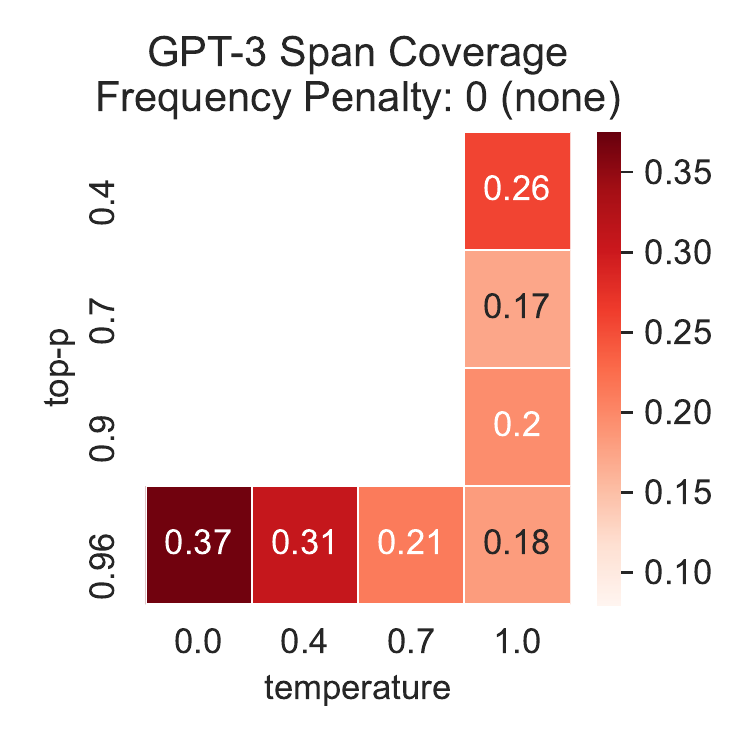}
\includegraphics[width=0.49\linewidth]{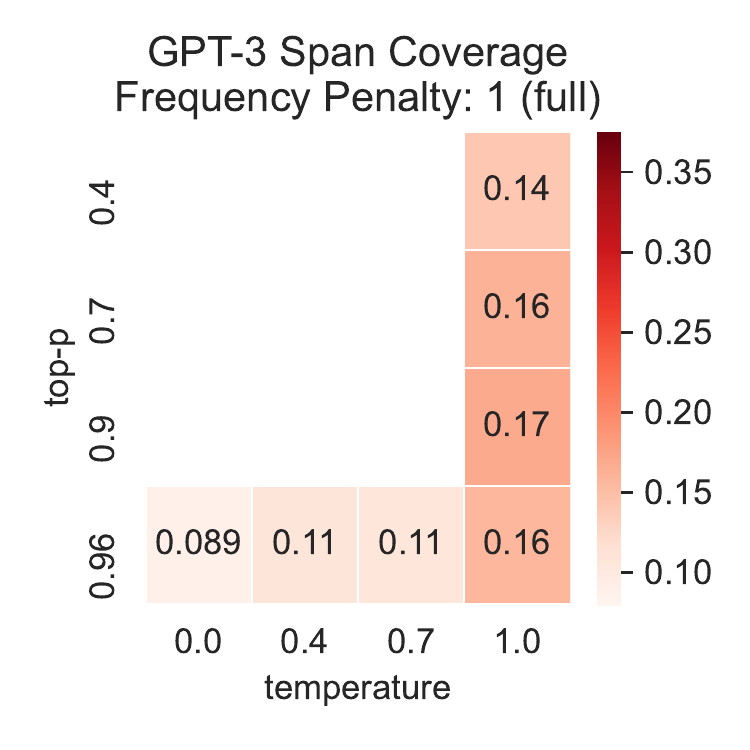}
\end{center}

\caption{
Comparison of \textit{frequency penalty} off (left) and full (right) for GPT-3 (removing reader issues and severity-1 \nameGrammarUsage~errors; argmax sampling is agnostic to the top-p value, so we simply plot it in the $p=0.96$ cell). We observe the frequency penalty improves average span coverage for all values of top-p and temperature. Furthermore its trend is reversed: with a frequency penalty, the least diverse sampling mechanisms (low temperature and low top-p) now produce text with the fewest error spans, rather than the most. (See Figure \ref{fig:model-variants} for confidence intervals on each value.)
}
\label{fig:heatmaps-frequency-penalty}
\end{figure}

%% file: fig/whats-next-figs.tex
\begin{figure*}[ht!]

\begin{center}

\includegraphics[height=0.30\linewidth]{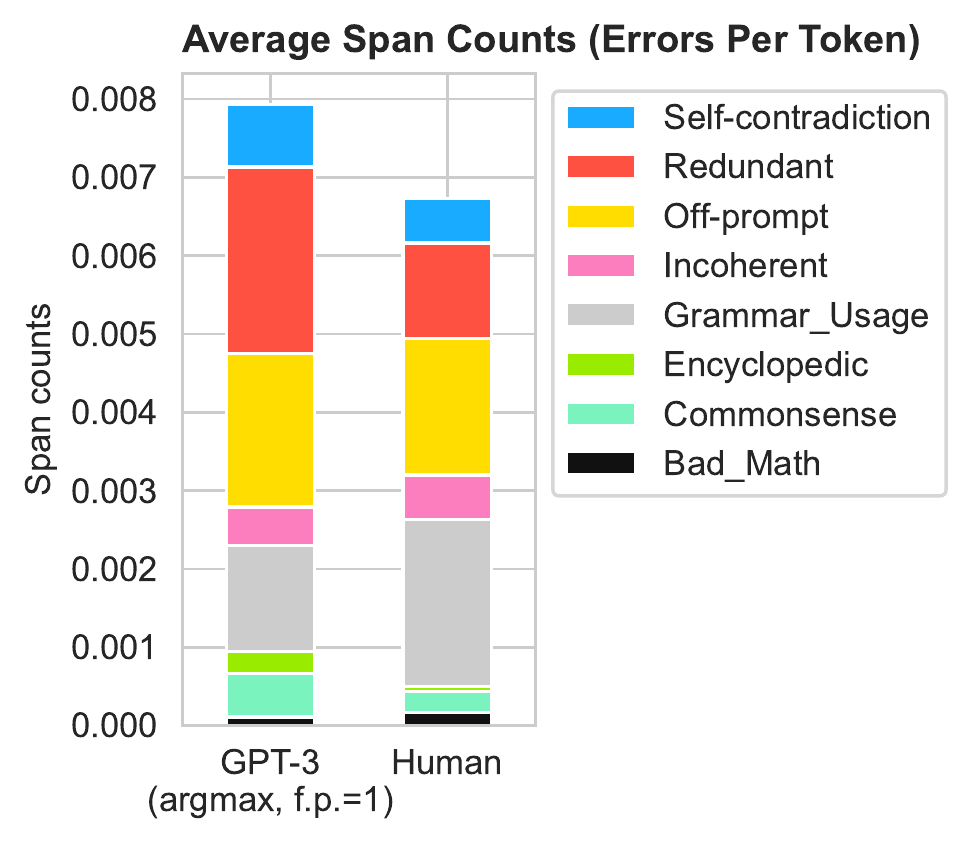}
\hspace{1em}
\includegraphics[height=0.30\linewidth]{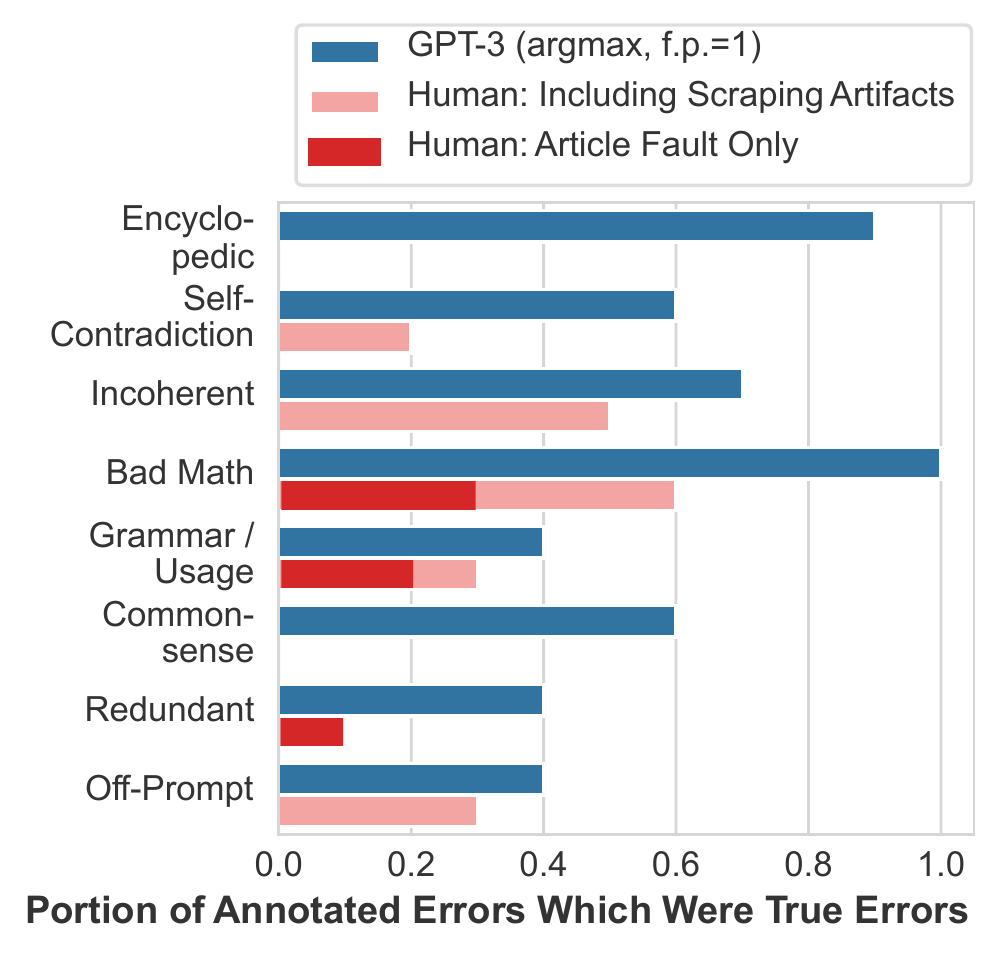}
\hspace{2em}
\includegraphics[height=0.30\linewidth]{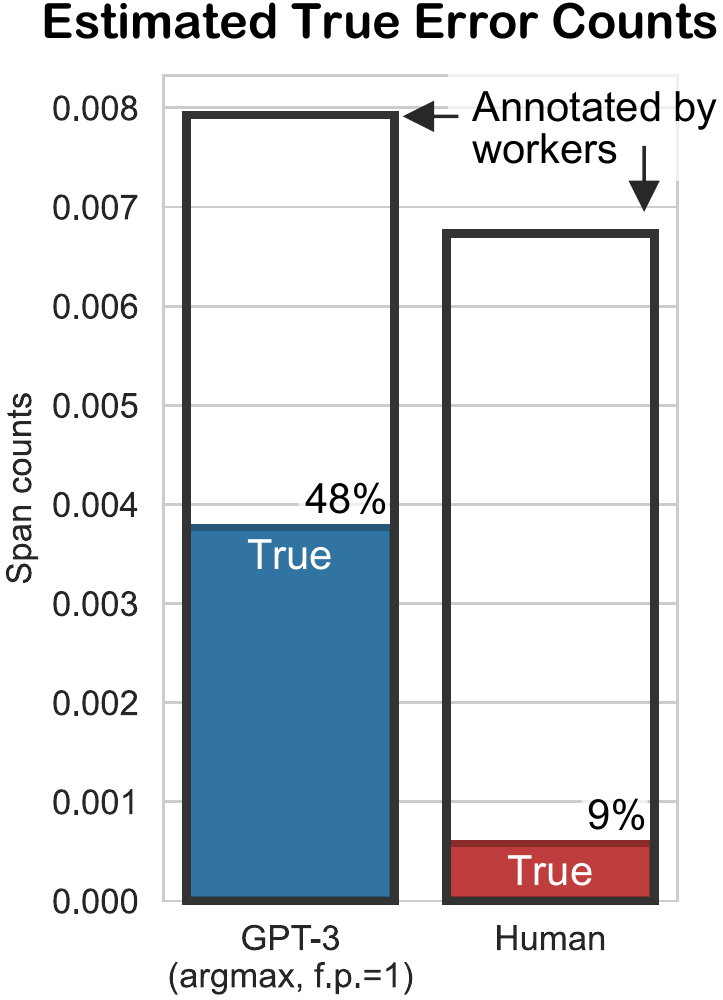}

\end{center}

\caption{
Analysis of the best GPT-3 configuration (\textit{argmax, freq. penalty = 1}) vs. human-authored text.
\textbf{Left:} A breakdown of errors by type. \textbf{Center:} Results of manually annotating 10 random spans from each type with whether the error was legitimate. For human-authored text, we also show errors marked on scraping artifacts that were present in the Common Crawl data. \textbf{Right:} Scaling each error type (\textit{left plot}, now shown in black outline) by the portion of errors found to be legitimate (\textit{center plot}), we estimate the true errors counts for each model (color-filled portions). \textbf{Takeaway:} Humans have more difficulty spotting errors in higher quality text; accounting for this difference dramatically increases the gap between model-authored and human-authored text. For simplicity, all plots use error \textit{counts} rather than error \textit{coverage}---i.e., they count the number of error spans, rather than scaling by the number of tokens covered.
}
\label{fig:whats-next-figs}
\end{figure*}

%% file: src/more-analysis.tex
\label{sec:further-analysis}


\subsection{Topics}

\input{fig/error-by-topic}

As noted in \S\ref{sec:data:prompt}, we collect data using prompts drawn primarily from 12--14 news topics. For conciseness, we show results only for GPT-3, and only for the standard apples-to-apples decoding configuration.

Figure~\ref{fig:errors-by-prompt} plots, based on the prompt topics, the average portion of the generation that is covered by error spans. While there is no significant difference between most topics, the results do indicate that generating text in more technical domains leads to higher span counts.

Figure \ref{fig:heatmaps-topic} shows individual span prevalence by topic. The top heatmap normalizes each topic (column) independently. \nameNeedsGoogle~issues and \nameOffPrompt~errors dominate the error types, with a few exceptions: for \textit{History}, and \textit{Nature} articles, \nameRedundant~trumps \nameOffPrompt~as a source of errors.

For the bottom, if we instead normalize by error label (row), we can observe which topics are more prone to certain error types than others. For example, we can see \nameBadMath~errors are most common in \textit{Business} and \textit{Health} generations; \textit{Entertainment} causes the most \nameSelfContradiction~errors; and \nameJargon~issues appears more frequently in articles about \textit{Business, Technology,} or \textit{Health}.

\subsection{Error explanations}

\input{fig/word-clouds}

\input{fig/explanation-lengths-fig}

\input{fig/error-type-explanations-fig}

Figure \ref{fig:word-clouds} displays word clouds for common unigrams and bigrams found in the error explanations for each error type, and Figure \ref{fig:explanation-lengths} shows the average explanation lengths for each error type. For \nameJargon, \nameRedundant, and \nameNeedsGoogle~error types, the prominent words do not provide much illumination and they have short average explanation length, indicating that the explanations are straightforward affirmations of the category (\textit{``I think this is financial jargon,''} \textit{``The information is repeated,''} or \textit{``I would need Google to check this.''}). But for categories like \nameEncyclopedic~and \nameBadMath, we observe some coarse trends: ``year'' is prevalent in both, ``movie'' appears in \nameEncyclopedic, and ``million'' is present in \nameBadMath, which suggests that the explanations are more likely from outside knowledge and needs some calculation (\textit{``The iPhone uses a lightening connector not a L-shaped connector,'' or \textit{``5000 feet is 1524 meters.''}}) 

Figure \ref{fig:error-type-examples} presents a few representative explanations for four error types, taking particular note of their explanation lengths (Figure \ref{fig:explanation-lengths}).
Both \nameSelfContradiction~and \nameRedundant~errors have antecedents, but their explanations are markedly different. Explanations for \nameSelfContradiction~contain more information describing the particular semantics that is reversed, which are less obvious at first glance than other errors. On the other hand, \nameRedundant~errors are more straightforward to spot, often involving simple lexical overlap, and so don't require elaboration.

Explanations for \nameCommonsense~contain the true commonsense knowledge that the text violates, which may take several words to explain. But an explanation for a \nameGrammarUsage~error simply corrects the error; as these errors are easier to fix, the explanation lengths are often short.

%% file: fig/error-by-topic.tex
\begin{figure}[t]

\begin{center}
\includegraphics[width=0.99\linewidth]{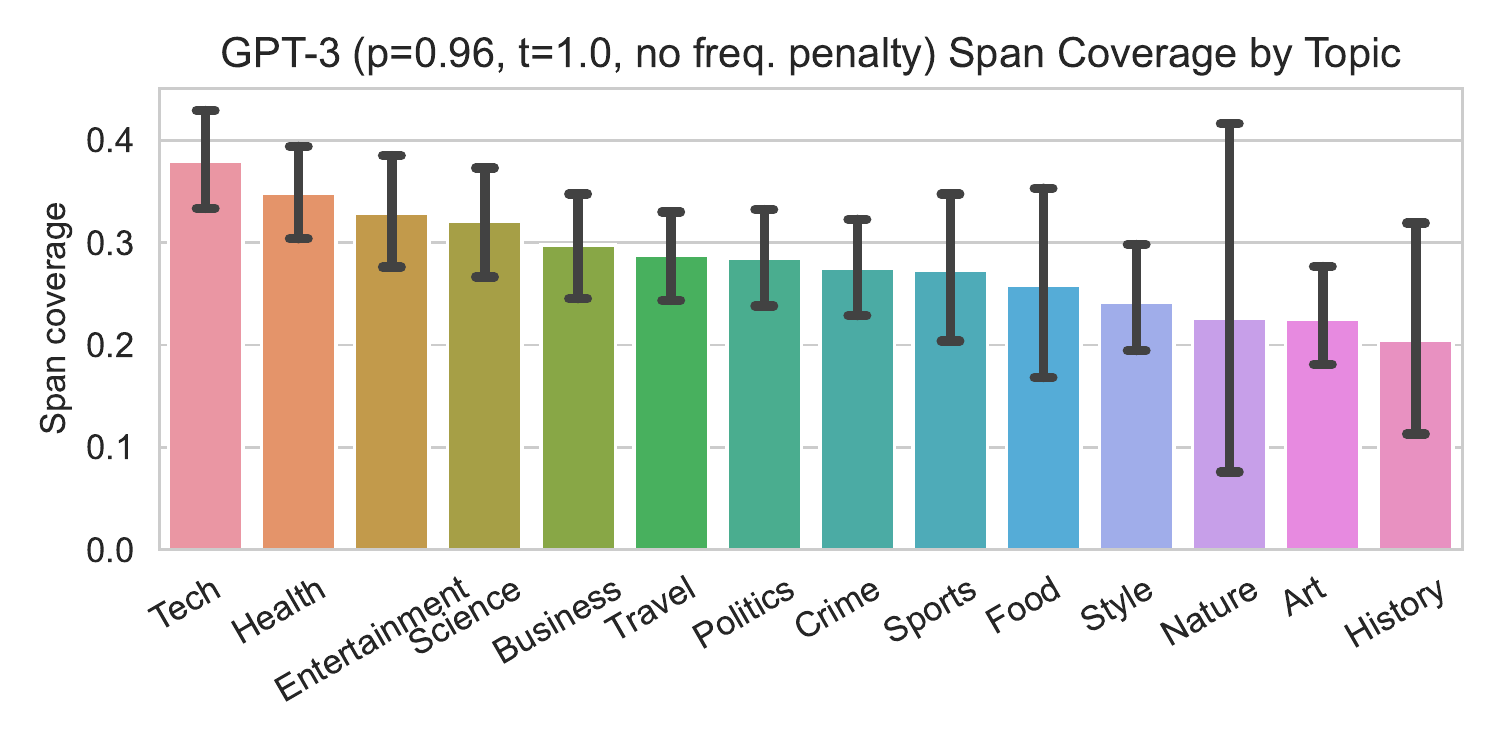}
\end{center}

\caption{
Average span coverage for different topics (GPT-3 generations with apples-to-apples decoding configuration), with 95\% confidence intervals. While the majority of topics display no significant trend, we observe that more technical topics such as \textit{Tech} and \textit{Health} are covered by a higher density of error spans than \textit{Style} and \textit{Art}.
}
\label{fig:errors-by-prompt}
\end{figure}

%% file: fig/word-clouds.tex
\begin{figure*}[ht!]

\begin{center}

\includegraphics[width=0.195\linewidth]{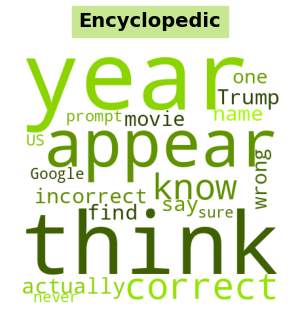}
\includegraphics[width=0.195\linewidth]{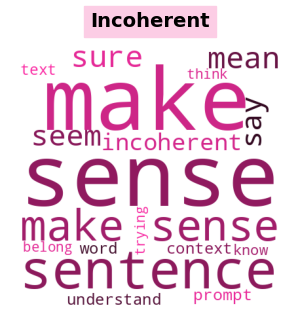}
\includegraphics[width=0.195\linewidth]{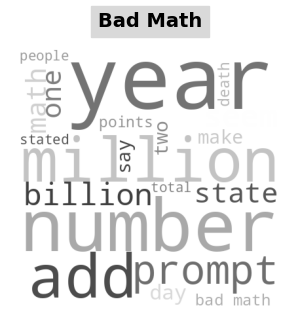}
\includegraphics[width=0.195\linewidth]{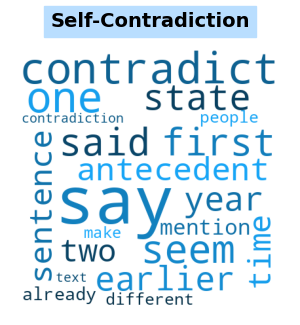}
\includegraphics[width=0.195\linewidth]{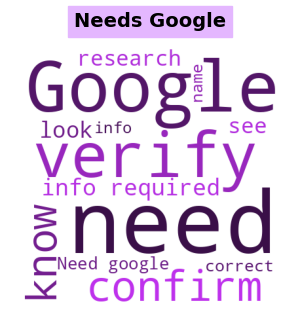}

\includegraphics[width=0.195\linewidth]{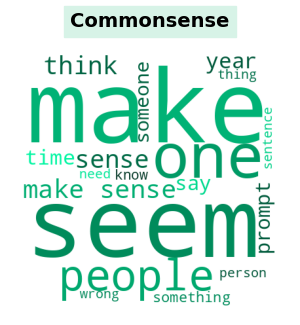}
\includegraphics[width=0.195\linewidth]{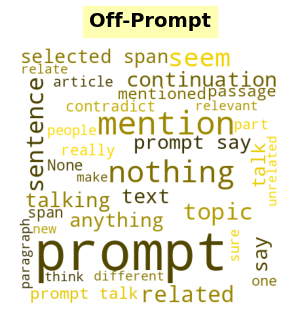}
\includegraphics[width=0.195\linewidth]{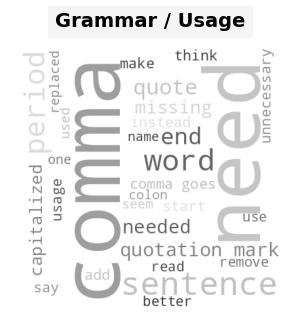}
\includegraphics[width=0.195\linewidth]{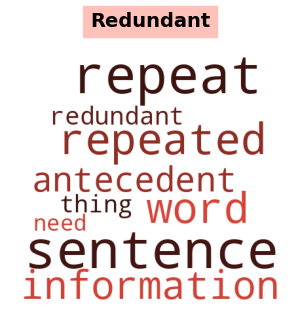}
\includegraphics[width=0.195\linewidth]{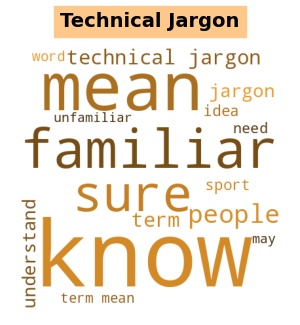}

\end{center}

\caption{
Common unigrams and bi-grams observed in the explanations written for each annotated span, grouped by error type.
}
\label{fig:word-clouds}
\end{figure*}

%% file: fig/explanation-lengths-fig.tex
\begin{figure}[ht]

\begin{center}
\includegraphics[width=0.99\linewidth]{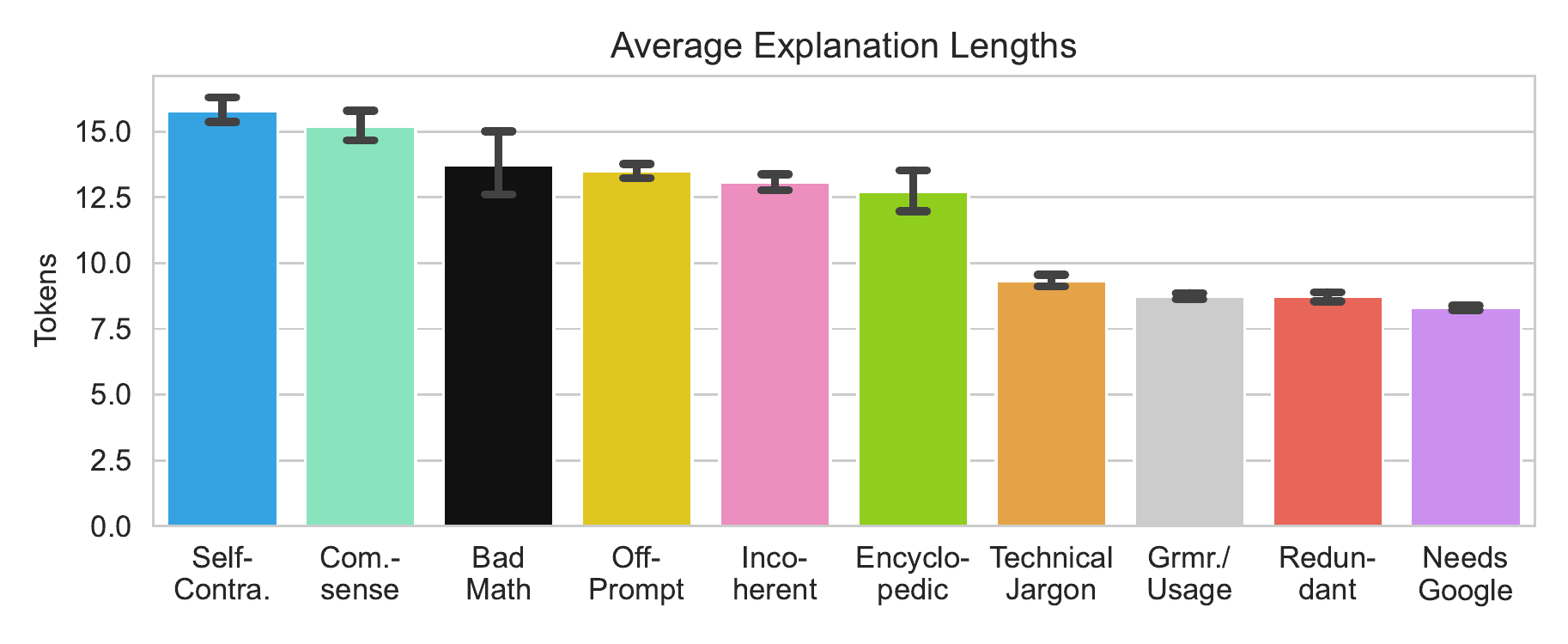}
\end{center}

\caption{
Average number of tokens in explanation for each error type.
We observe explanation length correlates with how obvious the error type is, where categories like \nameGrammarUsage~ and \nameJargon~are easier to find and explain than \nameSelfContradiction~and \nameCommonsense.
}
\label{fig:explanation-lengths}
\end{figure}

%% file: fig/error-type-explanations-fig.tex
\begin{figure}[t]

\begin{center}
\includegraphics[width=0.99\linewidth]{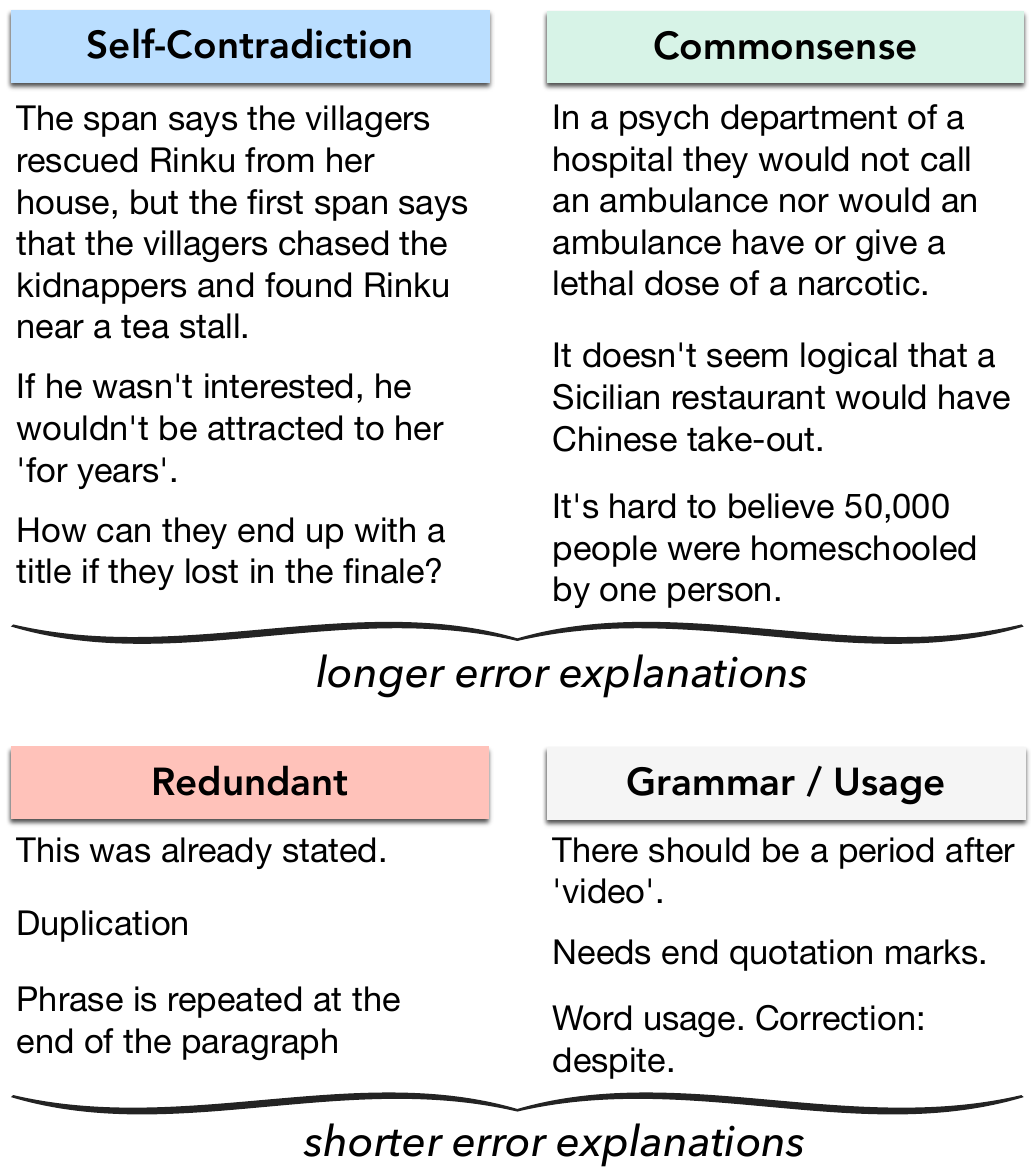}
\end{center}

\caption{
Examples of error explanations from different error types that favor longer (top) and shorter (bottom) descriptions.
}
\label{fig:error-type-examples}
\end{figure}

%% file: src/future-work.tex
\section{Future Work}
\label{sec:future-work}

We outline several further directions of study centering around the \method annotation framework, considering both natural implications and broader steps.

\subsection{\method Studies: Simple}

\paragraph{Find the best-performing GPT-3 decoding hyperparameters.}
We observed that for GPT-3, a frequency penalty value of 1 with argmax sampling produced fewer error spans than any other configuration (Fig. \ref{fig:model-variants}).
We have not tried varying the frequency penalty to values \textit{between} 0 and 1, or adding any \textit{presence penalty} (\S\ref{sec:decoding-strategies}), both of which then allow for fresh explorations of top-$p$ and temperature.

\paragraph{Study decoding parameters in smaller models.}
How good can (a finetuned) GPT-2 get? We saw decoding parameters considerably impacted GPT-3's performance, moving it from edging out Grover to error rates close to humans (Fig. \ref{fig:model-variants}). Could such decoding changes have a similar effect on a GPT-2-sized model?  Or might a smaller model favor different decoding hyperparameteres?

\paragraph{Back-off annotations.}
We observed good annotator agreement given the complexity of the task, but the odds that two annotators agree exactly on each span's type and boundaries remains only moderate (\S\ref{sec:data-quality}). We did not try backing-off (a) error types into coarser categories (e.g., language, factual, reader issue) or even to binary presence; (b) span boundaries into phrase or sentence-level annotations. Applying a type of back-off could also allow clustering methods to discover different error ontologies.

\paragraph{Improve automatic error detection.} While we present baseline results for automatic span error detection (\S\ref{sec:prediction}), we anticipate that significant progress is still available in this new task.

\subsection{\method Studies: Complex}

\paragraph{Align multiple annotations.}
In the current work, we largely treat annotators independently, with the exception of measuring their overlap to study agreement (\S\ref{sec:data-quality}) or taking their union to train prediction model (\S\ref{sec:prediction}). However, we might consider other ways of viewing the 10 annotations for each generation together. For example, we might consider the aggregate decision of \textit{whether} a token is labeled with \textit{any} span a measure of how noticeable or jarring an error is. This measure may be related to error severity, but may be distinct from it. 

One might also consider formal methods for computing annotation alignments. The Gamma measure, proposed by \newcite{mathet2015unified}, satisfies the long list of criteria needed to align and measure \method annotations: spans of multiple types, with gaps, full and partial span overlap, more than three annotators, and the potential to merge or split annotations (which we have not addressed in this paper). While we performed experiments with this measure, we experienced difficulties producing intuitive alignments with the authors' software, which disallows configuring parameters of the mixed-integer programming problem.\footnote{The mixed-integer programming approach is also computationally intensive; e.g., memory alone prevented us from computing alignments for pilot studies with twenty annotators, even on a machine with 500GB of RAM.} Emerging concurrent work \cite{titeux2021pygamma} offers a reimplementation of this measure that exposes additional parameters, which may be a promising avenue. However, it is possible that aligning annotations is a challenging task on its own that might require use of the explanations.

\paragraph{Characterize error nuance.}
Related to the previous point about error alignment, one might study whether model size affects span agreement. Anecdotally, errors from larger models like GPT-3---even of the same type, like \nameCommonsense~errors---are more difficult to describe without careful consideration, and may also be more difficult to identify.

\paragraph{Characterize repetition.}
Our quantitative studies of \nameRedundant~errors (e.g., Figs. \ref{fig:heatmaps-per-label} and \ref{fig:error-measuring-comparison}) point to semantic repetition as the major issue that emerges as models are scaled. Though this effect may be mitigated by changes to the decoding algorithm (like the frequency penalty), we still observe that models have difficulty striking a balance of repetition. With excessive paraphrasing, generated text seems \textit{stuck} on an idea. But equally, if a generation moves too quickly between ideas without linking them together or to an overall theme, the text lacks coherence. We posit that the issue of \nameRedundant~text emerges as the shadow of encompassing issues of narrative structure and discourse.

\subsection{Broadening \method}

\paragraph{Constrained generation}
This paper focuses on open-ended generation, but a natural extension of this method would be to assessing constrained generation tasks, such as machine translation.

\paragraph{New error types}
Especially if considering a novel task setting, new error types may prove useful. For example, in constrained generation, one might consider an \futureAdequacy~error, which---as in machine translation---would indicate that the meaning of a span diverges from what is expected given the generation constraints. Furthermore, one might need to introduce annotations on the provided (not generated) text to account for desired semantic components that are \textit{missing} from the generated text. Or, perhaps for a dialog setting, one might introduce a \futureGeneric~label, which would indicate that a portion of the generation is otherwise coherent and correct, but offers a lack of new information.\footnote{Such generic language may be seen as violating Grice's Maxims \cite{grice1975logic}, for example, by providing a dearth of information \textit{quantity}, or by flouting improper \textit{manner} by lacking brevity.} 

\paragraph{Corpus-level evaluation}
Other work has considered the evaluation of natural language generations at-scale, looking at distributional properties of the text \cite{caccia2020language,pillutla2021mauve}. We suggest that these views are complementary to instance-based, human evaluation proposed here, and combining the approaches could lead towards a more holistic view of generative evaluation. For example, while all \nameSelfContradiction~errors right now are \textit{within-document}, one could similarly identify \textit{cross-document} contradiction errors, where a model is inconsistent at a more global scale.

\subsection{Applications}

\paragraph{Detecting factuality}
One potential application of the \method data could be using the \nameNeedsGoogle~spans as a dataset of its own. In addition to training models to identify spans that require verification, one could go a step further and consider \textit{evidence retrieval} for each span, and even propose a classification task.\footnote{Minimally, \nameNeedsGoogle~spans from human-authored reputable news text should (hopefully) all be factually correct.}

\paragraph{Editing errors}
One errors can be detected, can they be fixed? The difficulty and scope of fixing \method-identified errors may depend on the error type, as error fixes may have cascading effects in the rest of the document.

%% file: acl.bbl
\begin{thebibliography}{40}
\expandafter\ifx\csname natexlab\endcsname\relax\def\natexlab#1{#1}\fi

\bibitem[{Banerjee and Lavie(2005)}]{banerjee2005meteor}
Satanjeev Banerjee and Alon Lavie. 2005.
\newblock Meteor: An automatic metric for mt evaluation with improved
  correlation with human judgments.
\newblock In \emph{Proceedings of the acl workshop on intrinsic and extrinsic
  evaluation measures for machine translation and/or summarization}, pages
  65--72.

\bibitem[{Branwen(2020)}]{branwen_2020}
Gwern Branwen. 2020.
\newblock \href {https://www.gwern.net/GPT-3} {Gpt-3 creative fiction}.

\bibitem[{Brown et~al.(2020)Brown, Mann, Ryder, Subbiah, Kaplan, Dhariwal,
  Neelakantan, Shyam, Sastry, Askell, Agarwal, Herbert-Voss, Krueger, Henighan,
  Child, Ramesh, Ziegler, Wu, Winter, Hesse, Chen, Sigler, Litwin, Gray, Chess,
  Clark, Berner, McCandlish, Radford, Sutskever, and
  Amodei}]{brown2020language}
Tom~B. Brown, Benjamin Mann, Nick Ryder, Melanie Subbiah, Jared Kaplan,
  Prafulla Dhariwal, Arvind Neelakantan, Pranav Shyam, Girish Sastry, Amanda
  Askell, Sandhini Agarwal, Ariel Herbert-Voss, Gretchen Krueger, Tom Henighan,
  Rewon Child, Aditya Ramesh, Daniel~M. Ziegler, Jeffrey Wu, Clemens Winter,
  Christopher Hesse, Mark Chen, Eric Sigler, Mateusz Litwin, Scott Gray,
  Benjamin Chess, Jack Clark, Christopher Berner, Sam McCandlish, Alec Radford,
  Ilya Sutskever, and Dario Amodei. 2020.
\newblock \href {http://arxiv.org/abs/2005.14165} {Language models are few-shot
  learners}.

\bibitem[{Caccia et~al.(2020)Caccia, Caccia, Fedus, Larochelle, Pineau, and
  Charlin}]{caccia2020language}
Massimo Caccia, Lucas Caccia, William Fedus, Hugo Larochelle, Joelle Pineau,
  and Laurent Charlin. 2020.
\newblock \href {http://arxiv.org/abs/1811.02549} {Language gans falling
  short}.

\bibitem[{Card et~al.(2020)Card, Henderson, Khandelwal, Jia, Mahowald, and
  Jurafsky}]{card.2020}
Dallas Card, Peter Henderson, Urvashi Khandelwal, Robin Jia, Kyle Mahowald, and
  Dan Jurafsky. 2020.
\newblock With little power comes great responsibility.
\newblock In \emph{Proceedings of EMNLP}.

\bibitem[{Carlini et~al.(2021)Carlini, Tram{\`e}r, Wallace, Jagielski,
  Herbert-Voss, Lee, Roberts, Brown, Song, Erlingsson, Oprea, and
  Raffel}]{Carlini2021ExtractingTD}
Nicholas Carlini, Florian Tram{\`e}r, Eric Wallace, Matthew Jagielski, Ariel
  Herbert-Voss, Katherine Lee, Adam Roberts, Tom~B. Brown, Dawn~Xiaodong Song,
  {\'U}lfar Erlingsson, Alina Oprea, and Colin Raffel. 2021.
\newblock Extracting training data from large language models.
\newblock In \emph{USENIX Security Symposium}.

\bibitem[{Celikyilmaz et~al.(2021)Celikyilmaz, Clark, and
  Gao}]{celikyilmaz2021evaluation}
Asli Celikyilmaz, Elizabeth Clark, and Jianfeng Gao. 2021.
\newblock \href {http://arxiv.org/abs/2006.14799} {Evaluation of text
  generation: A survey}.

\bibitem[{Clark et~al.(2021)Clark, August, Serrano, Haduong, Gururangan, and
  Smith}]{clark-etal-2021-thats}
Elizabeth Clark, Tal August, Sofia Serrano, Nikita Haduong, Suchin Gururangan,
  and Noah~A. Smith. 2021.
\newblock \href {https://doi.org/10.18653/v1/2021.acl-long.565} {All that{'}s
  {`}human{'} is not gold: Evaluating human evaluation of generated text}.
\newblock In \emph{Proceedings of the 59th Annual Meeting of the Association
  for Computational Linguistics and the 11th International Joint Conference on
  Natural Language Processing (Volume 1: Long Papers)}, pages 7282--7296,
  Online. Association for Computational Linguistics.

\bibitem[{Clark and Smith(2021)}]{clark-smith-2021-choose}
Elizabeth Clark and Noah~A. Smith. 2021.
\newblock \href {https://www.aclweb.org/anthology/2021.naacl-main.279} {Choose
  your own adventure: Paired suggestions in collaborative writing for
  evaluating story generation models}.
\newblock In \emph{Proceedings of the 2021 Conference of the North American
  Chapter of the Association for Computational Linguistics: Human Language
  Technologies}, pages 3566--3575, Online. Association for Computational
  Linguistics.

\bibitem[{Dugan et~al.(2020)Dugan, Ippolito, Kirubarajan, and
  Callison-Burch}]{dugan2020roft}
Liam Dugan, Daphne Ippolito, Arun Kirubarajan, and Chris Callison-Burch. 2020.
\newblock Roft: A tool for evaluating human detection of machine-generated
  text.
\newblock \emph{arXiv preprint arXiv:2010.03070}.

\bibitem[{Gao et~al.(2020)Gao, Fisch, and Chen}]{gao2020making}
Tianyu Gao, Adam Fisch, and Danqi Chen. 2020.
\newblock Making pre-trained language models better few-shot learners.
\newblock \emph{arXiv preprint arXiv:2012.15723}.

\bibitem[{Grice(1975)}]{grice1975logic}
Herbert~P Grice. 1975.
\newblock Logic and conversation.
\newblock In \emph{Speech acts}, pages 41--58. Brill.

\bibitem[{Gu et~al.(2021)Gu, yang Wu, and Yu}]{Gu2021PerceptionSA}
Jing Gu, Qing yang Wu, and Zhou Yu. 2021.
\newblock Perception score: A learned metric for open-ended text generation
  evaluation.
\newblock In \emph{AAAI}.

\bibitem[{Guan and Huang(2020)}]{Guan2020UNIONAU}
Jian Guan and Minlie Huang. 2020.
\newblock Union: An unreferenced metric for evaluating open-ended story
  generation.
\newblock In \emph{EMNLP}.

\bibitem[{Hashimoto et~al.(2019)Hashimoto, Zhang, and
  Liang}]{hashimoto-etal-2019-unifying}
Tatsunori Hashimoto, Hugh Zhang, and Percy Liang. 2019.
\newblock \href {https://doi.org/10.18653/v1/N19-1169} {Unifying human and
  statistical evaluation for natural language generation}.
\newblock In \emph{Proceedings of the 2019 Conference of the North {A}merican
  Chapter of the Association for Computational Linguistics: Human Language
  Technologies, Volume 1 (Long and Short Papers)}, pages 1689--1701,
  Minneapolis, Minnesota. Association for Computational Linguistics.

\bibitem[{Holtzman et~al.(2018)Holtzman, Buys, Forbes, Bosselut, Golub, and
  Choi}]{holtzman-etal-2018-learning}
Ari Holtzman, Jan Buys, Maxwell Forbes, Antoine Bosselut, David Golub, and
  Yejin Choi. 2018.
\newblock \href {https://doi.org/10.18653/v1/P18-1152} {Learning to write with
  cooperative discriminators}.
\newblock In \emph{Proceedings of the 56th Annual Meeting of the Association
  for Computational Linguistics (Volume 1: Long Papers)}, pages 1638--1649,
  Melbourne, Australia. Association for Computational Linguistics.

\bibitem[{Holtzman et~al.(2020)Holtzman, Buys, Forbes, and
  Choi}]{holtzman2019curious}
Ari Holtzman, Jan Buys, Maxwell Forbes, and Yejin Choi. 2020.
\newblock The curious case of neural text degeneration.
\newblock \emph{International Conference on Learning Representations}.

\bibitem[{Howcroft et~al.(2020)Howcroft, Belz, Clinciu, Gkatzia, Hasan,
  Mahamood, Mille, van Miltenburg, Santhanam, and
  Rieser}]{howcroft-etal-2020-twenty}
David~M. Howcroft, Anya Belz, Miruna-Adriana Clinciu, Dimitra Gkatzia, Sadid~A.
  Hasan, Saad Mahamood, Simon Mille, Emiel van Miltenburg, Sashank Santhanam,
  and Verena Rieser. 2020.
\newblock \href {https://www.aclweb.org/anthology/2020.inlg-1.23} {Twenty years
  of confusion in human evaluation: {NLG} needs evaluation sheets and
  standardised definitions}.
\newblock In \emph{Proceedings of the 13th International Conference on Natural
  Language Generation}, pages 169--182, Dublin, Ireland. Association for
  Computational Linguistics.

\bibitem[{Khashabi et~al.(2021)Khashabi, Stanovsky, Bragg, Lourie, Kasai, Choi,
  Smith, and Weld}]{khashabi2021genie}
Daniel Khashabi, Gabriel Stanovsky, Jonathan Bragg, Nicholas Lourie, Jungo
  Kasai, Yejin Choi, Noah~A. Smith, and Daniel~S. Weld. 2021.
\newblock \href {http://arxiv.org/abs/2101.06561} {Genie: A leaderboard for
  human-in-the-loop evaluation of text generation}.

\bibitem[{Krippendorff(2018)}]{krippendorff2018content}
Klaus Krippendorff. 2018.
\newblock \emph{Content analysis: An introduction to its methodology}.
\newblock Sage publications.

\bibitem[{Lin(2004)}]{lin2004rouge}
Chin-Yew Lin. 2004.
\newblock Rouge: A package for automatic evaluation of summaries.
\newblock In \emph{Text summarization branches out}, pages 74--81.

\bibitem[{Liu and Singh(2004)}]{liu2004conceptnet}
Hugo Liu and Push Singh. 2004.
\newblock Conceptnet—a practical commonsense reasoning tool-kit.
\newblock \emph{BT technology journal}, 22(4):211--226.

\bibitem[{Mathet et~al.(2015)Mathet, Widl{\"o}cher, and
  M{\'e}tivier}]{mathet2015unified}
Yann Mathet, Antoine Widl{\"o}cher, and Jean-Philippe M{\'e}tivier. 2015.
\newblock The unified and holistic method gamma ($\gamma$) for inter-annotator
  agreement measure and alignment.
\newblock \emph{Computational Linguistics}, 41(3):437--479.

\bibitem[{Monz and de~Rijke(2001)}]{monz2001light}
Christof Monz and Maarten de~Rijke. 2001.
\newblock Light-weight entailment checking for computational semantics.
\newblock In \emph{Proc. of the third workshop on inference in computational
  semantics (ICoS-3)}.

\bibitem[{Novikova et~al.(2018)Novikova, Dusek, and
  Rieser}]{Novikova2018RankMERH}
Jekaterina Novikova, Ondrej Dusek, and Verena Rieser. 2018.
\newblock Rankme: Reliable human ratings for natural language generation.
\newblock In \emph{NAACL}.

\bibitem[{Papineni et~al.(2002)Papineni, Roukos, Ward, and
  Zhu}]{papineni2002bleu}
Kishore Papineni, Salim Roukos, Todd Ward, and Wei-Jing Zhu. 2002.
\newblock Bleu: a method for automatic evaluation of machine translation.
\newblock In \emph{Proceedings of the 40th annual meeting of the Association
  for Computational Linguistics}, pages 311--318.

\bibitem[{Pillutla et~al.(2021)Pillutla, Swayamdipta, Zellers, Thickstun, Choi,
  and Harchaoui}]{pillutla2021mauve}
Krishna Pillutla, Swabha Swayamdipta, Rowan Zellers, John Thickstun, Yejin
  Choi, and Zaid Harchaoui. 2021.
\newblock \href {http://arxiv.org/abs/2102.01454} {Mauve: Human-machine
  divergence curves for evaluating open-ended text generation}.

\bibitem[{Qi et~al.(2020)Qi, Zhang, Zhang, Bolton, and
  Manning}]{qi-etal-2020-stanza}
Peng Qi, Yuhao Zhang, Yuhui Zhang, Jason Bolton, and Christopher~D. Manning.
  2020.
\newblock \href {https://doi.org/10.18653/v1/2020.acl-demos.14} {{S}tanza: A
  python natural language processing toolkit for many human languages}.
\newblock In \emph{Proceedings of the 58th Annual Meeting of the Association
  for Computational Linguistics: System Demonstrations}, pages 101--108,
  Online. Association for Computational Linguistics.

\bibitem[{Radford et~al.(2019)Radford, Wu, Child, Luan, Amodei, and
  Sutskever}]{radford2019language}
Alec Radford, Jeffrey Wu, Rewon Child, David Luan, Dario Amodei, and Ilya
  Sutskever. 2019.
\newblock Language models are unsupervised multitask learners.
\newblock \emph{OpenAI blog}, 1(8):9.

\bibitem[{Reynolds and McDonell(2021)}]{reynolds2021prompt}
Laria Reynolds and Kyle McDonell. 2021.
\newblock Prompt programming for large language models: Beyond the few-shot
  paradigm.
\newblock In \emph{Extended Abstracts of the 2021 CHI Conference on Human
  Factors in Computing Systems}, pages 1--7.

\bibitem[{Richardson et~al.(2013)Richardson, Burges, and
  Renshaw}]{richardson2013mctest}
Matthew Richardson, Christopher~JC Burges, and Erin Renshaw. 2013.
\newblock Mctest: A challenge dataset for the open-domain machine comprehension
  of text.
\newblock In \emph{Proceedings of the 2013 conference on empirical methods in
  natural language processing}, pages 193--203.

\bibitem[{Sai et~al.(2020)Sai, Mohankumar, and Khapra}]{sai2020survey}
Ananya~B Sai, Akash~Kumar Mohankumar, and Mitesh~M Khapra. 2020.
\newblock A survey of evaluation metrics used for nlg systems.
\newblock \emph{arXiv preprint arXiv:2008.12009}.

\bibitem[{Schank and Abelson(1977)}]{schank1977scripts}
Roger~C Schank and Robert~P Abelson. 1977.
\newblock \emph{Scripts, plans, goals, and understanding: An inquiry into human
  knowledge structures}.
\newblock Psychology Press.

\bibitem[{Titeux and Riad(2021)}]{titeux2021pygamma}
Hadrien Titeux and Rachid Riad. 2021.
\newblock pygamma-agreement: Gamma $\gamma$ measure for inter/intra-annotator
  agreement in python.

\bibitem[{Vaswani et~al.(2017)Vaswani, Shazeer, Parmar, Uszkoreit, Jones,
  Gomez, Kaiser, and Polosukhin}]{vaswani2017attention}
Ashish Vaswani, Noam Shazeer, Niki Parmar, Jakob Uszkoreit, Llion Jones,
  Aidan~N Gomez, Lukasz Kaiser, and Illia Polosukhin. 2017.
\newblock Attention is all you need.
\newblock \emph{arXiv preprint arXiv:1706.03762}.

\bibitem[{Wadden et~al.(2019)Wadden, Wennberg, Luan, and
  Hajishirzi}]{wadden-etal-2019-entity}
David Wadden, Ulme Wennberg, Yi~Luan, and Hannaneh Hajishirzi. 2019.
\newblock \href {https://doi.org/10.18653/v1/D19-1585} {Entity, relation, and
  event extraction with contextualized span representations}.
\newblock In \emph{Proceedings of the 2019 Conference on Empirical Methods in
  Natural Language Processing and the 9th International Joint Conference on
  Natural Language Processing (EMNLP-IJCNLP)}, pages 5784--5789, Hong Kong,
  China. Association for Computational Linguistics.

\bibitem[{Wood et~al.(2018)Wood, Long, Feltwell, Rowland, Brooker, Mahoney,
  Vines, Barnett, and Lawson}]{wood2018rethinking}
Gavin Wood, Kiel Long, Tom Feltwell, Scarlett Rowland, Phillip Brooker, Jamie
  Mahoney, John Vines, Julie Barnett, and Shaun Lawson. 2018.
\newblock Rethinking engagement with online news through social and visual
  co-annotation.
\newblock In \emph{Proceedings of the 2018 CHI Conference on Human Factors in
  Computing Systems}, pages 1--12.

\bibitem[{Zellers et~al.(2021)Zellers, Holtzman, Clark, Qin, Farhadi, and
  Choi}]{zellers-etal-2021-turingadvice}
Rowan Zellers, Ari Holtzman, Elizabeth Clark, Lianhui Qin, Ali Farhadi, and
  Yejin Choi. 2021.
\newblock \href {https://www.aclweb.org/anthology/2021.naacl-main.386}
  {{T}uring{A}dvice: A generative and dynamic evaluation of language use}.
\newblock In \emph{Proceedings of the 2021 Conference of the North American
  Chapter of the Association for Computational Linguistics: Human Language
  Technologies}, pages 4856--4880, Online. Association for Computational
  Linguistics.

\bibitem[{Zellers et~al.(2019)Zellers, Holtzman, Rashkin, Bisk, Farhadi,
  Roesner, and Choi}]{zellers2019neuralfakenews}
Rowan Zellers, Ari Holtzman, Hannah Rashkin, Yonatan Bisk, Ali Farhadi,
  Franziska Roesner, and Yejin Choi. 2019.
\newblock \href
  {http://papers.nips.cc/paper/9106-defending-against-neural-fake-news.pdf}
  {Defending against neural fake news}.
\newblock In H.~Wallach, H.~Larochelle, A.~Beygelzimer, F.~d\textquotesingle
  Alch\'{e}-Buc, E.~Fox, and R.~Garnett, editors, \emph{Advances in Neural
  Information Processing Systems 32}, pages 9054--9065. Curran Associates, Inc.

\bibitem[{Zhang et~al.(2019)Zhang, Kishore, Wu, Weinberger, and
  Artzi}]{zhang2019bertscore}
Tianyi Zhang, Varsha Kishore, Felix Wu, Kilian~Q Weinberger, and Yoav Artzi.
  2019.
\newblock Bertscore: Evaluating text generation with bert.
\newblock \emph{arXiv preprint arXiv:1904.09675}.

\end{thebibliography}
